\documentclass[10pt,twocolumn,letterpaper]{article}

\PassOptionsToPackage{table}{xcolor}
\usepackage[pagenumbers]{cvpr}      % review version
\usepackage{cuted}
\usepackage{amsthm}
\usepackage{mathtools}
\usepackage{bm}
\usepackage{multirow}
\usepackage{makecell}
\usepackage{pifont}
\usepackage{colortbl}
\usepackage{microtype}
\usepackage{algorithm}
\usepackage[noend]{algpseudocode}
\usepackage{parskip}
\usepackage{times}              % serif font like the reference
\usepackage{booktabs}
\usepackage{multirow}
\usepackage{makecell}
\usepackage{xcolor}
\usepackage{soul}
\usepackage[table]{xcolor}
\definecolor{rowgray}{RGB}{221,228,238}

\newcommand{\cmark}{\ding{51}}
\newcommand{\xmark}{\ding{55}}
\newcommand{\pmark}{\ensuremath{\triangle}}
\newcommand{\ours}{\textsc{DiffMark}\xspace}
\newcommand{\R}{\mathbb{R}}

\DeclareMathOperator{\sg}{sg}

\newcommand{\tbd}[1]{{\color{red}#1}}
\newcommand{\rev}[1]{{\color{blue}#1}}
\theoremstyle{plain}
\newtheorem{theorem}{Theorem}
\newtheorem{proposition}{Proposition}

\newtheorem{lemma}{Lemma}
\theoremstyle{definition}

\newtheorem{assumption}{Assumption}

\definecolor{darkred}{RGB}{139,0,0}
\definecolor{cvprblue}{rgb}{0.21,0.49,0.74}
\newcommand{\recomp}[1]{\tbd{[recompute\if\relax\detokenize{#1}\relax\else: #1\fi]}}
\newcommand{\LamI}{\Lambda_{\mathcal I}}
\usepackage[pagebackref,breaklinks,colorlinks,allcolors=cvprblue]{hyperref}

\AtEndPreamble{%
  \crefname{theorem}{Theorem}{Theorems}\Crefname{theorem}{Theorem}{Theorems}%
  \crefname{proposition}{Proposition}{Propositions}\Crefname{proposition}{Proposition}{Propositions}%
  \crefname{corollary}{Corollary}{Corollaries}\Crefname{corollary}{Corollary}{Corollaries}%
  \crefname{lemma}{Lemma}{Lemmas}\Crefname{lemma}{Lemma}{Lemmas}%
  \crefname{assumption}{Assumption}{Assumptions}\Crefname{assumption}{Assumption}{Assumptions}%
  \crefname{definition}{Definition}{Definitions}\Crefname{definition}{Definition}{Definitions}%
  \crefname{figure}{Fig.}{Figs.}\Crefname{figure}{Figure}{Figures}%
}

\def\paperID{*****}
\def\confName{CVPR}
\def\confYear{2027}

\title{Learning to Watermark the Denoising Trajectory: Fast and Robust Watermarks for Diffusion Models}

\author{%
Hong-Hanh Nguyen-Le\textsuperscript{1*} \quad
Van-Tuan Tran\textsuperscript{2*} \quad
Thuc D. Nguyen\textsuperscript{3} \quad
Nhien-An Le-Khac\textsuperscript{1}\\[0.8em]
{\small \textsuperscript{1}School of Computer Science, University College Dublin, Ireland}\\
{\small \textsuperscript{2}School of Computer Science and Statistics, Trinity College Dublin, Ireland}\\
{\small \textsuperscript{3}Department of Knowledge Engineering, University of Science, VNU-HCMC, Vietnam}\\[0.5em]
{\small \textsuperscript{*}Equal contribution}\\[0.5em]
{\tt\small hong-hanh.nguyen-le@ucdconnect.ie \quad honghanh.nguyen2104@gmail.com}\\
}

\begin{document}
\maketitle

% =====================================================================
\begin{abstract}
Attributing diffusion-generated images to individual users requires robust recovery of per-image keys, fast verification, and unchanged generator weights. To satisfy these requirements, we present \ours, which learns to embed arbitrary 256-bit keys along the late denoising trajectory of a frozen generator.
We embed the watermark during the final stages of image generation and reinforce it with complementary patterns to improve robustness. The watermark can then be recovered directly from the image, even after spatial transformations, without knowing the key or reversing the generation process. To reduce training memory requirements, we introduce dual-path training: a shortened generation process teaches the encoder to embed the watermark, while images from the full process teach the decoder how to recover it.
Theoretically, we bound how much denoising changes a perturbation added during the final sampling steps, give a key-separation condition that we verify on the tested samplers, and identify conditions under which short-path training improves watermark decoding with the full sampler. Empirically, against 10 baselines on three datasets, we show that \ours balances robustness, image quality, and extraction efficiency. Across SD~2.1, SD3, and FLUX on three datasets, DiffMark achieves $99.4$--$100\%$ average TPR@1\%FPR under distortion, removal, and adversarial attacks, with $89.4$\,ms extraction, approximately $17\times$ faster than Tree-Ring and Gaussian Shading.
\end{abstract}

% =====================================================================
\section{Introduction}
\label{sec:intro}

Diffusion models~\cite{rombach2022high,podellsdxl,esser2024scaling} now generate photorealistic images at platform scale, raising questions of provenance, copyright, and accountability~\cite{ref11_zelensky_deepfake,ref12_biden_robocall,ref9_taylor_swift,ref8_hk_deepfake_fraud}. For a generation platform, detecting synthetic content alone does not identify the user associated with an image. Watermarking supports this attribution by embedding a recoverable user-specific signal during generation~\cite{nguyen2025survey}. Such a signal must be \textbf{fast to extract} for large-scale verification and \textbf{robust to manipulation} as images are edited or attacked. We focus on achieving these two properties jointly while supporting arbitrary per-image keys and preserving image quality.

% \begin{figure}[t]
%   \centering
%   \includegraphics[width=0.9\columnwidth,keepaspectratio]{figs/user-specific-watermark.pdf}
%   \caption{\textbf{Fast and robust user-specific watermarking.} Diffusion watermarking methods embed an arbitrary key per image and recover it without diffusion inversion. Robust extraction supports attribution after image manipulation.}
%   \label{fig:teaser}
%   \vspace{-0.4cm}
% \end{figure}

Existing methods leave a gap between extraction speed and robustness (\cref{tab:paradigms,tab:main_results}). \emph{Initial-noise} watermarks~\cite{wen2023tree,ci2024ringid,li2025shallow,yang2024gaussian} recover their signal through DDIM inversion~\cite{songdenoising}, requiring repeated denoiser evaluations. FARI~\cite{yangfari} and OSI~\cite{chen2026osi} accelerate extraction with learned modules, while \emph{weight-based} methods~\cite{fernandez2023stable,feng2024aqualora} also support direct decoding. However, fast extraction alone does not ensure robust recovery: FARI and the evaluated weight-based methods lose detection performance under distortion or removal attacks (\cref{tab:main_results}). OptMark~\cite{xing2026optmark} additionally requires per-image latent optimization during generation. The challenge is to learn an embedding that withstands image manipulation while allowing direct, efficient extraction.

\begin{figure*}[t]
  \centering
  \begin{subfigure}[t]{0.45\textwidth}
    \centering
    \includegraphics[height=4cm,keepaspectratio]{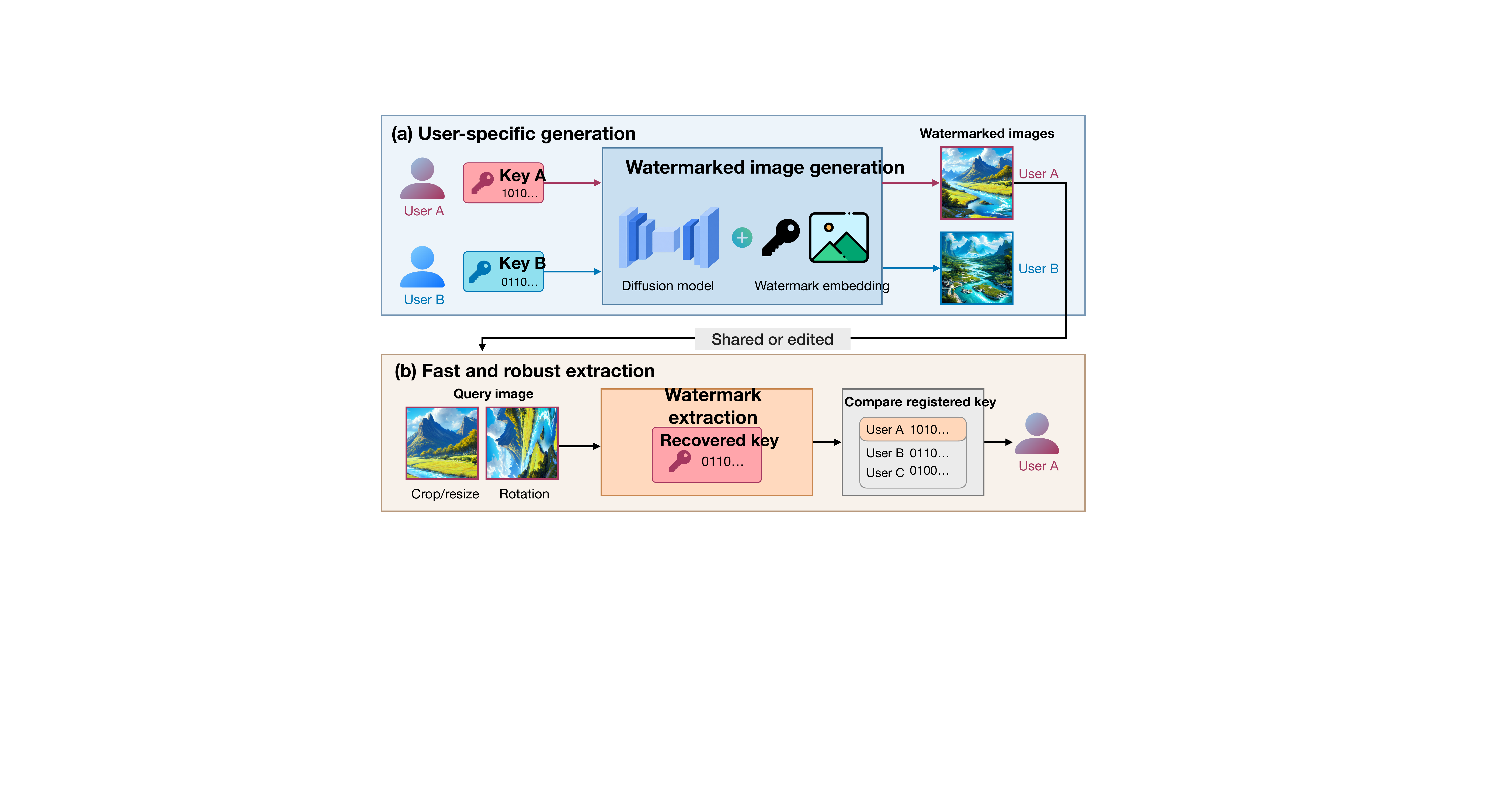}
    \caption{\textbf{Inference}: late-trajectory imprinting and fixed frequency-based patterns embed the key.}
    \label{fig:inference}
  \end{subfigure}\hfill
  \begin{subfigure}[t]{0.53\textwidth}
    \centering
    \includegraphics[height=4cm,keepaspectratio]{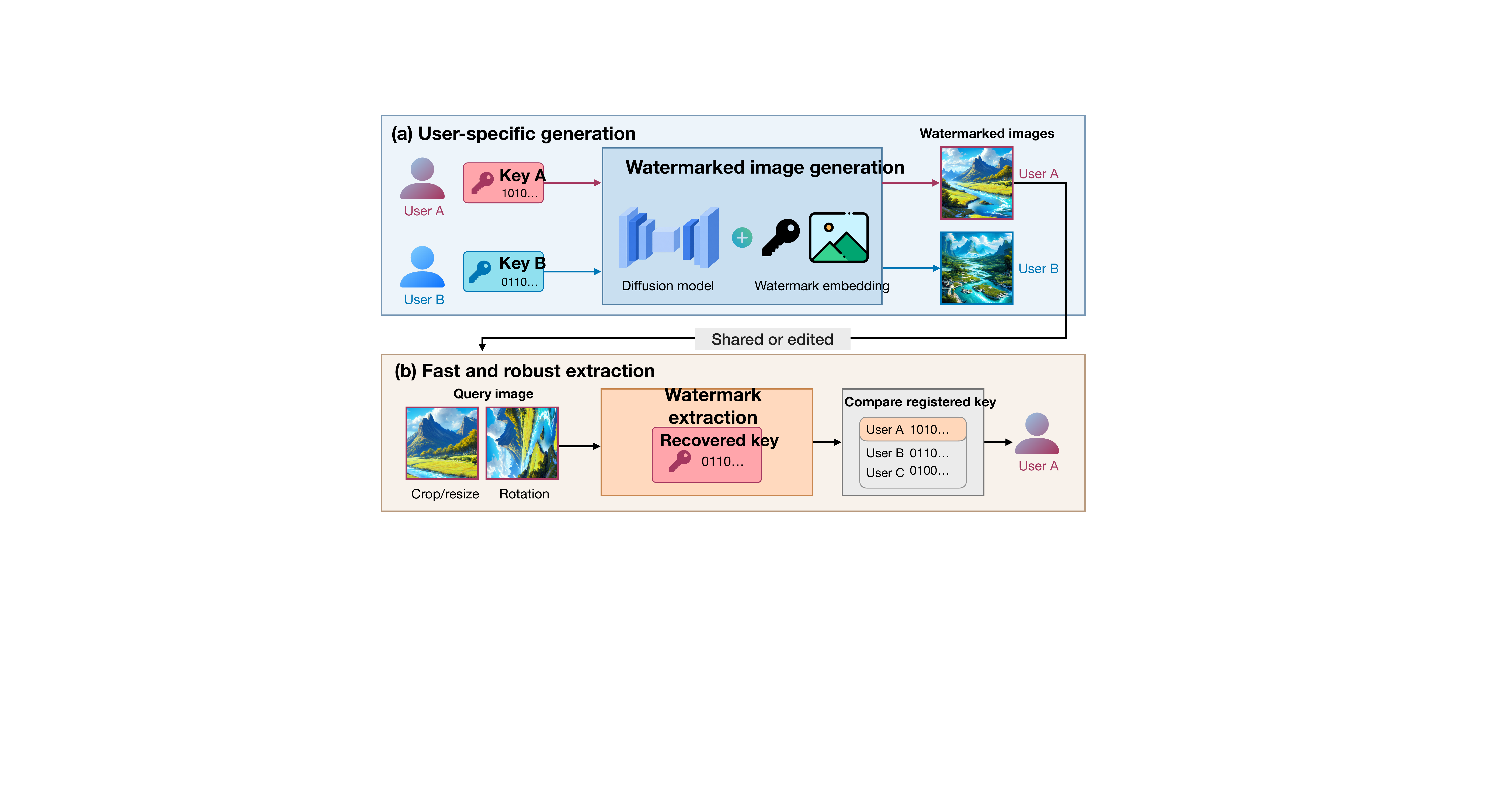}
    \caption{\textbf{Dual-path training}: a short differentiable run supplies gradients to $E_\phi$ via $\mathcal L_{\rm sh}$. The full sampler uses stop-gradient and calibrates $D_\psi$ via $\mathcal L_{\rm full}$ on clean and attacked latents.}
    \label{fig:dualpath}
  \end{subfigure}
  \caption{\textbf{Overview of \ours.} Robust embedding and direct image-only extraction are learned with dual-path training. The short path is used only during training.}
  \label{fig:overview}
  \vspace{-0.3cm}
\end{figure*}

To address this challenge, we propose \textbf{\ours}, centered on \textbf{learned late-trajectory imprinting} (\cref{fig:inference}). A watermark encoder maps an arbitrary 256-bit key to a latent perturbation. We repeatedly add this perturbation before each update in a short \emph{imprinting window} covering the final denoising steps, after most image content has formed. \textbf{For fast extraction}, a learned decoder and fixed frequency-based scores recover the key from VAE-encoded image latents without diffusion inversion. \textbf{For robust recovery}, fixed patterns embed the same bits at coarser spatial scales, embedding correction repairs weak bit margins, and a key-free alignment search compensates for spatial misalignment. These components combine direct decoding with recovery after spatial and pixel-level manipulation.

Learning this trajectory-based embedding requires gradients through sampling, but backpropagating through the full trajectory is memory-intensive and can be unstable~\cite{clarkdirectly}. We address this difficulty with \textbf{dual-path training}, which separates encoder learning from decoder calibration. A short differentiable run supplies encoder gradients, while a full run with a detached perturbation trains the decoder on clean and attacked latents. This split avoids backpropagation through the full sampler while exposing the decoder to its outputs. The generator weights remain unchanged, while the watermark modules are learned for the target backbone.

Our analysis bounds how much denoising changes the perturbation added during the final sampling steps (\cref{thm:channel}). In particular, a shorter imprinting window leaves fewer denoising steps to modify the perturbation. We also identify a condition under which distinct keys remain distinguishable at $z_0$ and verify it for every tested key pair on SD~2.1, SD3, and FLUX. For training, we give sufficient conditions for short-path updates to improve watermark decoding with the full sampler (\cref{thm:surrogate}) and empirically examine the alignment between short- and full-path gradients. These results support the embedding mechanism and its training procedure.

Our contributions are: \textbf{(1) Fast extraction without inversion.} \ours recovers arbitrary 256-bit keys directly from the image without diffusion inversion. Extraction takes $89.4$\,ms, approximately $17\times$ faster than Tree-Ring and Gaussian Shading. \textbf{(2) Robust watermark recovery.} Learned late-trajectory imprinting, frequency-based embedding, margin correction, and alignment search achieve $99.4$--$100\%$ average TPR@1\%FPR across attack groups on three datasets with SD~2.1, SD3, and FLUX. \textbf{(3) Theoretical analysis.} We characterize how the imprinting window affects watermark propagation, give a separation condition for distinct keys, and identify when short-path updates improve watermark decoding with the full sampler. We examine watermark propagation and gradient alignment on the tested samplers.

% =====================================================================
\begin{table}[t]
  \centering\footnotesize
  \setlength{\tabcolsep}{3pt}
  \caption{\textbf{Diffusion watermarking paradigms.} Fast: extraction without multi-step diffusion inversion. Key: arbitrary per-image keys. \cmark/\xmark/\pmark: yes/no/partial. $^\dagger$Requires per-image latent optimization during generation. Robustness is compared quantitatively in \cref{tab:main_results}.}
  \label{tab:paradigms}
  \resizebox{0.8\linewidth}{!}{%
  \begin{tabular}{@{}llccc@{}}
    \toprule
    Method & Carrier & Bits & Fast & Key \\
    \midrule
    TreeRing~\cite{wen2023tree} & $z_T$ & 0 & \xmark & \xmark \\
    RingID~\cite{ci2024ringid} & $z_T$ & 11 & \xmark & \pmark \\
    Shallow Diff.~\cite{li2025shallow} & $z_T$ & 0 & \xmark & \xmark \\
    Gauss.\ Shading~\cite{yang2024gaussian} & $z_T$ & 256 & \xmark & \cmark \\
    FARI-GS~\cite{yangfari} & $z_T$ & 256 & \cmark & \cmark \\
    OSI~\cite{chen2026osi} & $z_T$ & multi & \cmark & \cmark \\
    Stable Sig.~\cite{fernandez2023stable} & weights & 48 & \cmark & \xmark \\
    AquaLoRA~\cite{feng2024aqualora} & weights & 48 & \cmark & \pmark \\
    OptMark~\cite{xing2026optmark} & $z_t$ & multi & \cmark$^\dagger$ & \cmark \\
    \midrule
    \rowcolor{gray!15}\textbf{\ours (ours)} & trajectory, $z_0$ & \textbf{256} & \cmark & \cmark \\
    \bottomrule
  \end{tabular}}
  \vspace{-0.3cm}
\end{table}
\section{Related Work}
\label{sec:related}

\noindent\textbf{Watermarking diffusion models.} \emph{Initial-noise} methods encode the message in $z_T$ and detect it after DDIM inversion: TreeRing~\cite{wen2023tree} embeds Fourier rings for zero-bit detection, RingID~\cite{ci2024ringid} assigns ring patterns to users, Shallow Diffuse~\cite{li2025shallow} uses a low-dimensional subspace, and Gaussian Shading~\cite{yang2024gaussian} maps the message to the signs of $z_T$ via distribution-preserving sampling. ROBIN~\cite{huang2024robin} instead implants a signal at an intermediate state. FARI~\cite{yangfari} and OSI~\cite{chen2026osi} replace inversion with a learned one-step inverter or sign classifier. \emph{Weight-based} methods enable direct decoding by fine-tuning the VAE decoder~\cite{fernandez2023stable,xiong2023flexible} or merging a LoRA into the UNet~\cite{feng2024aqualora}. OptMark~\cite{xing2026optmark} optimizes latent watermarks per image during generation. \ours learns repeated late-trajectory imprinting together with direct decoding, and combines frequency-based embedding, margin correction, and alignment search to improve robustness without diffusion inversion.

\textbf{Differentiating through samplers.} Backpropagating through full sampling chains is memory-intensive and can be unstable, so DRaFT~\cite{clarkdirectly} truncates it to the last few steps, and consistency models~\cite{song2023consistency,luo2023latent} distill the probability-flow ODE into a few steps. We instead differentiate through a short run of the same sampler to learn the watermark encoder and calibrate the decoder on full-sampler outputs. Our analysis identifies conditions under which short-path updates improve watermark decoding with the full sampler.

% =====================================================================
\section{Method}
\label{sec:method}
\subsection{Preliminaries and goal}
\label{sec:prelim}
A latent generative model~\cite{rombach2022high} uses a VAE to encode an image $x$ as $z=\mathcal E(x)\in\R^{C\times h\times w}$. Generation starts from $z_T\sim\mathcal N(0,\mathbf I)$ and follows $N$ sampling steps conditioned on a prompt $c$. We write each update as $z_{k+1}=F_k(z_k)$ over timesteps $t_1>\dots>t_{N+1}$, with $z_1=z_T$ and $z_0:=z_{N+1}$. For a diffusion model with noise schedule $\bar\alpha_t$ and classifier-free guidance~\cite{ho2022classifier} of scale $w_g$, the DDIM update~\cite{songdenoising} is $F_k(z)=\sqrt{\bar\alpha_{t_{k+1}}}\,\frac{z-\sqrt{1-\bar\alpha_{t_k}}\,\hat\epsilon}{\sqrt{\bar\alpha_{t_k}}}+\sqrt{1-\bar\alpha_{t_{k+1}}}\,\hat\epsilon,$ where $\hat\epsilon=\epsilon_\theta(z,t_k,\varnothing)+w_g\,[\epsilon_\theta(z,t_k,c)-\epsilon_\theta(z,t_k,\varnothing)]$. For a flow-matching model~\cite{esser2024scaling}, the update $F_k(z)=z+(t_{k+1}-t_k)\,\hat v(z,t_k,c)$ is an Euler step using the guided velocity $\hat v$. Both updates are first-order solvers of a probability-flow ODE~\cite{song2021scorebased,lu2022dpmsolver}. The denoiser, VAE, and text encoders remain frozen.

\textbf{Goal.} Given a key $s\in\{0,1\}^L$ with $L=256$ and signs $\bar s=2s-1$, we learn a watermark encoder $E_\phi$ and decoder $D_\psi$, distinct from the VAE $(\mathcal E,\mathcal D)$. Our goals are fast image-only recovery without the prompt, initial noise, or diffusion inversion, and robust recovery under distortion, regeneration, and adversarial attacks. The watermarked image should remain close to an unwatermarked reference image generated from the same prompt and initial noise. Only the watermark modules are trained.

\subsection{Late trajectory imprinting}
\label{sec:late-injection}
The encoder maps the key to a latent perturbation $\delta=E_\phi(s)\in\R^{C\times h\times w}$ once per image. Before sampling step $k$, we add the scaled perturbation $\eta_k\delta$ to the latent  (\cref{fig:inference}):
\begin{equation}
  z_{k+1}=F_k\big(z_k+\eta_k\delta\big),\qquad k=1,\dots,N,
  \label{eq:injection}
\end{equation}
where $\eta_k\ge0$ controls the imprinting strength at step $k$, and $\bar\eta=\sum_k\eta_k$ is the total imprinting weight. We imprint only in a late window $\mathcal I$ containing the final $|\mathcal I|$ sampling steps, with equal weights $\eta_k=\bar\eta/|\mathcal I|$ for $k\in\mathcal I$ and $\eta_k=0$ otherwise. Imprinting is thus a repeated addition, once before every step of the window, rather than a one-shot stamp. Therefore, the same perturbation is added before each step in the window. Late imprinting occurs after the sampler has contracted low-variance latent directions. \cref{sec:theory} gives conditions that prevent cancellation across late steps and shows, under a Gaussian model, how late imprinting improves detectability per unit of $\delta$. The window $\mathcal I$ and total weight $\bar\eta$ can be adjusted at generation time using the same trained model. During training, we imprint at every step (\cref{sec:dualpath}).

\textbf{Watermark encoder.} $E_\phi$ sums embeddings specific to each bit's position and value, maps the result to spatial feature patterns, and applies three convolutional blocks with heads for the mean $\mu$ and log-variance. Architecture details are given in Appendix~\cref{sec:supp_architecture}.

\subsection{Frequency-Based Embedding and Correction}
\label{sec:frequency-based}
The learned perturbation mainly affects fine spatial details in the latent. To improve recovery under blur and small alignment errors, we also embed the same bits using fixed frequency-based patterns at coarser spatial scales. We restrict these patterns to a \emph{frequency band} corresponding to spatial periods between $p_{\rm lo}$ and $p_{\rm hi}$ latent pixels, where the two bounds specify the minimum and maximum retained periods. Within this band, we generate $L$ Gaussian fields with $n_c$ channels from a shared seed and orthonormalize them to obtain $\widetilde P_\ell$. An orthonormal matrix $V\in\R^{C\times n_c}$ maps the pattern channels to the $C$ latent channels: $P_\ell(:,y,x)=V\widetilde P_\ell(:,y,x)$. We choose $V$ using latent covariance, pattern responses after filtering and VAE re-encoding, and image distortion measured on a small set of generated images. The design favors strong extraction signals with small image changes and is computed once per generator with its weights frozen (Appendix~\cref{sec:supp_subcarrier}). After the last denoising step and before VAE decoding, we add the patterns once:
\begin{equation}
  z_0^{+}=z_0+\beta_{\rm sc}\,c_0\sum_{\ell=1}^{L}\bar s_\ell P_\ell,\; \text{with} \; c_0=\sqrt{Chw/L}.
  \label{eq:subcarrier}
\end{equation}
Because the patterns are orthonormal, $\beta_{\rm sc}$ is the root-mean-square amplitude of the added term over the latent coordinates. Adding the patterns only after denoising limits their effect on image content.

\textbf{Correction during embedding.} Since the key is known during embedding, we apply the extraction procedure in \cref{sec:extraction} to the generated image at its original alignment. For each bit whose signed margin $\bar s_\ell o_\ell$ falls below a target $\tau_c$, we strengthen its frequency-based pattern in proportion to the margin deficit divided by its direct frequency-score response (Appendix~\cref{sec:supp_correction}). We repeat this for up to $R$ rounds and retain the latent state producing the fewest bit errors, following informed embedding~\cite{miller2004applying}.

\subsection{Fast and Robust Watermark Extraction}
\label{sec:extraction}
Extraction uses VAE encoding, a learned decoder, and fixed-template correlations, avoiding the repeated denoiser evaluations of diffusion inversion. To suppress high-frequency perturbations, we first apply a narrow Gaussian filter and obtain $z=\mathcal E(G_{\sigma_f}*x)$. Geometric distortions are handled by the alignment search below.

\textbf{Watermark decoder.} $D_\psi$ extracts features from the latent $z$ using a strided CNN
and pools them to a $2\times2$ grid. An MLP maps the flattened features to two logits per bit. Their difference defines the decoder margin, $o^{\rm dec}_\ell(z)=[D_\psi(z)]_{\ell,1}-[D_\psi(z)]_{\ell,0}$. Positive margins favor bit~1. Architecture details are given in
Appendix~\cref{sec:supp_impl}.

\textbf{Frequency-based scores.} We compute $o^{\rm sc}_\ell(z)=\langle z,H_\ell\rangle$ with $H_\ell=\Lambda P_\ell$. The matrix $\Lambda\in\R^{C\times C}$ weights latent channels using measured covariance and normalizes the measured responses of the selected embedding directions. The resulting templates $H_\ell$ are computed once per generator and remain fixed during extraction (Appendix~\cref{sec:supp_subcarrier}).

\textbf{Alignment search.} Alignment requires the watermark patterns to match the spatial
positions, orientation, and scale expected by the extractor. To correct rotation and resized cropping without knowing the key, we use agreement between the decoder and frequency-based scores:
\begin{equation}
  \operatorname{Agr}(z)=\frac{1}{L}\sum_{\ell=1}^{L}
  \tanh\left(\frac{o^{\rm dec}_\ell(z)}{2}\right)
  \tanh\left(\frac{o^{\rm sc}_\ell(z)}{2}\right).
  \label{eq:agreement}
\end{equation}
We rank inverse rotations and crop corrections by agreement, apply the best candidates to the image, and refine after VAE re-encoding. We accept a correction only when its agreement
stands out among candidates and its decoder margins are sufficiently strong; otherwise, we retain the original alignment.

\textbf{Recovery and verification.} At the selected alignment, we combine the two scores,
$o(z)=o^{\rm dec}(z)+o^{\rm sc}(z)$, and recover each bit as
$\hat s_\ell=\mathbf 1[o_\ell(z)>0]$. We accept a registered key $s$ when $\text{Match}(x,s)=\sum_{\ell=1}^{L}\mathbf1[\hat s_\ell=s_\ell]\ge\tau$; among accepted keys, the highest count identifies the user. We calibrate $\tau$ on unwatermarked images using the complete extraction pipeline, including alignment search.

\subsection{Dual-path training}
\label{sec:dualpath}
Backpropagating through all $N$ denoising steps is memory-intensive and can be unstable~\cite{clarkdirectly}. We train $E_\phi$ and $D_\psi$ using two sampling paths of the frozen diffusion model that share the key $s$, initial noise $z_T$, and prompt $c$ (\cref{fig:dualpath}).

\textbf{Short path (encoder gradients).} A differentiable $K$-step run ($K\ll N$) with updates $F_k^{\text{sh}}$ imprints the perturbation before each step:
\begin{equation}
  z^{\rm sh}_{k+1}=F^{\rm sh}_k\big(z^{\rm sh}_k+ \eta^{\rm sh}_k\delta\big),\qquad k=1,\dots,K,
  \label{eq:shortpath}
\end{equation}
with $z^{\rm sh}_1=z_T$, short-path weights $\eta^{\rm sh}_k\ge0$, and $z^{\rm sh}_0:=z^{\rm sh}_{K+1}$. We use $\eta^{\rm sh}_k=1/K$. Its loss $\mathcal L_{\rm sh}=\mathcal L_{\rm rec}(o(z_0^{\rm sh}),s)$ and the imperceptibility terms train the encoder.

\textbf{Full path (decoder calibration).} The full sampler follows \cref{eq:injection} with the detached perturbation $\sg(\delta)$ and training imprinting weights $\eta_k=1/N$. Its outputs train the decoder on full-sampler latents and their transformed versions, with $\nabla_\phi\mathcal L_{\rm full}=0$. \cref{thm:surrogate} gives a sufficient condition for short-path updates to improve watermark decoding with the full sampler.

\textbf{Bit Reconstruction.} For bit $\ell$, we combine binary cross-entropy with a signed-margin hinge:
\begin{equation}
    \mathcal L_{\rm rec}(o,s)
      =\lambda_L\Big(\frac{1}{L}\sum_{\ell=1}^{L}\omega_\ell\ell_\ell
      +w_{\rm tail}\,\operatorname{TopMean}_{r}(\omega_\ell\ell_\ell)\Big),
  \label{eq:reconstruction}
\end{equation}
where $\ell_\ell = \log\big(1+e^{-\bar s_\ell o_\ell}\big)+ w_{\rm m}\,[\tau_{\rm m}-\bar s_\ell o_\ell]_+,$ $[a]_+=\max(a,0)$, $\tau_{\rm m}$ is the target margin, and $\operatorname{TopMean}_{r}$ averages the largest fraction $r$ of weighted bit losses. The weights $\omega_\ell$ emphasize bits with higher running error rates; $w_{\rm m}$, $w_{\rm tail}$, and $\lambda_L$ control the loss contributions and scale.

\textbf{Joint objective.} We optimize
\begin{equation}
  \begin{aligned}
    \mathcal L={}&\mathcal L_{\rm sh}+\mathcal L_{\rm full}
      + \chi_t\,\mathcal L_{\rm rec}\big(o(z_0^{\rm full,rt}),s\big) + w_{\rm mag}\mathcal L_{\rm mag} \\
      &+ \beta_t\mathcal L_{\rm KL} + w_{\rm div}\mathcal L_{\rm div}
      + w_{\rm prvl}\mathcal L_{\rm prvl} \\
      &+w_{\rm ld}\mathcal L_{\rm ld}
        + w_{\rm neg}\mathcal L_{\rm neg},
  \end{aligned}
  \label{eq:training_objective}
\end{equation}
where $z_0^{\rm full,rt}=\mathcal E(\mathcal D(z_0^{\rm full}))$ is a VAE round trip and $\chi_t \in \{0,1\}$ indicates whether its supervision is included at training step $t$. The latent penalty $\mathcal L_{\rm ld}=\|z_0^{\rm sh}-\sg(z_0^{\rm sh,\varnothing})\|_2^2$ compares against the clean short path from the same noise. The pixel penalty $\mathcal L_{\rm prvl}$~\cite{feng2024aqualora} measures the largest mean absolute difference from the corresponding unwatermarked image over local image regions. $\mathcal L_{\rm mag}$ controls perturbation amplitude, $\mathcal L_{\rm KL}$ regularizes an auxiliary Gaussian encoder distribution, and $\mathcal L_{\rm div}$ penalizes mean pairwise cosine similarity between perturbations. $\mathcal L_{\rm neg}$ is the negative mean decoder entropy on unwatermarked full-path latents, discouraging confident bit predictions. Details of these loss functions are provided in Appendix \cref{sec:supp_training}.

% =====================================================================
\section{Theoretical Analysis}
\label{sec:theory}

We analyze how late imprinting preserves the watermark and how short-path training supports recovery. \cref{thm:channel} bounds how much denoising changes the added perturbation at $z_0$ and gives a separation condition for distinct keys. \cref{thm:surrogate} gives sufficient conditions for short-path updates to decrease the full-sampler decoding loss. Proofs are in Appendix~\cref{sec:supp_proofs}, and Appendix~\cref{sec:supp_measured,sec:gradient-alignment} empirically examines watermark propagation and gradient alignment on the tested samplers.

\subsection{Late imprinting as direct addition}
Fix $z_T$ and $c$. The clean trajectory follows $z^\circ_{k+1}=F_k(z^\circ_k)$ from $z^\circ_1=z_T$. Let $\Phi(\delta)=z_{N+1}$ denote the final latent of the imprinted trajectory in \cref{eq:injection}, so $\Phi(0)=z^\circ_{N+1}$. Let $\nabla$ denote the Jacobian and define $J_k(\delta)=\nabla F_N\cdots\nabla F_k$, where each $\nabla F_j$ is evaluated at the input $z_j+\eta_j\delta$ of the imprinted trajectory. This Jacobian describes the first-order response of $z_0$ to a perturbation at the input of step $k$. We write $J_k=J_k(0)$ and $A=\nabla\Phi(0)=\sum_k\eta_kJ_k$. For $\delta\ne\delta'$ such that $\Phi$ is continuously differentiable along the straight-line path connecting the two perturbations, $\delta^{(s)}=\delta'+s(\delta-\delta')$, $s\in[0,1]$, let $u=(\delta-\delta')/\|\delta-\delta'\|$ and define the \emph{signed gain}
\begin{equation}
  \bar m_{\delta,\delta'}=\int_0^1u^\top\nabla\Phi(\delta^{(s)})\,u\,ds,
  \label{eq:signed_gain}
\end{equation}
which measures the average response along the direction between the two perturbations. Norms are Euclidean for latents and spectral for matrices, and $\lambda_{\min}$ denotes the smallest eigenvalue.

\begin{assumption}[Local Lipschitz regularity~{\cite{dmHairer1993}}]
\label{asm:regular}
Each $F_k$ is continuously differentiable on a neighborhood of the closed ball of radius $R_0>0$ centered at $z_k^\circ$. On this ball, it satisfies $\|\nabla F_k-\mathbf I\|\le e^{\ell_k}-1$ for some $\ell_k\ge0$.
\end{assumption}
Here, $e^{\ell_k}-1$ bounds how much step $k$ changes a small perturbation relative to its magnitude. To combine these bounds across successive steps, define $\Lambda_k=\sum_{j=k}^{N}\ell_j$ for the steps from $k$ to $N$. For the late imprinting window $\mathcal I=\{N-|\mathcal I|+1,\dots,N\}$, define $\LamI=\sum_{k\in\mathcal I}\ell_k$. The accumulated relative change is then bounded by $e^{\Lambda_k}-1$ over the remaining steps and by $e^{\LamI}-1$ over the window.

For exact updates of the sampling ODE, whose vector field has Lipschitz constant $L(t)$, we can choose $\ell_k=\int_{t_{k+1}}^{t_k}L(t)\,dt$~\cite{song2021scorebased,lu2022dpmsolver,dmHairer1993}. In this case, $\LamI$ depends on the time interval covered by the window, independently of how it is divided into sampling steps. The uniform choice $\ell_k=L_f/N$ instead uses the same worst-case constant at every step and gives $\LamI=L_f\rho$, where $\rho=|\mathcal I|/N$.

\begin{proposition}[Imprinting response at $z_0$]
\label{thm:channel}
Let $\eta_k\ge0$ with $\bar\eta=\sum_k\eta_k>0$.
\begin{enumerate}[label=(\roman*),leftmargin=*]
  \item \textbf{Late imprinting as direct addition.} Under \cref{asm:regular}, suppose $\eta_k=0$ outside $\mathcal I$. Let $r_{\mathcal I}=R_0e^{-\LamI}/\bar\eta$ and $\varepsilon_{\mathcal I} =\sum_{k\in\mathcal I}\eta_k(e^{\Lambda_k}-1) \le\bar\eta(e^{\LamI}-1)$. On $\{\delta:\|\delta\|\le r_{\mathcal I}\}$, $\Phi$ is $C^1$ with $\|\nabla\Phi(\delta)-\bar\eta\mathbf I\| \le\varepsilon_{\mathcal I}$. Hence, for any $\delta,\delta'$ in this ball, $\Phi(\delta)-\Phi(\delta')=\bar\eta(\delta-\delta')+E$ with $\|E\|\le\varepsilon_{\mathcal I}\|\delta-\delta'\|$, and for $\delta\ne\delta'$ the signed gain in \cref{eq:signed_gain} satisfies
  \begin{equation}
    \bar m_{\delta,\delta'}\ge\bar\eta-\varepsilon_{\mathcal I}\ge\bar\eta\big(2-e^{\LamI}\big).
    \label{eq:worst_case_gain}
  \end{equation}
  \item \textbf{Separation.} For any $\delta\ne\delta'$ such that $\Phi$ is $C^1$ on the segment between them,
  \begin{equation}
    \|\Phi(\delta)-\Phi(\delta')\|\ge u^\top\big(\Phi(\delta)-\Phi(\delta')\big)=\bar m_{\delta,\delta'}\|\delta-\delta'\|,
    \label{eq:channel_monotone}
  \end{equation}
  so the two perturbations remain distinguishable at $z_0$ whenever $\bar m_{\delta,\delta'}>0$. The signed gain is bounded below, uniformly over directions, by $\bar m_{\delta,\delta'}\ge\min_{s\in[0,1]}\lambda_{\min}\big(\tfrac12(\nabla\Phi+\nabla\Phi^\top)(\delta^{(s)})\big)$ and, along the direction $u$, in terms of the propagation Jacobians $J_k$:
  \begin{equation}
    \bar m_{\delta,\delta'}\ge\sum_{k=1}^{N}\eta_k
      \Bigl(1-\sup_{s\in[0,1]}\|J_k(\delta^{(s)})u-u\|\Bigr).
    \label{eq:directional_margin}
  \end{equation}
  This part does not require \cref{asm:regular} or a bound on the perturbation norms.
\end{enumerate}
\end{proposition}

\textbf{Main findings.} (a) \emph{Late imprinting approximately adds the perturbation to $z_0$.} Under the conditions in part~(i), the final latent difference deviates from $\bar\eta(\delta-\delta')$ by at most a fraction $e^{\LamI}-1$. This bound decreases with the sum of the stepwise constants $\LamI$ and is independent of the step count. Therefore, a shorter window limits how much the sampler can change the key perturbation, consistent with the window ablation in \cref{tab:ablation}.
(b) \emph{Distinct keys remain distinguishable.} The identity in part~(ii) allows us to compute $\bar m_{\delta,\delta'}$ directly from two generated latents without derivatives. A positive value guarantees a positive lower bound on their distance, so key separation can be checked for any pair.

\begin{table*}[ht]
    \centering
    \caption{\textbf{Watermark detection and robustness results with SD 2.1 on DiffusionDB, MS-COCO, and DALL-E3.} All values are TPR@1\%FPR, reported as mean with the standard deviation in subscript. The best result is in \textbf{bold}.}
    \label{tab:main_results}
    \resizebox{\textwidth}{!}{%
    \footnotesize
    \setlength{\tabcolsep}{4pt}
    \begin{tabular}{l cccc @{\hspace{10pt}} cccc @{\hspace{10pt}} cccc}
    \toprule
    \multirow{2}{*}{\textbf{Method}}
    & \multicolumn{4}{c}{\textbf{DiffusionDB}}
    & \multicolumn{4}{c}{\textbf{MS-COCO}}
    & \multicolumn{4}{c}{\textbf{DALL-E3}} \\
    \cmidrule(lr){2-5} \cmidrule(lr){6-9} \cmidrule(lr){10-13}
    & Clean & \makecell{Avg\\Distortion} & \makecell{Avg\\Removal} & \makecell{Avg\\Adversarial}
    & Clean & \makecell{Avg\\Distortion} & \makecell{Avg\\Removal} & \makecell{Avg\\Adversarial}
    & Clean & \makecell{Avg\\Distortion} & \makecell{Avg\\Removal} & \makecell{Avg\\Adversarial} \\
    \cmidrule(lr){2-2} \cmidrule(lr){3-3} \cmidrule(lr){4-4} \cmidrule(lr){5-5}
    \cmidrule(lr){6-6} \cmidrule(lr){7-7} \cmidrule(lr){8-8} \cmidrule(lr){9-9}
    \cmidrule(lr){10-10} \cmidrule(lr){11-11} \cmidrule(lr){12-12} \cmidrule(lr){13-13}
    % ---------- Zero-bit ----------
    \rowcolor{rowgray}
    TreeRing
    & \textbf{1.000}$_{\pm 0.000}$ & 0.778$_{\pm 0.006}$ & 0.985$_{\pm 0.003}$ & 0.841$_{\pm 0.018}$
    & \textbf{1.000}$_{\pm 0.000}$ & 0.774$_{\pm 0.009}$ & 0.971$_{\pm 0.003}$ & 0.859$_{\pm 0.005}$
    & \textbf{1.000}$_{\pm 0.000}$ & 0.773$_{\pm 0.009}$ & 0.959$_{\pm 0.003}$ & 0.840$_{\pm 0.003}$ \\
    ShallowDiffuse
    & \textbf{1.000}$_{\pm 0.000}$ & 0.628$_{\pm 0.002}$ & 0.978$_{\pm 0.005}$ & 0.545$_{\pm 0.011}$
    & \textbf{1.000}$_{\pm 0.000}$ & 0.629$_{\pm 0.002}$ & 0.991$_{\pm 0.002}$ & 0.703$_{\pm 0.008}$
    & \textbf{1.000}$_{\pm 0.000}$ & 0.628$_{\pm 0.002}$ & 0.981$_{\pm 0.002}$ & 0.708$_{\pm 0.017}$ \\
    \rowcolor{rowgray}
    FARI-T
    & \textbf{1.000}$_{\pm 0.000}$ & 0.760$_{\pm 0.003}$ & 0.974$_{\pm 0.003}$ & 0.907$_{\pm 0.013}$
    & \textbf{1.000}$_{\pm 0.000}$ & 0.760$_{\pm 0.004}$ & 0.959$_{\pm 0.008}$ & 0.939$_{\pm 0.009}$
    & \textbf{1.000}$_{\pm 0.000}$ & 0.759$_{\pm 0.004}$ & 0.939$_{\pm 0.003}$ & 0.901$_{\pm 0.005}$ \\
    \midrule
    % ---------- Multi-bit ----------
    StegaStamp
    & \textbf{1.000}$_{\pm 0.000}$ & 0.758$_{\pm 0.010}$ & 0.912$_{\pm 0.018}$ & \textbf{1.000}$_{\pm 0.000}$
    & \textbf{1.000}$_{\pm 0.000}$ & 0.741$_{\pm 0.005}$ & 0.922$_{\pm 0.004}$ & \textbf{1.000}$_{\pm 0.000}$
    & \textbf{1.000}$_{\pm 0.000}$ & 0.714$_{\pm 0.008}$ & 0.905$_{\pm 0.002}$ & \textbf{1.000}$_{\pm 0.000}$ \\
    \rowcolor{rowgray}
    RingID
    & \textbf{1.000}$_{\pm 0.000}$ & 0.883$_{\pm 0.003}$ & \textbf{1.000}$_{\pm 0.000}$ & 0.873$_{\pm 0.003}$
    & \textbf{1.000}$_{\pm 0.000}$ & 0.887$_{\pm 0.003}$ & \textbf{1.000}$_{\pm 0.000}$ & 0.929$_{\pm 0.006}$
    & \textbf{1.000}$_{\pm 0.000}$ & 0.882$_{\pm 0.003}$ & 0.998$_{\pm 0.001}$ & 0.911$_{\pm 0.004}$ \\
    StableSignature
    & 0.999$_{\pm 0.001}$ & 0.866$_{\pm 0.002}$ & 0.004$_{\pm 0.001}$ & 0.999$_{\pm 0.001}$
    & \textbf{1.000}$_{\pm 0.000}$ & 0.873$_{\pm 0.001}$ & 0.015$_{\pm 0.007}$ & \textbf{1.000}$_{\pm 0.000}$
    & 0.999$_{\pm 0.001}$ & 0.864$_{\pm 0.003}$ & 0.008$_{\pm 0.005}$ & 0.998$_{\pm 0.002}$ \\
    \rowcolor{rowgray}
    AquaLoRA
    & 0.996$_{\pm 0.002}$ & 0.852$_{\pm 0.003}$ & 0.906$_{\pm 0.003}$ & 0.978$_{\pm 0.004}$
    & 0.990$_{\pm 0.003}$ & 0.837$_{\pm 0.007}$ & 0.859$_{\pm 0.029}$ & 0.947$_{\pm 0.008}$
    & 0.966$_{\pm 0.007}$ & 0.820$_{\pm 0.009}$ & 0.825$_{\pm 0.024}$ & 0.908$_{\pm 0.009}$ \\
    GaussianShading
    & \textbf{1.000}$_{\pm 0.000}$ & 0.757$_{\pm 0.003}$ & \textbf{1.000}$_{\pm 0.000}$ & 0.666$_{\pm 0.003}$
    & \textbf{1.000}$_{\pm 0.000}$ & 0.757$_{\pm 0.002}$ & \textbf{1.000}$_{\pm 0.000}$ & 0.844$_{\pm 0.007}$
    & \textbf{1.000}$_{\pm 0.000}$ & 0.752$_{\pm 0.002}$ & \textbf{1.000}$_{\pm 0.000}$ & 0.833$_{\pm 0.017}$ \\
    \rowcolor{rowgray}
    FARI-GS
    & \textbf{1.000}$_{\pm 0.000}$ & 0.756$_{\pm 0.002}$ & \textbf{1.000}$_{\pm 0.000}$ & 0.770$_{\pm 0.009}$
    & \textbf{1.000}$_{\pm 0.000}$ & 0.756$_{\pm 0.001}$ & \textbf{1.000}$_{\pm 0.000}$ & 0.944$_{\pm 0.005}$
    & \textbf{1.000}$_{\pm 0.000}$ & 0.754$_{\pm 0.001}$ & \textbf{1.000}$_{\pm 0.000}$ & 0.913$_{\pm 0.009}$ \\
    PRC
    & \textbf{1.000}$_{\pm 0.000}$ & 0.641$_{\pm 0.007}$ & 0.843$_{\pm 0.005}$ & 0.533$_{\pm 0.004}$
    & \textbf{1.000}$_{\pm 0.000}$ & 0.638$_{\pm 0.004}$ & 0.804$_{\pm 0.014}$ & 0.631$_{\pm 0.014}$
    & \textbf{1.000}$_{\pm 0.000}$ & 0.637$_{\pm 0.008}$ & 0.837$_{\pm 0.006}$ & 0.662$_{\pm 0.010}$ \\
    \midrule
    % ---------- Ours ----------
    \rowcolor{rowgray}
    \textbf{\ours}
    & \textbf{1.000}$_{\pm 0.000}$ & \textbf{1.000}$_{\pm 0.000}$ & \textbf{1.000}$_{\pm 0.000}$ & \textbf{1.000}$_{\pm 0.000}$
    & \textbf{1.000}$_{\pm 0.000}$ & \textbf{1.000}$_{\pm 0.000}$ & \textbf{1.000}$_{\pm 0.000}$ & \textbf{1.000}$_{\pm 0.000}$
    & \textbf{1.000}$_{\pm 0.000}$ & \textbf{0.999}$_{\pm 0.000}$ & \textbf{1.000}$_{\pm 0.000}$ & \textbf{1.000}$_{\pm 0.000}$ \\
    \bottomrule
    \end{tabular}%
    }
\end{table*}

\begin{figure*}[t]
    \centering
    \begin{minipage}[t]{0.49\textwidth}
        \centering
        \includegraphics[width=\linewidth]{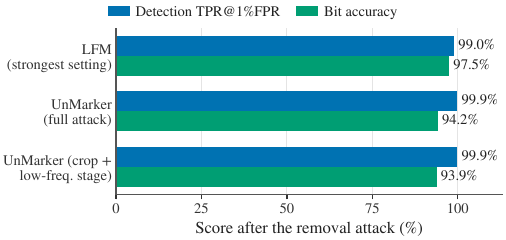}
        \caption{\textbf{Robustness to stronger removal attacks on SD~2.1 with DiffusionDB.} Detection TPR@1\%FPR and bit accuracy after LFM at its strongest tested setting and two UnMarker variants.}
        \label{fig:strong_removal}
    \end{minipage}
    \hfill
    \begin{minipage}[t]{0.49\textwidth}
        \centering
        \includegraphics[width=\linewidth]{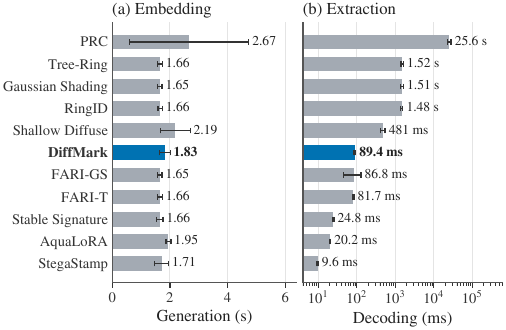}
        \caption{\textbf{Per-image watermark embedding and extraction latency on SD~2.1.}}
        \label{fig:latency}
    \end{minipage}
    \vspace{-8pt}
\end{figure*}

\textbf{Supporting results.} \cref{lem:psd} (Appendix~\cref{sec:supp_margin})  provides an idealized explanation for why a late step can reduce a perturbation without reversing it. In a Gaussian latent model, \cref{prop:gauss} (Appendix~\cref{sec:supp_proof_prop1}) shows how applying imprinting later improves detectability per unit of $\delta$ and gives a positive lower bound on short- and full-path gradient alignment. Appendix~\cref{sec:supp_measured} empirically examines these properties on SD~2.1, SD3, and FLUX.

\subsection{Descent with short-path gradients}
Fix the key $s$, condition $c$, initial latent $z_T$, decoder $D_\psi$, and all sampling noise. For $P\in\{{\rm sh},{\rm full}\}$, let $z^P_0(\delta)$ be the final latent of path $P$ (\cref{eq:shortpath,eq:injection}) and $\ell_P(\delta)=\mathcal L_{\rm rec}\big(o(z^P_0(\delta)),s\big)$ its per-sample decoding loss. With $A_P=\partial z^P_0/\partial\delta$ and $g_P=\nabla_z\mathcal L_{\rm rec}(o(z),s)\big|_{z=z^P_0}$, the chain rule gives $\nabla_\delta\ell_P=A_P^\top g_P$. Here, $A_{\rm full}=\nabla\Phi(\delta)$, with $\Phi$ taken at the training weights $\eta_k=1/N$, equals $A$ at $\delta=0$. Following inexact-gradient analysis~\cite{dmBottou2018,dmAjalloeian2020}, we compare the two gradients; the full-sampler gradient serves only as a theoretical reference and is not computed during training.

\begin{theorem}[Descent directions from short-path gradients]
\label{thm:surrogate}
For any scale $\kappa>0$, define the bias $b_\delta={}\big\|(\kappa^{-1}A_{\rm sh}-A_{\rm full})^\top g_{\rm sh}\big\| + \big\|A_{\rm full}^\top(g_{\rm sh}-g_{\rm full})\big\|.$ At any point where both losses are differentiable,
$\|\kappa^{-1}\nabla_\delta\ell_{\rm sh}-\nabla_\delta\ell_{\rm full}\|\le b_\delta$.
If $b_\delta<\|\nabla_\delta\ell_{\rm full}\|$, then
\begin{equation}
  \cos\angle\big(\nabla_\delta\ell_{\rm sh},\nabla_\delta\ell_{\rm full}\big)
  \ge\sqrt{1-\frac{b_\delta^2}{\|\nabla_\delta\ell_{\rm full}\|^2}}>0,
  \label{eq:descent}
\end{equation}
so $-\nabla_\delta\ell_{\rm sh}$ is a descent direction for the full-sampler loss $\ell_{\rm full}$.
At a point where $\delta=E_\phi(s)$ is differentiable in the encoder parameters $\phi$, $-\nabla_\phi\ell_{\rm sh}$ is a descent direction for $\ell_{\rm full}$ if
$\|\partial\delta/\partial\phi\|\,b_\delta<\|\nabla_\phi\ell_{\rm full}\|$.
\end{theorem}
The scale $\kappa$ accounts for the different imprinting strengths of the two paths. After rescaling, $b_\delta$ is small when the two paths respond similarly to $\delta$ along the decoder gradient, $\kappa^{-1}A_{\rm sh}^\top g_{\rm sh}\approx A_{\rm full}^\top g_{\rm sh}$, and have similar decoder gradients, $g_{\rm sh}\approx g_{\rm full}$.

When $g_{\rm sh}=g_{\rm full}$, a scale $\kappa$ with $b_\delta<\|\nabla_\delta\ell_{\rm full}\|$ exists if and only if the two gradients have a positive inner product (Appendix~\cref{sec:supp_proof_thm2}), so their cosine is the quantity the theorem turns on. We therefore check this hypothesis directly through the measured cosine, rather than through a bound on the smallest singular value of $A_{\rm full}$, which is small on the deployed samplers; Appendix~\cref{sec:gradient-alignment} shows that the cosine on SD~2.1 is positive with respect to both $\delta$ and $\phi$ at every tested checkpoint.

\begin{table*}[t]
    \centering
    \caption{\textbf{Image quality results with SD 2.1 on MS-COCO}. References are real images and unwatermarked companions. Values are mean with the standard deviation in subscript. The best result is in \textbf{bold}.}
    \label{tab:quality_sd21}
    \resizebox{0.85\textwidth}{!}{%
    \footnotesize
    \setlength{\tabcolsep}{4pt}
    \begin{tabular}{l cccc @{\hspace{10pt}} cccc}
    \toprule
    \multirow{2}{*}{\textbf{Method}}
    & \multicolumn{4}{c}{\textbf{MS-COCO (real images)}}
    & \multicolumn{4}{c}{\textbf{MS-COCO (unwatermarked images)}} \\
    \cmidrule(lr){2-5} \cmidrule(lr){6-9}
    & PSNR$\uparrow$ & LPIPS$\downarrow$ & FID$\downarrow$ & CLIP-FID$\downarrow$
    & PSNR$\uparrow$ & LPIPS$\downarrow$ & FID$\downarrow$ & CLIP-FID$\downarrow$ \\
    \cmidrule(lr){2-2} \cmidrule(lr){3-3} \cmidrule(lr){4-4} \cmidrule(lr){5-5}
    \cmidrule(lr){6-6} \cmidrule(lr){7-7} \cmidrule(lr){8-8} \cmidrule(lr){9-9}
    % ---------- Zero-bit ----------
    \rowcolor{rowgray}
    TreeRing
    & 8.83$_{\pm 0.04}$ & 0.769$_{\pm 0.002}$ & 118.9$_{\pm 1.1}$ & 24.61$_{\pm 0.25}$
    & 12.92$_{\pm 0.05}$ & 0.479$_{\pm 0.003}$ & 58.3$_{\pm 1.1}$ & 4.73$_{\pm 0.08}$ \\
    RingID
    & 8.57$_{\pm 0.06}$ & 0.773$_{\pm 0.002}$ & 124.1$_{\pm 1.1}$ & 24.77$_{\pm 0.30}$
    & 11.64$_{\pm 0.04}$ & 0.542$_{\pm 0.003}$ & 64.9$_{\pm 0.8}$ & 5.36$_{\pm 0.08}$ \\
    \rowcolor{rowgray}
    ShallowDiffuse
    & 8.74$_{\pm 0.05}$ & 0.778$_{\pm 0.002}$ & 117.3$_{\pm 1.3}$ & 24.22$_{\pm 0.23}$
    & \textbf{30.34}$_{\pm 0.10}$ & \textbf{0.027}$_{\pm 0.000}$ & \textbf{5.4}$_{\pm 0.1}$ & \textbf{0.20}$_{\pm 0.00}$ \\
    FARI-T
    & 8.83$_{\pm 0.04}$ & 0.769$_{\pm 0.002}$ & 118.9$_{\pm 1.1}$ & 24.61$_{\pm 0.25}$
    & 12.92$_{\pm 0.05}$ & 0.479$_{\pm 0.003}$ & 58.3$_{\pm 1.1}$ & 4.73$_{\pm 0.08}$ \\
    \midrule
    % ---------- Multi-bit ----------
    StegaStamp
    & \textbf{13.97}$_{\pm 0.15}$ & \textbf{0.445}$_{\pm 0.003}$ & \textbf{67.4}$_{\pm 1.1}$ & \textbf{9.72}$_{\pm 0.14}$
    & 27.99$_{\pm 0.06}$ & 0.099$_{\pm 0.001}$ & 40.1$_{\pm 1.1}$ & 6.65$_{\pm 0.17}$ \\
    \rowcolor{rowgray}
    StableSignature
    & 8.57$_{\pm 0.05}$ & 0.769$_{\pm 0.002}$ & 117.2$_{\pm 1.2}$ & 23.85$_{\pm 0.22}$
    & 28.01$_{\pm 0.11}$ & 0.039$_{\pm 0.001}$ & 7.0$_{\pm 0.1}$ & 0.75$_{\pm 0.02}$ \\
    AquaLoRA
    & 8.50$_{\pm 0.04}$ & 0.777$_{\pm 0.001}$ & 120.5$_{\pm 1.0}$ & 23.25$_{\pm 0.19}$
    & 18.17$_{\pm 0.09}$ & 0.253$_{\pm 0.003}$ & 44.1$_{\pm 0.8}$ & 3.19$_{\pm 0.04}$ \\
    \rowcolor{rowgray}
    GaussianShading
    & 8.58$_{\pm 0.06}$ & 0.776$_{\pm 0.001}$ & 119.1$_{\pm 0.9}$ & 24.63$_{\pm 0.16}$
    & 8.68$_{\pm 0.04}$ & 0.716$_{\pm 0.002}$ & 64.7$_{\pm 0.7}$ & 5.46$_{\pm 0.10}$ \\
    FARI-GS
    & 8.58$_{\pm 0.06}$ & 0.776$_{\pm 0.001}$ & 119.1$_{\pm 0.9}$ & 24.63$_{\pm 0.16}$
    & 8.68$_{\pm 0.04}$ & 0.716$_{\pm 0.002}$ & 64.7$_{\pm 0.7}$ & 5.46$_{\pm 0.10}$ \\
    \rowcolor{rowgray}
    PRC
    & 8.70$_{\pm 0.06}$ & 0.782$_{\pm 0.002}$ & 120.0$_{\pm 1.0}$ & 25.30$_{\pm 0.20}$
    & 8.96$_{\pm 0.03}$ & 0.730$_{\pm 0.002}$ & 64.6$_{\pm 0.6}$ & 5.45$_{\pm 0.11}$ \\
    \midrule
    % ---------- Ours ----------
    \rowcolor{rowgray}
    \textbf{\ours}
    & 8.65$_{\pm 0.05}$ & 0.778$_{\pm 0.002}$ & 118.6$_{\pm 1.4}$ & 23.66$_{\pm 0.26}$
    & 24.20$_{\pm 0.08}$ & 0.162$_{\pm 0.001}$ & 30.5$_{\pm 0.5}$ & 6.02$_{\pm 0.12}$ \\
    \bottomrule
    \end{tabular}%
    }
\end{table*}

\begin{figure*}[t]
  \centering
  \includegraphics[width=0.9\textwidth,keepaspectratio]{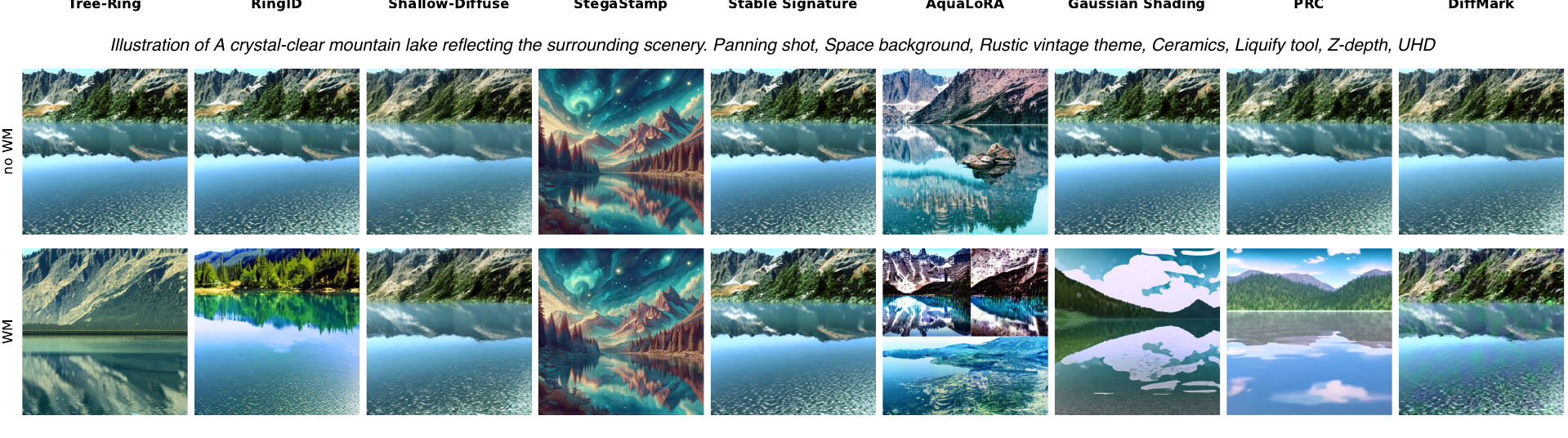}
  \caption{\textbf{Qualitative comparison on DALL-E3 dataset.} Unwatermarked (no WM) and watermarked (WM) generations for matched prompts and initial noise across methods.}
  \label{fig:qualitative}
  \vspace{-0.3cm}
\end{figure*}

% =====================================================================
\section{Experiments}
\label{sec:experiments}

\subsection{Experimental setup}
\label{sec:setup}
\textbf{Data and baselines.} We evaluate on 1,000 prompts each from DiffusionDB~\cite{wang2023diffusiondb}, MS-COCO 2017~\cite{lin2014microsoft}, and the DALL-E3 prompt set~\cite{anwaves}, following prior work~\cite{yangfari, li2025shallow, feng2024aqualora, yang2024gaussian}. The 10,000 DiffusionDB training prompts are used to train \ours's \emph{watermark encoder} $E_\phi$ and \emph{watermark decoder} $D_\psi$, while the frozen generator produces the short- and full-path latents used by the objectives in \cref{sec:dualpath}. We compare with the 10 baselines in \cref{tab:main_results}: StegaStamp~\cite{tancik2020stegastamp}, Stable Signature~\cite{fernandez2023stable}, AquaLoRA~\cite{feng2024aqualora}, Tree-Ring~\cite{wen2023tree}, RingID~\cite{ci2024ringid}, Shallow Diffuse~\cite{li2025shallow}, Gaussian Shading~\cite{yang2024gaussian}, FARI-T and FARI-GS~\cite{yangfari}, and PRC~\cite{gunn2025prc}.

\textbf{Metrics.} Following WAVES~\cite{anwaves}, we report TPR@1\%FPR for detection and bit accuracy (BACC) for message recovery. Following WAVES~\cite{anwaves}, we evaluate 8 distortion attacks (rotation, resized cropping, random erasing, brightness and contrast adjustments, Gaussian blur, Gaussian noise, and JPEG compression), 5 removal configurations (VAE regeneration, diffusion regeneration, and Rinse at settings 20, 60, and 100), and 2 adversarial attacks (Adv-KLVAE8 and Adv-RN18). In Appendix, we further evaluate \ours under 2 stronger removal attacks: LFM~\cite{aistov2026exploring} and UnMarker~\cite{kassis2025unmarker}. To measure image quality, we report PSNR, LPIPS~\cite{zhang2018unreasonable}, FID~\cite{heusel2017gans} and CLIP-FID~\cite{anwaves}.

\textbf{Implementation.} We use SD~2.1-base~\cite{rombach2022high} (DDIM, 50 steps, guidance 7.5), SD3-medium~\cite{esser2024scaling} (flow-matching Euler, 28 steps, guidance 7.0, without T5), and FLUX.1-schnell~\cite{flux2024} (4 distilled steps, no guidance), all at $512{\times}512$ and kept frozen. The imprinting window covers the last 5 steps with $\bar\eta=0.5$ on SD~2.1, the last 5 steps with $\bar\eta=2$ on SD3, and the last 2 steps with $\bar\eta=3$ on FLUX. We train the watermark encoder and decoder on SD~2.1 and adapt these modules for each target generator. Training uses $K=2$ short-path steps on SD~2.1 and SD3, and $K=4$ on FLUX. All hyperparameters are in Appendix~\cref{sec:supp_impl}.

\subsection{Robustness}
\label{sec:robustness}

\cref{tab:main_results} shows consistently high detection across datasets: \ours reaches $1.000$ TPR on clean images and at least $0.999$ average TPR across distortion, removal, and adversarial attacks in all three datasets. Across SD~2.1, SD3, and FLUX, \ours achieves $99.4$--$100\%$ average TPR across these attack groups, with per-attack results in Appendix~\cref{sec:supp_distortion_results,sec:supp_removal_adv_results}. Under stronger removal attacks on SD~2.1 with DiffusionDB (Fig.~\ref{fig:strong_removal}), \ours achieves $99.0\%$ TPR@1\%FPR and $97.5\%$ BACC under LFM~\cite{aistov2026exploring}, and $99.9\%$ TPR with $93.9$--$94.2\%$ BACC across two UnMarker variants~\cite{kassis2025unmarker}.

\subsection{Fast extraction} \ours takes $89.4$\,ms for extraction and $1.83$\,s for generation on SD~2.1 (\cref{fig:latency}), while achieving the highest robustness across attacks (\cref{tab:main_results}). Extraction latency is comparable to FARI-T/FARI-GS ($81.7$/$86.8$\,ms), with higher robustness under distortion and adversarial attacks (\cref{tab:main_results}). It is about $17\times$ faster than Gaussian Shading ($1.51$\,s), and $5.4\times$ faster than Shallow Diffuse ($481$\,ms). This efficiency comes from avoiding diffusion inversion: bits are recovered from VAE-encoded latents using decoder margins and correlations with fixed frequency-based templates $H_\ell$.

\subsection{Image quality}
\label{sec:image-quality}
\cref{tab:quality_sd21} reports image quality on MS-COCO with SD~2.1. Against real images, \ours performs similarly to other generation-time methods. Against unwatermarked images from the same prompts and initial noise, it achieves $24.20$\,dB PSNR, $0.162$ LPIPS, and $30.5$ FID, outperforming AquaLoRA and initial-noise methods other than Shallow Diffuse on these metrics. These initial-noise methods alter image content by changing $z_T$. Stable Signature and Shallow Diffuse achieve higher PSNR ($28.0$ and $30.3$\,dB), but lose the watermark under removal and distortion attacks, respectively (\cref{tab:main_results}). However, \ours has the highest CLIP-FID ($6.02$), indicating changes in CLIP features. Qualitative results (\cref{fig:qualitative}) and residual maps (Appendix~\cref{sec:supp_residuals}) show changes in fine texture and local color while preserving scene layout. \cref{tab:ablation} and Appendix~\cref{sec:supp-payload} examine the trade-offs among image quality, robustness, and payload length.

\subsection{Analysis and Ablation}
\label{sec:analysis-ablation}

\begin{table}[ht]
    \centering
    \caption{\textbf{Ablation of \ours components}. The best result in each column is in \textbf{bold}.}
    \label{tab:ablation}
    \resizebox{0.9\columnwidth}{!}{%
    \footnotesize
    \setlength{\tabcolsep}{3pt}
    \begin{tabular}{l ccccc}
    \toprule
    \textbf{Component}
    & \makecell{\textbf{Clean}\\\textbf{BACC}} & \makecell{\textbf{Avg}\\\textbf{TPR@1\%FPR}} & \makecell{\textbf{Avg}\\\textbf{BACC}} & \textbf{PSNR} & \textbf{LPIPS} \\
    \cmidrule(lr){2-2} \cmidrule(lr){3-3} \cmidrule(lr){4-4} \cmidrule(lr){5-5} \cmidrule(lr){6-6}
    \rowcolor{rowgray}
    \textbf{\ours (full)} & \textbf{1.000} & \textbf{1.000} & 0.966 & \textbf{26.15} & \textbf{0.204} \\
    \midrule
    w/o imprinting window & 1.000 & 1.000 & \textbf{0.968} & 22.26 & 0.381 \\
    \rowcolor{rowgray}
    w/o pre-filter & 0.999 & 0.990 & 0.950 & 26.15 & 0.204 \\
    \makecell[l]{w/o frequency-based\\embedding} & 0.988 & 0.933 & 0.920 & 26.15 & 0.204 \\
    \rowcolor{rowgray}
    w/o short-path & 0.919 & 0.891 & 0.874 & 26.15 & 0.204 \\
    w/o correction & 0.881 & 0.871 & 0.885 & 26.15 & 0.204 \\
    \bottomrule
    \end{tabular}%
    }
    \vspace{-8pt}
\end{table}

\textbf{Component contributions.} In \cref{tab:ablation}, removing frequency-based embedding, short-path training, or embedding correction lowers average TPR from $1.000$ to $0.933$, $0.891$, and $0.871$, respectively. Correction also has the largest effect on clean BACC ($1.000$ to $0.881$), supporting its role in repairing weak bit margins. The imprinting window improves image quality while preserving perfect average TPR. Compared with imprinting throughout sampling, the late-window configuration raises PSNR by $3.9$\,dB and nearly halves LPIPS ($0.381$ to $0.204$).

\begin{figure}[ht]
    \centering
    \includegraphics[width=\linewidth]{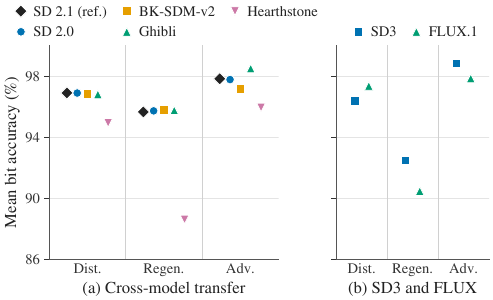}
    \caption{\textbf{Robustness across generators on DiffusionDB.} Mean BACC by attack group for (a) compatible checkpoints using the SD~2.1 watermark modules and (b) SD3 and FLUX after adapting the watermark encoder and decoder.}
    \label{fig:cross-model}
\end{figure}

\textbf{Robustness across generators.} Applying the SD~2.1 modules to compatible SD~2.0~\cite{stabilityai2022sd2}, BK-SDM-v2~\cite{kim2024bksdm}, Ghibli~\cite{nitrosocke_ghibli}, and Hearthstone~\cite{norod78_hearthstone} checkpoints achieves $\sim100\%$ TPR@1\%FPR under distortion and adversarial attacks (\cref{fig:cross-model}a). These checkpoints use latent representations compatible with the trained modules, allowing direct reuse without retraining. SD3 and FLUX use different VAE latent spaces and sampling dynamics, requiring adaptation of $E_\phi$ and $D_\psi$ to their latent dimensions and watermark responses. After adaptation, \ours achieves roughly $96$--$99\%$ BACC under distortion and adversarial attacks and $90$--$92\%$ under regeneration (\cref{fig:cross-model}b), while keeping generator weights unchanged.

\begin{figure}[ht]
    \centering
    \includegraphics[width=\linewidth,keepaspectratio]{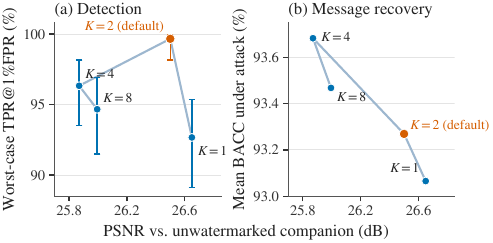}
    \caption{\textbf{Short-path training steps $K$}. PSNR versus (a) worst-case TPR@1\%FPR and (b) mean BACC over all attacks on DiffusionDB.}
    \label{fig:lcmstep}
    \vspace{-0.3cm}
\end{figure}

\textbf{Short-path steps ablation.} We vary the number $K$ of short-path steps of the frozen sampler on SD~2.1 (\cref{fig:lcmstep}). With $K=2$, worst-case detection is near-perfect at about $26.5$\,dB PSNR. Increasing $K$ from 1 to 4 improves mean BACC by only $0.6$ percentage points while reducing PSNR by $0.8$\,dB. These results support $K=2$ as a balance between image quality and robustness, and show that mean BACC alone can conceal vulnerability to individual attacks.

Additional experiments in the appendix examine watermark propagation and sampler Jacobians on SD~2.1, SD3, and FLUX (\cref{sec:supp_measured}), short- and full-path gradient alignment (\cref{sec:gradient-alignment}), the trade-off among payload length, image quality, and robustness (\cref{sec:supp-payload}), the imprinting window (\cref{sec:injection-window}), key flexibility (\cref{sec:key-flexibility}) and robustness to sampling and guidance settings (\cref{sec:robust-sample-guidance}).

% =====================================================================
\section{Conclusion}
\label{sec:conclusion}
\ours combines fast image-only extraction with robust recovery of arbitrary 256-bit keys. Learned late-trajectory \rev{imprinting}, fixed frequency-based embedding, and margin correction support recovery under image manipulation, while direct latent decoding avoids diffusion inversion. Dual-path training learns the embedding through a short sampling path and calibrates the decoder on full-sampler outputs. Across three datasets with SD~2.1, SD3, and FLUX, \ours achieves $99.4$--$100\%$ average TPR@1\%FPR across attack groups. On SD~2.1, extraction takes $89.4$\,ms, approximately $17\times$ faster than Tree-Ring and Gaussian Shading.

\newpage
{
  \small
  \bibliographystyle{ieeenat_fullname}
  \bibliography{references}
}

% WARNING: the supplementary pages are appended after the references so
% that \cref works across files. Remove them from the main-paper PDF and
% submit them as the separate supplementary file.

% Readability revisions relative to the supplied appendix are marked in dark green.
% This color is local to revised passages; the main paper needs no changes.
\definecolor{diffmarkreadabilitygreen}{rgb}{0,0.392157,0}
\DeclareRobustCommand{\diffmarkreadability}[1]{{\hypersetup{allcolors=diffmarkreadabilitygreen}\color{diffmarkreadabilitygreen}#1}}
% Proof revisions are marked in dark red (definition copied from supp_revised.tex).
\definecolor{diffmarkproofred}{RGB}{139,0,0}
\DeclareRobustCommand{\diffmarkproofrevision}[1]{{\hypersetup{allcolors=diffmarkproofred}\color{diffmarkproofred}#1}}

% =====================================================================

\clearpage
\maketitlesupplementary

\setcounter{section}{0}
\renewcommand{\thesection}{\Alph{section}}

This supplement provides method details (\cref{sec:additional-diffmark}), notation and proofs (\cref{sec:supp_proofs}), implementation details (\cref{sec:supp_impl}), attack settings (\cref{sec:supp_attacks}), additional results (\cref{sec:supp_results}), extended related work (\cref{sec:supp_related}), and limitations (\cref{sec:supp_limitations}).

% =====================================================================

\section{Additional Details of \ours}
\label{sec:additional-diffmark}
This section supplements the method in \cref{sec:method} with network architectures, pattern design, and algorithms for embedding, extraction, and training.

\subsection{Watermark encoder and decoder}
\label{sec:supp_architecture}
The watermark networks $E_\phi$ and $D_\psi$ are separate from the frozen VAE $(\mathcal E,\mathcal D)$. The encoder takes only the key, while the decoder takes only an image latent and has no timestep input.

\textbf{Encoder.} For each bit position $\ell$ and value $v\in\{0,1\}$, a learned vector $b_{\ell,v}\in\R^{256}$ represents that pair. The key embedding is $e(s)=\sqrt{64/L}\sum_{\ell=1}^L b_{\ell,s_\ell}$, where the factor controls its scale across key lengths. A learned spatial basis $B\in\R^{256\times64\times64}$ maps this vector to spatial features $X_{j,y,x}=e_j(s)B_{j,y,x}$. Three $3\times3$ convolutional blocks map channels $256\to128\to64\to32$, using SiLU activations and BatchNorm in the first two blocks. Two $3\times3$ heads map the resulting features to $C$-channel fields $\mu$ and $\log\sigma^2$. With a learned scalar $a$, the output is $\delta=a\mu$ in deterministic mode or $\delta=a(\mu+\sigma\odot\epsilon)$ with $\epsilon\sim\mathcal N(0,\mathbf I)$ in stochastic pretraining.

\textbf{Decoder.} A $3\times3$ convolution maps the $C$ latent channels to 8 features, followed by SiLU. Five $4\times4$ convolutions with stride 2 then map channels $8\to16\to32\to64\to128\to256$, reducing a $64\times64$ latent to $2\times2$. Each stage uses SiLU; all but the final stage also use BatchNorm. Adaptive average pooling fixes the output grid at $2\times2$. An MLP with widths $1024\to2048\to1024\to2L$ and SiLU hidden activations produces two logits per bit. The difference between the bit-1 and bit-0 logits gives $o^{\rm dec}_\ell$ in \cref{sec:extraction}. The fixed frequency-based score $o^{\rm sc}_\ell$ is added separately to form the final margin.

\begin{figure*}[ht]
    \centering
    \includegraphics[width=\linewidth]{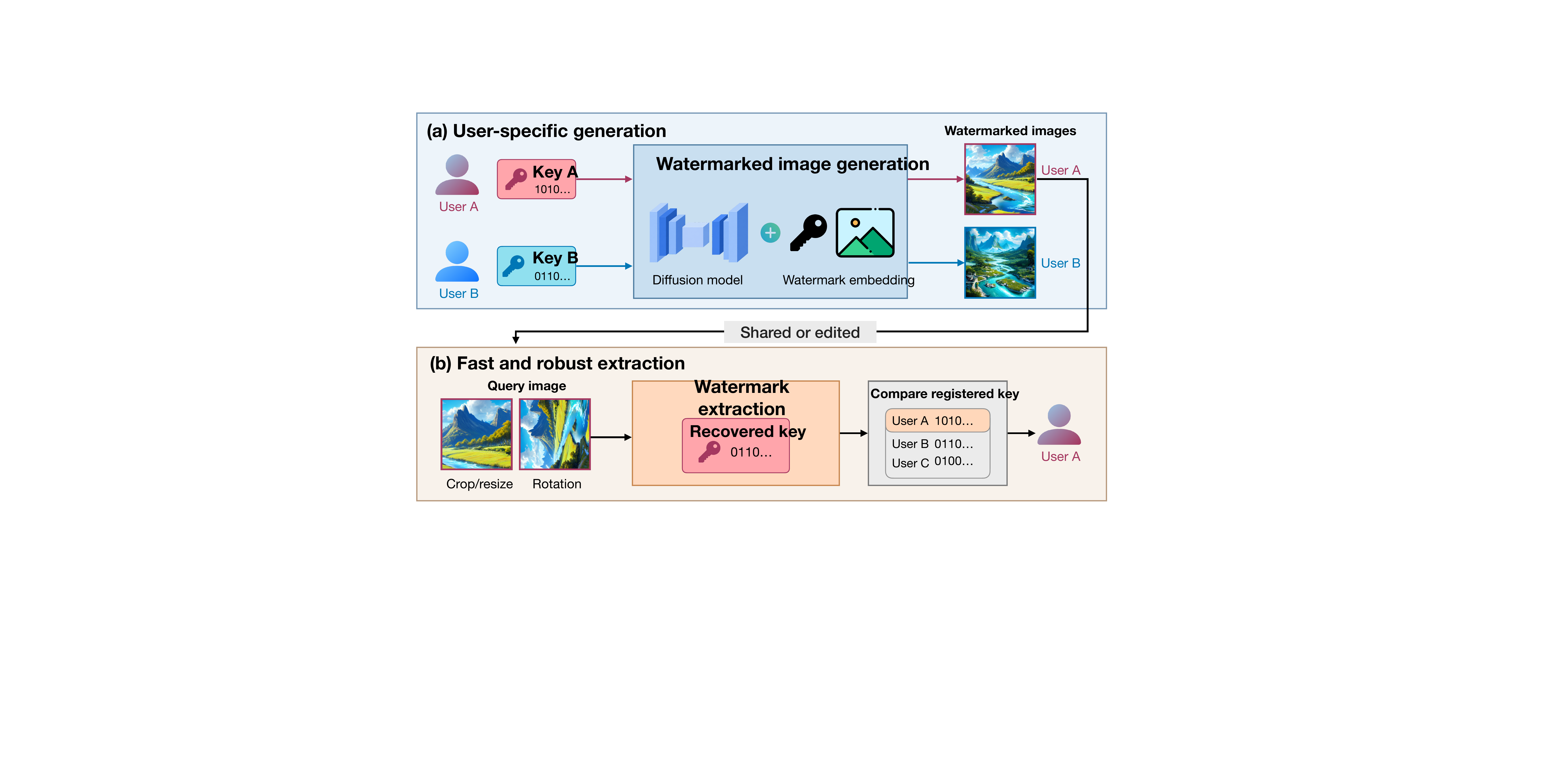}
    \caption{\textbf{Design of frequency-based patterns.} Gaussian fields drawn from a shared seed are band-pass filtered, jointly orthonormalized, and mapped to latent channels through $V$ to obtain embedding patterns $P_\ell$. Calibration on generated images determines $V$ and $\Lambda$ from latent covariance, pattern responses, and image distortion. The scoring templates $H_\ell=\Lambda P_\ell$ provide correlation scores for extraction. The embedding patterns and scoring templates are computed once per generator and reused across user keys.}
    \label{fig:frequency-based-patterns}
\end{figure*}

\subsection{Design of frequency-based patterns}
\label{sec:supp_subcarrier}

\textbf{Goal.} Frequency-based patterns complement the learned trajectory perturbation by embedding the same bits at coarser spatial scales, improving robustness to blur and small alignment errors. Their correlation scores provide complementary evidence for bit recovery and, through agreement with the learned decoder, guide key-free alignment correction. We use measured pattern responses and image distortion to balance recovery strength and image quality.

We draw $L$ Gaussian fields with $n_c$ channels and restrict them to a frequency band satisfying $1/p_{\rm hi}\le\|f\|_2\le1/p_{\rm lo}$, where $f$ is the spatial frequency in cycles per latent pixel. The bounds $p_{\rm lo}$ and $p_{\rm hi}$ are the minimum and maximum retained spatial periods, measured in latent pixels. We filter the fields in the Fourier domain, transform them back to spatial coordinates, and apply QR orthonormalization to their flattened representations, obtaining $\widetilde P_\ell$. An orthonormal matrix $V\in\R^{C\times n_c}$ maps the $n_c$ pattern channels to the $C$ latent channels, giving $P_\ell(:,y,x)=V\widetilde P_\ell(:,y,x)$. Since $V^\top V=\mathbf I_{n_c}$, the mapped patterns satisfy $\langle P_\ell,P_j\rangle=\mathbf1[\ell=j]$. The frequency band must support at least $L$ independent patterns. Therefore, the added term in \cref{eq:subcarrier} has squared norm $\beta_{\rm sc}^2Chw$, making $\beta_{\rm sc}$ its root-mean-square amplitude over the latent coordinates.

\textbf{Statistics estimated from generated images.}
We choose $V$ and the scoring matrix $\Lambda$ using a small set of images generated with the frozen model. These images let us estimate natural latent variation, measure the image distortion caused by watermark patterns, and assess how well the patterns are recovered during extraction. The matrix $S\in\R^{C\times C}$ is the covariance across latent channels within the selected frequency band, measured after applying the extraction pre-filter and VAE encoding. The response matrix $G\in\R^{C\times C}$ describes how an embedded channel direction appears after VAE decoding, pre-filtering, and re-encoding. The matrix $D_{\rm img}\in\R^{C\times C}$ captures image distortion per unit of embedding along different channel directions. We also measure $G_{\rm clean}$ using the same VAE decoding and re-encoding procedure without the pre-filter.

\textbf{Choosing embedding directions.}
For a channel direction $v\in\R^C$, we balance the extraction response against natural latent variation and image distortion through
\begin{equation}
  \ensuremath{
  J(v)=\frac{
    \underbrace{v^\top G^\top S^{-1}Gv}_{\text{covariance-weighted pattern response}}
  }{
    \underbrace{v^\top D_{\rm img}v}_{\text{image distortion}}
  }.
  }
  \label{eq:supp_direction_design}
\end{equation}
The weighting $S^{-1}$ reduces the contribution of directions with large natural variation, where image content could obscure the watermark. The denominator penalizes directions that cause larger image changes. We select the leading $n_c$ generalized eigenvectors of $G^\top S^{-1}G$ relative to $D_{\rm img}$ and orthonormalize them to form $V$. Its columns span channel combinations with strong extraction responses relative to their image distortion, while preserving the orthonormality of the spatial patterns.

\textbf{Choosing scoring weights.}
We construct extraction templates that account for latent covariance and normalize the measured response of the selected embedding directions:
\begin{equation}
  \ensuremath{
  \begin{aligned}
    W&=S^{-1}G_{\rm clean}V
       \big(V^\top G_{\rm clean}^\top S^{-1}G_{\rm clean}V\big)^{-1},\\
    \Lambda&=WV^\top,\qquad H_\ell=\Lambda P_\ell.
  \end{aligned}
  }
  \label{eq:supp_pattern_design}
\end{equation}
Here, $S^{-1}$ downweights latent directions dominated by natural image variation, while the inverse matrix gives $W^\top G_{\rm clean}V=\mathbf I_{n_c}$ in the measured linear-response model. This normalizes the reference response and separates the contributions of the selected channel directions. The resulting template $H_\ell$ weights latent channels when computing $o^{\rm sc}_\ell(z)=\langle z,H_\ell\rangle$. We compute the statistics and design matrices once per generator, with its weights fixed, and reuse the resulting $P_\ell$ and $H_\ell$ for all user keys. The settings are provided in \cref{sec:supp_inference_settings}.

\subsection{Correction and alignment}
\label{sec:supp_correction}
\textbf{Correction during embedding.} Starting from $z^{(0)}=z_0^+$, we decode, quantize, pre-filter, and re-encode each candidate before computing its combined margins $o^{(t)}$. Let $\mathcal W_t=\{\ell:\bar s_\ell o^{(t)}_\ell<\tau_c\}$ contain the weak bits and let $r_\ell=\langle P_\ell,H_\ell\rangle>0$ be the direct frequency-score response. The update is
\begin{equation}
  z^{(t+1)}=z^{(t)}+\gamma_c\sum_{\ell\in\mathcal W_t}
  \frac{\tau_c-\bar s_\ell o^{(t)}_\ell}{r_\ell}\,\bar s_\ell P_\ell.
  \label{eq:supp_correction}
\end{equation}
All candidates are evaluated at the original alignment. We apply at most $R$ correction updates. If every bit meets the target margin ($\mathcal W_t=\varnothing$), no further updates are needed. We then select the latent state with the fewest bit errors from the initial state and all updated states. The direct frequency-score response only approximates the full extraction response. We therefore retain the best candidate in case an update increases bit errors. This procedure uses the known key and requires no gradient optimization.

\textbf{Alignment during extraction.} For SD~2.1, we first search for rotation and crop corrections on the latent to reduce computation. We apply the candidates with the highest agreement scores to the image and re-encode each result with the VAE. We then refine the best correction. This image-based check is needed because the latent search only approximates transforming and re-encoding the image. For SD3 and FLUX, we perform the entire search by transforming and re-encoding the image. We accept a correction only if its agreement score is high relative to other candidates and its decoder margins are strong. Otherwise, we keep the original alignment. A separate threshold verifies the recovered key.

\subsection{Training objectives and stages}
\label{sec:supp_training}
We first pretrain the watermark encoder $E_\phi$ and decoder $D_\psi$ directly on perturbations $\delta=E_\phi(s)$. We then fine-tune with the frozen short and full sampling paths, $\{F_k^{\rm sh}\}_{k=1}^{K}$ and $\{F_k\}_{k=1}^{N}$, together with the image-quality losses and under attacks. This first teaches $D_\psi$ to recover the key, before introducing losses that reduce the amplitude of $\delta$.

\textbf{Loss definitions.}
For a minibatch of $B$ keys, write $\delta_b=E_\phi(s_b)$ and let $n_z = Chw$ be the number of latent coordinates. The encoder heads define the auxiliary distribution $q_\phi(\cdot\mid s_b)=\mathcal N(\mu_b,\operatorname{diag}(\sigma_b^2))$.

\textit{Diversity.} We penalize the mean cosine similarity between different perturbations:
\begin{equation}
  \ensuremath{
    \mathcal L_{\rm div}
    =\frac{2}{B(B-1)}
      \sum_{b<b'}
      \frac{\langle\delta_b,\delta_{b'}\rangle}
           {\|\delta_b\|_2\|\delta_{b'}\|_2}.
  }
  \label{eq:supp_diversity_loss}
\end{equation}
This discourages different keys from producing perturbations with similar directions. 

\textit{Gaussian regularization.} The KL loss compares the auxiliary encoder distribution with a standard Gaussian:
\begin{equation}
  \ensuremath{
    \mathcal L_{\rm KL}
    =\frac{1}{2B}\sum_{b=1}^{B}\sum_{j=1}^{n_z}
      \left(\mu_{b,j}^2+\sigma_{b,j}^2
      -\log\sigma_{b,j}^2-1\right).
  }
  \label{eq:supp_kl_loss}
\end{equation}
This term regularizes both encoder heads, even when the forward pass uses the deterministic mean.

\textit{Perturbation magnitude.} Let $\xi=\operatorname{std}(\delta)-\sigma^\star$, where $\sigma^\star$ is the target amplitude. Stage~3 uses $\mathcal L_{\rm mag}=\xi^2$ while stage~4 uses the asymmetric penalty
\begin{equation}
  \ensuremath{
    \mathcal L_{\rm mag}
    =10[\xi]_+^2+[\xi]_+^4+5[-\xi]_+^2,
  }
  \label{eq:supp_magnitude_loss}
\end{equation}
where $[a]_+=\max(a,0)$. The terms penalize amplitudes below and above the target, discouraging perturbations that vanish or become too large. Stage~1 instead uses the upper-bound penalty $[\operatorname{std}(\delta)-\sigma_{\rm pre}]_+^2$, with $\sigma_{\rm pre}$ specified in \cref{sec:implement_training}.

\textit{Latent fidelity.} Let $z_{0,b}^{\rm sh,\varnothing}$ be the clean short-path output generated from the same prompt and initial noise as $z_{0,b}^{\rm sh}$. We minimize their mean squared difference:
\begin{equation}
  \ensuremath{
    \mathcal L_{\rm ld}
    = \frac{1}{Bn_z}\sum_{b=1}^{B}
      \left\|z_{0,b}^{\rm sh} - \sg(z_{0,b}^{\rm sh,\varnothing})\right\|_2^2.
  }
  \label{eq:supp_latent_loss}
\end{equation}

\textit{Regional pixel fidelity.}
Define $x_b=\mathcal D(z_{0,b}^{\rm sh})$ and its clean companion $x_b^\varnothing=\sg(\mathcal D(z_{0,b}^{\rm sh,\varnothing}))$. For a set $\mathcal W_{\rm img}$ of spatial windows, we penalize the largest regional mean absolute difference:
\begin{equation}
  \ensuremath{
    \mathcal L_{\rm prvl}
    =\frac{1}{B}\sum_{b=1}^{B}
      \max_{\Omega\in\mathcal W_{\rm img}}
      \frac{\|(x_b-x_b^\varnothing)|_\Omega\|_1}
           {3|\Omega|}.
  }
  \label{eq:supp_pixel_loss}
\end{equation}
Here, $|\Omega|$ counts spatial positions, and the restriction includes all three image channels. This limits concentrated image changes that a global average could obscure. Window size and image normalization are specified in \cref{sec:implement_training}.

\textit{False-positive suppression.} Let $z_{0,b}^{\rm full,\varnothing}$ be an unwatermarked full-path latent generated from fresh noise, and let $p_{b,\ell}=\operatorname{sigmoid}(o_\ell(z_{0,b}^{\rm full,\varnothing}))$. We minimize negative mean bit entropy:
\begin{equation}
      \mathcal L_{\rm neg}
      =\frac{1}{BL}\sum_{b=1}^{B}\sum_{\ell=1}^{L}
      \Big[p_{b,\ell}\log p_{b,\ell}
      +(1-p_{b,\ell})\log(1-p_{b,\ell})\Big].
  \label{eq:supp_negative_loss}
\end{equation}
Minimization discourages confident bit predictions on unwatermarked images.

\textit{Robustness supervision.}
For $P\in\{\mathrm{sh},\mathrm{full}\}$, an attacked latent is $z_{0,b}^{P,\rm atk}=\mathcal E(Q_8(\mathcal A(\mathcal D(z_{0,b}^{P}))))$, where $\mathcal A$ is an image attack and $Q_8$ rounds the image to eight bits. During Stage~4, we add $w_{\rm atk}(t)\mathcal L_{\rm atk}$ to \cref{eq:training_objective}, with
\begin{equation}
  \ensuremath{
    \mathcal L_{\rm atk}
    = \frac{1}{B}\sum_{b=1}^{B}
      \sum_{P\in\{\mathrm{sh},\mathrm{full}\}}
      \mathcal L_{\rm rec}
      \big(o(z_{0,b}^{P,\rm atk}),s_b\big).
  }
  \label{eq:supp_attack_loss}
\end{equation}
The round-trip term similarly applies $\mathcal L_{\rm rec}$ to $z_{0,b}^{\rm full,rt}=\mathcal E(\mathcal D(z_{0,b}^{\rm full}))$. Short-path reconstruction terms update both $\phi$ and $\psi$. Full-path latents are detached from $E_\phi$, so full-path, round-trip, and negative-entropy terms update only $\psi$. Attack sampling, quantization gradients, loss weights, and activation schedules are specified in \cref{sec:implement_training}.

% \diffmarkreadability{\textbf{Geometric augmentation.} During robustness fine-tuning, geometric attacks are followed by an approximate inverse transform. The remaining alignment error trains the decoder to handle imperfect alignment correction. The loss weights, gradient handling, and training schedule are specified in \cref{sec:implement_training}.}

\textbf{Fine-tuning for each generator backbone.} SD~2.1, SD3, and FLUX differ in latent dimensions, sampling updates $F_k$, and VAE mappings $(\mathcal E,\mathcal D)$. At $512\times512$ image resolution, SD~2.1 uses $4\times64\times64$ latents, while SD3 and FLUX use $16\times64\times64$ latents. These differences affect the output shape of $E_\phi$, the input shape of $D_\psi$, and how $\delta$ propagates into the extracted latent. Therefore, we adapt both watermark networks for each backbone. The generator remains frozen, and the adapted networks are reused for all user keys.

% =====================================================================
\subsection{Algorithms}
\label{sec:supp_algorithms}
Algorithms~\ref{alg:supp_generation}--\ref{alg:supp_training} summarize generation, extraction, and dual-path training. They use $u(s)=\beta_{\rm sc}\sqrt{Chw/L}\sum_\ell\bar s_\ell P_\ell$ for frequency-based embedding. We write $\operatorname{Margins}(z)=(o^{\rm dec}(z),o^{\rm sc}(z))$ and $\operatorname{ImageMargins}(x,T)=\operatorname{Margins}(\mathcal E(G_{\sigma_f}*T(x)))$, where the image is first resized to the input resolution. Here, $T$ is an inverse alignment transform and $I$ is the identity. The quantizers $Q_{\rm gen}$ and $Q_{\rm train}$ follow the generation and training conventions in \cref{sec:supp_inference_settings}.

\begin{algorithm}[tbp]
    \caption{\ours generation with embedding correction}
    \label{alg:supp_generation}
    \small
    \begin{algorithmic}[1]
        \Require Prompt $c$, key $s$, initial noise $z_T$; late window $\mathcal I$, total weight $\bar\eta$, correction limit $R$
        \Ensure Watermarked image $x$
        \State $\delta\gets E_\phi(s)$ in deterministic mode; $z\gets z_T$
        \For{$k=1,\ldots,N$}
          \State $\eta_k\gets(\bar\eta/|\mathcal I|)\mathbf1[k\in\mathcal I]$
          \State $z\gets F_k(z+\eta_k\delta;c)$
        \EndFor
        \State $z\gets z+u(s)$
        \State $z^*\gets z$; $e^*\gets+\infty$
        \State $r_\ell\gets\langle P_\ell,H_\ell\rangle>0$ for every bit $\ell$
        \For{$t=0,\ldots,R$}
          \State $x_t\gets Q_{\rm gen}(\mathcal D(z))$
          \State $(o^{\rm dec},o^{\rm sc})\gets\operatorname{ImageMargins}(x_t,I)$
          \State $o\gets o^{\rm dec}+o^{\rm sc}$; $\hat s_\ell\gets\mathbf1[o_\ell>0]$
          \State $e\gets\sum_\ell\mathbf1[\hat s_\ell\ne s_\ell]$
          \If{$e\le e^*$}
            \State $z^*\gets z$; $e^*\gets e$
          \EndIf
          \State $\mathcal W\gets\{\ell:\bar s_\ell o_\ell<\tau_c\}$
          \If{$\mathcal W=\varnothing$ or $t=R$}
            \State \textbf{break}
          \EndIf
          \State $z\gets z+\gamma_c\sum_{\ell\in\mathcal W}
            \dfrac{\tau_c-\bar s_\ell o_\ell}{r_\ell}\bar s_\ell P_\ell$
        \EndFor
        \State \Return $Q_{\rm gen}(\mathcal D(z^*))$
    \end{algorithmic}
\end{algorithm}

\begin{algorithm}[tbp]
\caption{\ours extraction and verification}
\label{alg:supp_extraction}
\small
\begin{algorithmic}[1]
\Require Image $x$, inverse-transform grid $\mathcal T$ containing $I$, registered keys $\mathcal S$, threshold $\tau$
\Ensure Recovered bits $\hat s$ and an accepted key, or no match
\State Resize $x$ to the input resolution; $\hat T\gets I$
\If{SD~2.1}
  \State $z\gets\mathcal E(G_{\sigma_f}*x)$
  \For{$T\in\mathcal T$}
    \State $\operatorname{Agr}_T\gets\operatorname{Agr}(\operatorname{Warp}(z,T))$ 
  \EndFor
  \State Search is needed if $\max_T \operatorname{Agr}_T>\operatorname{Agr}_I$
\Else \Comment{SD3 and FLUX}
  \State $\operatorname{Agr}_I\gets\operatorname{Agr}(\mathcal E(G_{\sigma_f}*x))$
  \State Search is needed if identity fails its early-accept test
  \If{search is needed}
    \For{$T\in\mathcal T\setminus\{I\}$}
      \State $\operatorname{Agr}_T\gets\operatorname{Agr}(\mathcal E(G_{\sigma_f}*T(x)))$
    \EndFor
  \EndIf
\EndIf
\If{search is needed}
  \State Compute median $\operatorname{med}_{\rm Agr}$ and scaled MAD $\operatorname{mad}_{\rm Agr}$ of $\{\operatorname{Agr}_T\}_{T\in\mathcal T}$
  \State Select distinct high-agreement candidates $\mathcal C$
  \State Evaluate $\mathcal C$ after image transformation and encoding
  \State Refine the best candidate to obtain $T^*$
  \State $(o^{\rm dec},o^{\rm sc})\gets\operatorname{ImageMargins}(x,T^*)$
  \If{agreement and confidence pass the acceptance checks}
    \State $\hat T\gets T^*$
  \EndIf
\EndIf
\State $(o^{\rm dec},o^{\rm sc})\gets\operatorname{ImageMargins}(x,\hat T)$
\State $\hat s_\ell\gets\mathbf1[o^{\rm dec}_\ell+o^{\rm sc}_\ell>0]$ for every bit $\ell$
\State $\text{Match}_s\gets\sum_\ell\mathbf1[\hat s_\ell=s_\ell]$ for each $s\in\mathcal S$
\State $\mathcal A\gets\{s\in\mathcal S:\text{Match}_s\ge\tau\}$
\If{$\mathcal A=\varnothing$}
  \State \Return $(\hat s,\text{no match})$
\EndIf
\State \Return $(\hat s,\arg\max_{s\in\mathcal A}\text{Match}_s)$
\end{algorithmic}
\end{algorithm}

Candidate selection, refinement, and acceptance checks in Algorithm~\ref{alg:supp_extraction} use the backbone-specific rules in \cref{sec:supp_alignment_settings}. The search does not access registered keys. 
% The threshold $\tau$ is calibrated by running the same extraction procedure on unwatermarked images, as specified in \cref{sec:supp_eval}.

For training, $\operatorname{Sample}_P(z,d;c)$ denotes path $P\in\{\mathrm{sh},\mathrm{full}\}$ with initial noise $z$, prompt $c$, and perturbation $d$ imprinted at every step using \cref{tab:supp_hparams}. We abbreviate $\ell(z,s)=\mathcal L_{\rm rec}(o(z),s)$ and let $\mathcal R_t$ collect the magnitude, KL, diversity, regional pixel, and latent penalties with their stage-dependent weights, including the additional Fourier penalty in Stage~3. The clean short-path output is the reference for the image-quality losses.

\begin{algorithm}[ht]
    \caption{One dual-path training step of \ours}
    \label{alg:supp_training}
    \small
    \begin{algorithmic}[1]
    \Require Batch $(c,s,z_T)$, independent noise $z'_T$, stage $j\in\{3,4\}$, training step $t$
    \Ensure Updated watermark parameters $\phi,\psi$
    \State $\delta\gets E_\phi(s)$ in deterministic mode
    \State $u_j\gets u(s)$ if $j=4$, otherwise $0$
    \State $z_{\rm sh}\gets\operatorname{Sample}_{\rm sh}(z_T,\delta;c)+u_j$
    \State $z_{\rm sh}^{\varnothing}\gets\sg(\operatorname{Sample}_{\rm sh}(z_T,0;c))$
    \If{FLUX}
      \State $z_{\rm full}\gets\sg(z_{\rm sh})$
      \State $z_{\rm neg}\gets\sg(\operatorname{Sample}_{\rm sh}(z'_T,0;c))$
    \Else
      \State $z_{\rm full}\gets\sg(\operatorname{Sample}_{\rm full}(z_T,\delta;c))+u_j$
      \State $z_{\rm neg}\gets\sg(\operatorname{Sample}_{\rm full}(z'_T,0;c))$
    \EndIf
    \State $\mathcal L\gets\ell(z_{\rm sh},s)+\ell(z_{\rm full},s)+\mathcal R_t$
    \State $\mathcal L\gets\mathcal L+w_{\rm neg}(t)\mathcal L_{\rm neg}(z_{\rm neg})$
    \If{round-trip supervision is sampled at step $t$}
      \State $z_{\rm rt}\gets\mathcal E(\mathcal D(z_{\rm full}))$
      \State $\mathcal L\gets\mathcal L+\ell(z_{\rm rt},s)$
    \EndIf
    \If{$j=4$}
      \State Draw a training attack $\mathcal A_t$
      \For{$P\in\{\mathrm{sh},\mathrm{full}\}$}
        \State $x_P\gets\mathcal D(z_P)$
        \State $z_P^{\rm atk}\gets\mathcal E(Q_{\rm train}(\mathcal A_t(x_P)))$
        \State $\mathcal L\gets\mathcal L+w_{\rm atk}(t)\ell(z_P^{\rm atk},s)$
      \EndFor
    \EndIf
    \State Update $(\phi,\psi)$ with $\nabla_{\phi,\psi}\mathcal L$
    \State Update per-bit accuracy estimates and loss weights
    \end{algorithmic}
\end{algorithm}

Algorithm~\ref{alg:supp_training} follows perturbation pretraining in Stages~1 and 2. All generator weights remain frozen. The short-path computation remains differentiable, whereas full-path sampling is evaluated without a gradient graph. Thus, full-path terms update only the watermark decoder. Stage~4 uses provisional frequency-based patterns and scoring templates. After selecting a checkpoint, we replace them with the fixed design in \cref{sec:supp_subcarrier}, without further training. Training schedules, attack weights $w_{\rm atk}(t)$, and quantization gradients are given in \cref{sec:implement_training}.

% \subsection{Why staged training is needed}
% \label{sec:supp_collapse}
% At initialization, the encoder gradient is dominated by two terms,
% \begin{equation}
%   \nabla_\phi\mathcal L\supset\underbrace{\nabla_\phi\mathcal L_{\rm sh}}_{\text{needs }\|\delta\|>0}+\,w_{\rm ld}\underbrace{\nabla_\phi\mathcal L_{\rm ld}}_{\text{drives }\delta\to0}.
% \end{equation}
% The reconstruction gradient flows through $\mathcal L_{\rm sh}\to D_\psi\to z_0^{\rm sh}\to\text{sampler}\to\delta\to E_\phi$, but a randomly initialized decoder predicts nearly uniform bits regardless of $z_0$, so $\nabla_{z_0}\mathcal L_{\rm sh}\approx0$. The footprint gradient, in contrast, is informative from the first step, since any $\delta\neq0$ moves $z_0^{\rm sh}$; $\mathcal L_{\rm prvl}$ behaves alike. This imbalance collapses $\|\delta\|$ to the trivial solution $\delta=0$ before the decoder can exploit the signal. Stages 1 and 2 train the decoder to read $\delta$ before stage~3 adds the imperceptibility terms, which yields an informative $\nabla_{z_0}\mathcal L_{\rm sh}$ that counterbalances them. In the language of \cref{thm:surrogate}, the decoder-loss gradient $g_{\rm sh}$, and hence $\nabla_\delta\ell_{\rm sh}=A_{\rm sh}^\top g_{\rm sh}$, must be informative before the encoder can follow it.

% =====================================================================

\begin{table*}[ht]
  \centering\footnotesize
  \caption{\textbf{Key notation.} Symbols used in the method and proofs.}
  \label{tab:supp_notation}
  \renewcommand{\arraystretch}{1.12}
  \begin{tabular}{@{}
    p{0.17\linewidth}@{\hspace{0.02\linewidth}}
    p{0.29\linewidth}@{\hspace{0.04\linewidth}}
    p{0.17\linewidth}@{\hspace{0.02\linewidth}}
    p{0.29\linewidth}@{}}
    \toprule
    \textbf{Symbol} & \textbf{Meaning}
      & \textbf{Symbol} & \textbf{Meaning} \\
    \midrule
    \multicolumn{4}{@{}l}{\emph{Generation and watermarking}}\\
    $z\in\R^{C\times h\times w}$ & Image VAE latent
      & $\mathcal E$, $\mathcal D$ & Frozen VAE encoder and decoder \\
    $F_k$ & Sampler update at step $k$
      & $N$ & Number of sampling steps \\
    $z_T=z_1$ & Initial noise
      & $z_0:=z_{N+1}$ & Final latent \\
    $z_k^\circ$ & Clean latent from the same $z_T$
      & $s$, $\bar s=2s-1$ & Key bits and signs \\
    $L$, $\ell$ & Payload length and bit index ($L=256$)
      & $E_\phi$, $D_\psi$ & Watermark encoder and decoder \\
    $\delta=E_\phi(s)$ & Key perturbation
      & $\eta_k$, $\bar\eta$ & Step weight and total $\sum_k\eta_k$ \\
    $\mathcal I$, $\rho$ & Late window and fraction $|\mathcal I|/N$
      & $K$, $\eta_k^{\rm sh}=1/K$ & Short-path steps and weights \\
    $z_0^{\rm sh}$, $z_0^{\rm full}$ & Short- and full-path final latents
      & $P_\ell$, $H_\ell$ & Frequency patterns and scoring templates \\
    $\beta_{\rm sc}$ & Pattern amplitude
      & $o_\ell^{\rm dec}$, $o_\ell^{\rm sc}$, $o_\ell$ & Decoder margin, frequency score, and sum \\
    $\mathcal L_{\rm rec}$ & Bit-reconstruction loss
      & $\mathcal L_{\rm sh}$, $\mathcal L_{\rm full}$ & Short- and full-path objectives \\
    \addlinespace[3pt]
    \multicolumn{4}{@{}l}{\emph{Theoretical analysis}}\\
    $\Phi(\delta)$ & Final imprinted latent
      & $\nabla$ & Jacobian \\
    $\|\cdot\|$ & Euclidean or spectral norm
      & $\lambda_{\min}$ & Smallest eigenvalue \\
    $\sigma_{\min}$, $\sigma_{\max}$ & Smallest and largest singular values
      & $R_0$ & Radius around clean latents \\
    $\ell_k$ & Constant for the step-$k$ Jacobian bound
      & $\Lambda_k$ & Remaining-step total $\sum_{j=k}^N\ell_j$ \\
    $\LamI$ & Window total $\sum_{k\in\mathcal I}\ell_k$
      & $J_k(\delta)$ & Jacobian from the input of step $k$ to $z_0$ \\
    $J_k=J_k(0)$ & Jacobian on the clean trajectory
      & $A$ & $\nabla\Phi(0)=\sum_k\eta_kJ_k$ \\
    $r_{\mathcal I}$ & Allowed radius $R_0e^{-\LamI}/\bar\eta$
      & $m$ & Lower bound on clean-trajectory signed gain \\
    $\delta^{(s)}$ & Segment $\delta'+s(\delta-\delta')$
      & $u$ & Unit direction along that segment \\
    $\bar m_{\delta,\delta'}$ & Signed gain averaged along the segment
      & $s_k$ & Identity coefficient in $\nabla F_k$ \\
    $\Sigma_k(z)$ & Posterior covariance of $x_0$ given $z_k=z$
      & $\ell_P(\delta)$ & Decoding loss for path $P$ \\
    $A_P$, $g_P$ & Path Jacobian and latent loss gradient
      & $\kappa$, $b_\delta$ & Gradient scale and bias bound \\
    $e_i$, $\sigma_i$ & Prior eigenvector and standard deviation
      & $\pi_{k,i}$ & Eigenvalue of $z_T\mapsto z_k^\circ$ along $e_i$ \\
    $a_i^{\rm full}$, $a_i^{\rm sh}$ & Eigenvalues of $A_{\rm full}$ and $A_{\rm sh}$
      & $\kappa_i$ & Ratio $a_i^{\rm sh}/a_i^{\rm full}$ \\
    $\gamma_i$ & Late-imprinting gain
      & $d$ & Detectability index \\
    \bottomrule
  \end{tabular}
\end{table*}

\section{Additional Theoretical Results and Proofs}
\label{sec:supp_proofs}
We summarize the notation in \cref{sec:supp_notation} and prove \cref{thm:channel} in \cref{sec:supp_proof_thm1}. \cref{sec:supp_margin} explains when the bound guarantees positive signed gain in every direction and examines the one-step Jacobians of an ideal denoiser (\cref{lem:psd}). \cref{sec:supp_measured} examines sampled watermark propagation, the imprinting window, and separation of encoded key pairs. \cref{sec:supp_proof_thm2} proves \cref{thm:surrogate}, and \cref{sec:supp_proof_prop1} presents an illustrative Gaussian model. The analysis uses ODE stability~\cite{dmHairer1993}, matrix perturbation theory~\cite{dmStewart1990}, and inexact-gradient analysis~\cite{dmBottou2018,dmAjalloeian2020}.

\subsection{Notation}
\label{sec:supp_notation}
\cref{tab:supp_notation} lists the key notation. Throughout, $C$ is the number of latent channels and $R$ is the number of correction rounds.

\subsection{Proof of \cref{thm:channel}}
\label{sec:supp_proof_thm1}
\begin{proof}
Let $\zeta_k=e^{\ell_k}$. \cref{asm:regular} gives $\|\nabla F_k\|\le\zeta_k$ within the allowed distance of the clean latent at step $k$. For $j\le k$, define $\Pi_{j:k} = \prod_{i=j}^{k}\zeta_i = \exp\big(\sum_{i=j}^{k}\ell_i\big)$Set this product to $1$ when it contains no factors ($j>k$). We use $\Pi_{k:N}=e^{\Lambda_k}$ for $k\in\mathcal I$ and $\Pi_{j:k}\le e^{\LamI}$ for $j,k\in\mathcal I$. The second bound holds because all included steps lie in the window. In part~(i), imprinting is restricted to $\mathcal I=\{N-|\mathcal I|+1,\dots,N\}$, with $\eta_k=0$ for $k\notin\mathcal I$. Define the accumulated weight $\bar\eta_k=\sum_{j\le k}\eta_j$ and its final value is $\bar\eta_N=\bar\eta$.

\textbf{Part (i): the trajectory stays within the assumed neighborhoods.} Fix $\|\delta\|\le r_{\mathcal I}$. Define the difference from the clean latent $\Delta_k=z_k-z_k^\circ$, the input $y_k=z_k+\eta_k\delta$ of step $k$, and its deviation $\widetilde\Delta_k=y_k-z_k^\circ=\Delta_k+\eta_k\delta$. Before the window, $\Delta_k=\widetilde\Delta_k=0$. We verify the following bounds at the first step in $\mathcal I$ and show that they remain valid from step $k$ to step $k+1$:
\begin{equation}
  \begin{aligned}
    \|\Delta_k\|&\le\|\delta\|\sum_{j<k}\eta_j\,{\Pi_{j:k-1}},\\
    \|\widetilde\Delta_k\|&\le\Big(\sum_{j<k}\eta_j\,{\Pi_{j:k-1}}+\eta_k\Big)\|\delta\|
      \le {e^{\LamI}}\bar\eta_k\|\delta\|\le R_0,
  \end{aligned}
  \label{eq:supp_trajectory_bound}
\end{equation}
where the last two inequalities use $\Pi_{j:k-1}\le e^{\LamI}$ for $j,k\in\mathcal I$ and $e^{\LamI}\bar\eta\,r_{\mathcal I}=R_0$. At the first window step, $\Delta_k=0$ and $\|\widetilde\Delta_k\|=\eta_k\|\delta\|$. If \cref{eq:supp_trajectory_bound} holds at step $k$, the line segment $\{z_k^\circ+t\widetilde\Delta_k:0\le t\le1\}$ lies within the radius-$R_0$ neighborhood of $z_k^\circ$ in \cref{asm:regular}, so
\begin{equation*}
  \begin{aligned}
    \Delta_{k+1}&=F_k(y_k)-F_k(z_k^\circ)=\int_0^1\nabla F_k(z_k^\circ+t\widetilde\Delta_k)\,\widetilde\Delta_k\,dt,\\
    \|\Delta_{k+1}\|&\le{\zeta_k}\|\widetilde\Delta_k\|\le\|\delta\|\sum_{j<k+1}\eta_j\,{\Pi_{j:k}},
  \end{aligned}
\end{equation*}
which is \cref{eq:supp_trajectory_bound} at $k+1$, because $\zeta_k\Pi_{j:k-1}=\Pi_{j:k}$ and $\zeta_k\eta_k=\eta_k\Pi_{k:k}$. Therefore, every input $y_k$ stays within distance $R_0$ of $z_k^\circ$, where \cref{asm:regular} applies at every step. Each update is continuously differentiable, and their composition $\Phi$ is $C^1$ for $\|\delta\|\le r_{\mathcal I}$.

textbf{The Jacobian of $\Phi$.} Let $Z_k=\nabla_\delta z_k$, $Y_k=Z_k+\eta_k\mathbf I$ and $G_k=\nabla F_k(y_k)$, the one-step Jacobian at the input of the imprinted trajectory. Differentiating the updates gives $Z_1=0$ and $Z_{k+1}=G_kY_k$. Expanding this recursion gives
\begin{equation}
  \nabla\Phi(\delta)=Z_{N+1}=\sum_{k=1}^N\eta_k\,G_N\cdots G_k=\sum_{k\in\mathcal I}\eta_k\,J_k(\delta),
  \label{eq:supp_jacobian_sum}
\end{equation}
which reduces to $A=\sum_k\eta_kJ_k$ at $\delta=0$, where $G_k=\nabla F_k(z_k^\circ)$. The recursion uses only the differentiability of each $F_k$ at the inputs $y_k$, so the first two equalities in \cref{eq:supp_jacobian_sum} hold for arbitrary weights $\eta_k$ wherever the imprinted trajectory is differentiable. This is the decomposition $\nabla\Phi(\delta) = \sum_k\eta_kJ_k(\delta)$ stated before \cref{asm:regular}, which part~(ii) uses without the radius condition. Expanding the product difference gives
\begin{equation}
  J_k(\delta)-\mathbf I=\sum_{j=k}^{N}G_N\cdots G_{j+1}\,(G_j-\mathbf I),
  \label{eq:supp_telescoping}
\end{equation}
with the empty product equal to $\mathbf I$ for $j=N$, together with \cref{asm:regular} at the inputs $y_j$ gives, for $k\in\mathcal I$,
\begin{equation}
  \begin{aligned}
      \|J_k(\delta)-\mathbf I\|&\le\sum_{j=k}^N\Pi_{j+1:N}\,(\zeta_j-1)\\
      &=\sum_{j=k}^N\big(\Pi_{j:N}-\Pi_{j+1:N}\big)=\Pi_{k:N}-1\\
      &=e^{\Lambda_k}-1\le e^{\LamI}-1,
  \end{aligned}
  \label{eq:supp_near_identity}
\end{equation}
The first inequality bounds each matrix product by the product of its factor norms. The sum equals $\Pi_{k:N}-1$ because intermediate terms cancel, even when every $\zeta_j=1$. The last inequality uses $\Lambda_k\le\LamI$ for $k\in\mathcal I$. Only steps $k,\dots,N$ contribute to this bound. For the uniform choice $\ell_j=L_f/N$, the bound becomes $e^{L_f(N-k+1)/N}-1\le e^{L_f\rho}-1$. Multiplying \cref{eq:supp_near_identity} by $\eta_k$ and summing over $k\in\mathcal I$ gives
\begin{equation}
  \begin{aligned}
  \|\nabla\Phi(\delta)-\bar\eta\mathbf I\|&\le\sum_{k\in\mathcal I}\eta_k\|J_k(\delta)-\mathbf I\|\\
  &\le{\sum_{k\in\mathcal I}\eta_k\,(e^{\Lambda_k}-1)\le\bar\eta\,(e^{\LamI}-1)},
  \end{aligned}
  \label{eq:supp_direct_addition}
\end{equation}
This proves the Jacobian bound $\|\nabla\Phi(\delta)-\bar\eta\mathbf I\|\le\varepsilon_{\mathcal I}$ in part~(i).

\textbf{Comparing two perturbations.} For $\|\delta\|,\|\delta'\|\le r_{\mathcal I}$, every point $\delta^{(s)}=\delta'+s(\delta-\delta')$ on the straight line between them satisfies $\|\delta^{(s)}\|\le(1-s)\|\delta'\|+s\|\delta\|\le r_{\mathcal I}$. Thus, $\Phi$ is $C^1$ and \cref{eq:supp_direct_addition} holds along the entire line segment. The fundamental theorem of calculus gives the exact difference
\begin{equation}
  \begin{aligned}
    \Phi(\delta)-\Phi(\delta')&=\int_0^1\nabla\Phi(\delta^{(s)})(\delta-\delta')\,ds\\
    &=\bar\eta(\delta-\delta')+\underbrace{\int_0^1\big(\nabla\Phi(\delta^{(s)})-\bar\eta\mathbf I\big)(\delta-\delta')\,ds}_{E},
  \end{aligned}
\end{equation}
and \cref{eq:supp_direct_addition} bounds $\|E\|\le{\sum_{k\in\mathcal I}\eta_k(e^{\Lambda_k}-1)}\|\delta-\delta'\|{\le\bar\eta(e^{\LamI}-1)\|\delta-\delta'\|}$, which proves the bound on deviation from direct addition in \cref{thm:channel}(i). For the signed gain of \cref{eq:signed_gain}, the same Jacobian bound gives, for every $s\in[0,1]$ and the unit vector $u=(\delta-\delta')/\|\delta-\delta'\|$,
\begin{equation}
  \begin{aligned}
    u^\top\nabla\Phi(\delta^{(s)})u
    &=\bar\eta+
      \underbrace{
        u^\top\big(\nabla\Phi(\delta^{(s)})-\bar\eta\mathbf I\big)u
      }_{\substack{
        \ge-\|\nabla\Phi(\delta^{(s)})-\bar\eta\mathbf I\|\\
        \text{Cauchy--Schwarz}
      }}\\
    &\ge\bar\eta-
      \underbrace{
        \big\|\nabla\Phi(\delta^{(s)})-\bar\eta\mathbf I\big\|
      }_{\substack{
        \le\varepsilon_{\mathcal I}\\
        \text{by \cref{eq:supp_direct_addition}}
      }}
      \ge\bar\eta-\varepsilon_{\mathcal I}.
  \end{aligned}
\end{equation}
The bound holds for every $s\in[0,1]$. Integrating over $s$ and using
$\varepsilon_{\mathcal I}\le\bar\eta(e^{\LamI}-1)$
proves \cref{eq:worst_case_gain}.
The same result follows from \cref{eq:directional_margin} in part~(ii), proved below, using
$\|J_k(\delta^{(s)})u-u\|
\le\|J_k(\delta^{(s)})-\mathbf I\|
\le e^{\Lambda_k}-1$
from \cref{eq:supp_near_identity} and
$\sum_{k\in\mathcal I}\eta_k=\bar\eta$.
The bounds expressed using $\LamI$ depend on $\bar\eta$, $R_0$, and $\LamI$.

\textbf{Equal weights and a single final step.} With $\eta_k=\bar\eta/|\mathcal I|$ inside the window, the Jacobian bound of part~(i) becomes $\|\nabla\Phi(\delta)-\bar\eta\mathbf I\|\le(\bar\eta/|\mathcal I|)\sum_{k\in\mathcal I}(e^{\Lambda_k}-1)$. This averages the bounds for the individual steps instead of using the largest bound at every step. For a single final step, $\mathcal I=\{N\}$, the bound is $\bar\eta(e^{\ell_N}-1)$ and depends only on the final step. The uniform form instead uses $e^{L_f/N}-1$, with $L_f$ determined by the worst case over the full trajectory.

\textbf{Part (ii): separation of the final latents.} Let $\Delta_\delta=\delta-\delta'$, $u=\Delta_\delta/\|\Delta_\delta\|$ and $\delta^{(s)}=\delta'+s\Delta_\delta$. Integrating the continuous Jacobian of $\Phi$ along $\delta^{(s)}$ for $s\in[0,1]$ gives
\begin{equation}
  \begin{aligned}
    \Phi(\delta)-\Phi(\delta')&=\int_0^1\nabla\Phi(\delta^{(s)})\,\Delta_\delta\,ds,\\
    u^\top\big(\Phi(\delta)-\Phi(\delta')\big)&=\|\Delta_\delta\|\int_0^1u^\top\nabla\Phi(\delta^{(s)})u\,ds\\
    &=\bar m_{\delta,\delta'}\|\Delta_\delta\|.
  \end{aligned}
\end{equation}
The Cauchy--Schwarz inequality gives $\|\Phi(\delta)-\Phi(\delta')\|\ge u^\top(\Phi(\delta)-\Phi(\delta'))$, which is \cref{eq:channel_monotone}. Therefore, the distance is positive whenever $\bar m_{\delta,\delta'}>0$.

Two lower bounds on the signed gain follow from the integrand. First, $u^\top\nabla\Phi(\delta^{(s)})u=u^\top\tfrac12(\nabla\Phi+\nabla\Phi^\top)(\delta^{(s)})u$, which is at least the smallest eigenvalue of the symmetric part of $\nabla\Phi(\delta^{(s)})$; integrating over $s$ gives $\bar m_{\delta,\delta'}\ge\min_{s\in[0,1]}\lambda_{\min}\big(\tfrac12(\nabla\Phi+\nabla\Phi^\top)(\delta^{(s)})\big)$. Second, when each $F_k$ is differentiable at the inputs of the imprinted trajectory of $\delta^{(s)}$, the recursion behind \cref{eq:supp_jacobian_sum} gives $\nabla\Phi(\delta^{(s)})=\sum_{k=1}^N\eta_kJ_k(\delta^{(s)})$ for arbitrary weights $\eta_k\ge0$, and hence
\begin{equation}
  \begin{aligned}
    u^\top\nabla\Phi(\delta^{(s)})u&=\sum_{k=1}^N\eta_k\Big(1+u^\top\big(J_k(\delta^{(s)})u-u\big)\Big)\\
    &\ge\sum_{k=1}^N\eta_k\Big(1-\big\|J_k(\delta^{(s)})u-u\big\|\Big),
  \end{aligned}
  \label{eq:supp_per_step_gain}
\end{equation}
where the inequality follows from applying Cauchy--Schwarz to each term. Bounding each $\|J_k(\delta^{(s)})u-u\|$ uniformly over $s$ and integrating gives \cref{eq:directional_margin}. 
\end{proof}

\subsection{Scope and structural conditions for positive signed gain}
\label{sec:supp_margin}
This subsection separates the scope of \cref{thm:channel} from an additional structural result for ideal samplers. We first explain when the general bound guarantees separation and distinguish a clean-trajectory statement from a finite-perturbation guarantee. We then prove the one-step posterior-mean identity in \cref{lem:psd} and specify the additional conditions that make it useful for multi-step propagation. The proof of \cref{thm:channel} does not rely on this lemma.

\subsubsection{Scope of the bound on deviation from direct addition}
\label{sec:supp_bound_scope}
\textbf{When the bound guarantees separation.} Under \cref{asm:regular}, \cref{thm:channel}(i) applies to perturbations satisfying $\|\delta\|,\|\delta'\|\le r_{\mathcal I}$. It gives
\begin{equation}
  \begin{aligned}
    \bar m_{\delta,\delta'}&\ge\bar\eta-\varepsilon_{\mathcal I},\\
    \varepsilon_{\mathcal I}
      &=\sum_{k\in\mathcal I}\eta_k(e^{\Lambda_k}-1)
      \le\bar\eta(e^{\LamI}-1).
  \end{aligned}
  \label{eq:supp_scope_sharp_bound}
\end{equation}
Thus, $\varepsilon_{\mathcal I}<\bar\eta$ guarantees positive signed gain for every pair in this ball. The simpler condition $\LamI<\log2$ is sufficient because it makes the last upper bound smaller than $\bar\eta$. Failure of this simpler condition does not rule out $\varepsilon_{\mathcal I}<\bar\eta$. Failure of either sufficient condition also does not imply that two perturbations become indistinguishable. The response estimate remains valid under its stated assumptions even when its lower bound on signed gain is nonpositive. Imprinting at every step is covered by choosing $\mathcal I=\{1,\dots,N\}$, provided the same regularity and radius requirements hold. We discuss window-length measurements and key-pair separation in \cref{sec:supp_measured}.

\textbf{Clean-trajectory gain and finite perturbations.} Wherever the imprinted trajectory is continuously differentiable, define
\begin{equation}
    m(\delta) = \sum_{k=1}^N\eta_k\lambda_{\min}\!\left(
          \frac{J_k(\delta)+J_k(\delta)^\top}{2}\right), \; \text{and} \;
    m=m(0).
  \label{eq:supp_pointwise_margin}
\end{equation}
Since $\nabla\Phi(\delta)=\sum_k\eta_kJ_k(\delta)$ and $\eta_k\ge0$, every unit vector $u$ satisfies
\begin{equation}
  \begin{aligned}
    u^\top\nabla\Phi(\delta)u
      &=\sum_k\eta_k u^\top
          \frac{J_k(\delta)+J_k(\delta)^\top}{2}u\\
      &\ge m(\delta).
  \end{aligned}
  \label{eq:supp_pointwise_gain}
\end{equation}
At $\delta=0$, this gives $\lambda_{\min}(\tfrac12(A+A^\top))\ge m$. If $m>0$, Cauchy--Schwarz gives $\|Au\|\ge u^\top Au\ge m$, hence $\sigma_{\min}(A)\ge m$. This guarantees positive first-order signed gain in every direction on the clean trajectory. By continuity, the signed gain remains bounded below by a positive constant in some sufficiently small neighborhood of $0$. However, this argument alone does not specify a radius that contains the perturbations used for generation.

For prescribed distinct perturbations, the relevant Jacobians are those along $\delta^{(s)}=\delta'+s(\delta-\delta')$. Integrating \cref{eq:supp_pointwise_gain} gives
\begin{equation}
  \bar m_{\delta,\delta'}
    \ge\int_0^1 m(\delta^{(s)})\,ds
    \ge\underbrace{\inf_{s\in[0,1]}m(\delta^{(s)})}_{\substack{
      \text{uniform lower bound}\\
      \text{along the segment}
    }}.
  \label{eq:supp_segment_margin}
\end{equation}
If this uniform lower bound is strictly positive, separation is guaranteed for that pair. Under the assumptions of part~(i), \cref{eq:supp_near_identity} yields, throughout the prescribed ball,
\begin{equation}
    m(\delta) \ge \sum_{k\in\mathcal I}\eta_k(2-e^{\Lambda_k})
      = \bar\eta-\varepsilon_{\mathcal I}
      \ge\bar\eta(2-e^{\LamI}).
    \label{eq:supp_scope_margin}
\end{equation}
This is the connection between a bound on each propagation Jacobian and the finite-pair separation statement.

\subsubsection{Ideal posterior-mean samplers}
\label{sec:supp_ideal_samplers}
The following lemma identifies a separate source of nonnegative gain at one sampling step. It supports an idealized explanation.

\begin{lemma}[One-step Jacobians of ideal posterior-mean samplers]
\label{lem:psd}
Let $x_0\in\mathbb R^d$ satisfy $\mathbb E\|x_0\|^2<\infty$, and let $\epsilon\sim\mathcal N(0,\mathbf I)$ be independent of $x_0$. Assume exact posterior-mean prediction without guidance. The posterior is defined by the forward corruption model at step $k$, and
$\Sigma_k(z)=\operatorname{Cov}(x_0\mid z_k=z)$. Any conditioning information is held fixed.

For deterministic DDIM with $0<\bar\alpha_k<\bar\alpha_{k+1}\le1$ and $z_k=\sqrt{\bar\alpha_k}x_0+\sqrt{1-\bar\alpha_k}\epsilon$,
\begin{equation}
  \begin{gathered}
    \nabla F_k(z)=s_k\mathbf I+c_k\Sigma_k(z), \; \text{with} \\
    s_k=\sqrt{\frac{1-\bar\alpha_{k+1}}{1-\bar\alpha_k}},\;
    c_k=\frac{\sqrt{\bar\alpha_k}}{1-\bar\alpha_k}
      \left(\sqrt{\bar\alpha_{k+1}}-s_k\sqrt{\bar\alpha_k}\right)>0.
  \end{gathered}
  \label{eq:supp_psd_ddim}
\end{equation}
For the flow-matching Euler update with $0\le\sigma_{k+1}<\sigma_k\le1$ and $z_k=(1-\sigma_k)x_0+\sigma_k\epsilon$,
\begin{equation}
  \begin{gathered}
    \nabla F_k(z)=s_k\mathbf I+c_k\Sigma_k(z), \; \text{with}\\
    s_k=\frac{\sigma_{k+1}}{\sigma_k},\qquad
    c_k=\frac{(\sigma_k-\sigma_{k+1})(1-\sigma_k)}{\sigma_k^3}\ge0.
  \end{gathered}
  \label{eq:supp_psd_flow}
\end{equation}
Here $c_k>0$ when $0<\sigma_k<1$, and $c_k=0$ when $\sigma_k=1$. In both cases, $\nabla F_k(z)$ is symmetric positive semidefinite and $\lambda_{\min}(\nabla F_k(z))\ge s_k$.
\end{lemma}

\begin{proof}
\textbf{Posterior-mean derivative.} We first establish the Gaussian conditional-mean derivative identity directly.\footnote{See also A.~Dytso, H.~V.~Poor, and S.~Shamai (Shitz), ``Conditional Mean Estimation in Gaussian Noise: A Meta Derivative Identity With Applications,'' \emph{IEEE Transactions on Information Theory}, 69(3):1883--1898, 2023, \href{https://doi.org/10.1109/TIT.2022.3216012}{doi:10.1109/TIT.2022.3216012}.}
Let $X=x_0$, $Z=aX+b\epsilon$, $a\ge0$, and $b>0$. Write
\begin{equation*}
  \mu(z)=\mathbb E[X\mid Z=z],\qquad
  \Sigma(z)=\operatorname{Cov}(X\mid Z=z).
\end{equation*}
Let $P_X$ be the law of $X$, which need not have a density. Define
\begin{equation}
  \begin{aligned}
    L(z,x)&=\exp\!\left(-\frac{\|z-ax\|^2}{2b^2}\right),\\
    D(z)&=\int L(z,x)\,dP_X(x)>0,\\
    M_i(z)&=\int x_iL(z,x)\,dP_X(x).
  \end{aligned}
  \label{eq:supp_posterior_integrals}
\end{equation}
Bayes' formula gives $\mu_i(z)=M_i(z)/D(z)$. For each observation coordinate $j$,
\begin{equation}
  \partial_j L(z,x)=-\frac{z_j-ax_j}{b^2}L(z,x).
  \label{eq:supp_likelihood_derivative}
\end{equation}
On a compact set of $z$-values, $|\partial_jL(z,x)|$ is bounded by a constant times $1+|x_j|$, and $|x_i\partial_jL(z,x)|$ is bounded by a constant times $|x_i|+|x_ix_j|$. These functions are integrable because $\mathbb E\|X\|^2<\infty$. Dominated convergence therefore justifies differentiation under the integrals and continuity of the resulting derivatives. Substituting \cref{eq:supp_likelihood_derivative} gives
\begin{equation}
  \begin{aligned}
    \frac{\partial_jD(z)}{D(z)}
      &=-\frac{z_j-a\mu_j(z)}{b^2},\\
    \frac{\partial_jM_i(z)}{D(z)}
      &=-\frac{z_j\mu_i(z)-a\mathbb E[X_iX_j\mid Z=z]}{b^2}.
  \end{aligned}
  \label{eq:supp_posterior_integral_derivatives}
\end{equation}
The quotient rule now gives
\begin{equation}
  \begin{aligned}
    \partial_j\mu_i(z)
      &=\frac{\partial_jM_i(z)}{D(z)}
        -\mu_i(z)\frac{\partial_jD(z)}{D(z)}\\
      &=\frac{a}{b^2}\left(
        \mathbb E[X_iX_j\mid Z=z]-\mu_i(z)\mu_j(z)\right)\\
      &=\frac{a}{b^2}\Sigma_{ij}(z).
  \end{aligned}
  \label{eq:supp_posterior_quotient_rule}
\end{equation}
Thus,
\begin{equation}
  \nabla\mu(z)=\frac{a}{b^2}\Sigma(z).
  \label{eq:supp_posterior_mean_jacobian}
\end{equation}
This is an exact identity. The prior distribution can be non-Gaussian, and the dependence of $\Sigma(z)$ on $z$ allows the posterior mean to be nonlinear.

\textbf{DDIM update.} Set $\alpha=\bar\alpha_k$ and $\beta=\bar\alpha_{k+1}$, and write $\mu_k(z)=\mathbb E[x_0\mid z_k=z]$. Taking conditional expectations in $z_k=\sqrt\alpha x_0+\sqrt{1-\alpha}\epsilon$ gives
\begin{equation}
  \widehat\epsilon_k(z)
    =\frac{z-\sqrt\alpha\mu_k(z)}{\sqrt{1-\alpha}}.
  \label{eq:supp_ddim_posterior_noise}
\end{equation}
Substituting this expression into deterministic DDIM yields
\begin{equation}
  \begin{aligned}
    F_k(z)
      &=\sqrt\beta\mu_k(z)+\sqrt{1-\beta}\widehat\epsilon_k(z)\\
      &=s_kz+\left(\sqrt\beta-s_k\sqrt\alpha\right)\mu_k(z),\\
    s_k&=\sqrt{\frac{1-\beta}{1-\alpha}}.
  \end{aligned}
  \label{eq:supp_ddim_posterior_update}
\end{equation}
Applying \cref{eq:supp_posterior_mean_jacobian} with $a=\sqrt\alpha$ and $b=\sqrt{1-\alpha}$ gives
\begin{equation}
  \nabla\mu_k(z)=\frac{\sqrt\alpha}{1-\alpha}\Sigma_k(z).
  \label{eq:supp_ddim_posterior_jacobian}
\end{equation}
Differentiating \cref{eq:supp_ddim_posterior_update} therefore proves \cref{eq:supp_psd_ddim}. To check $c_k>0$, observe that
\begin{equation*}
  \beta-s_k^2\alpha
    =\beta-\frac{\alpha(1-\beta)}{1-\alpha}
    =\frac{\beta-\alpha}{1-\alpha}>0.
\end{equation*}
Since $\sqrt\beta$ and $s_k\sqrt\alpha$ are nonnegative, this implies $\sqrt\beta>s_k\sqrt\alpha$. The prefactor $\sqrt\alpha/(1-\alpha)$ is also positive. At the final step $\beta=1$, the update is $F_k=\mu_k$ and the same calculation applies.

\textbf{Flow-matching update.} Set $\sigma=\sigma_k$ and $\tau=\sigma_{k+1}$, and let $\mu_\sigma(z)=\mathbb E[x_0\mid Z_\sigma=z]$ for $Z_\sigma=(1-\sigma)x_0+\sigma\epsilon$. Differentiating this interpolation with fixed endpoints gives $dZ_\sigma/d\sigma=\epsilon-x_0$. Moreover, $Z_\sigma-x_0=\sigma(\epsilon-x_0)$, so the exact conditional velocity is
\begin{equation}
  \begin{aligned}
    v_\sigma(z)
      &=\mathbb E[\epsilon-x_0\mid Z_\sigma=z]\\
      &=\frac{z-\mu_\sigma(z)}{\sigma}.
  \end{aligned}
  \label{eq:supp_flow_posterior_velocity}
\end{equation}
The Euler update is consequently
\begin{equation}
  \begin{aligned}
    F_k(z)
      &=z+(\tau-\sigma)v_\sigma(z)\\
      &=\frac{\tau}{\sigma}z+
        \frac{\sigma-\tau}{\sigma}\mu_\sigma(z).
  \end{aligned}
  \label{eq:supp_flow_posterior_update}
\end{equation}
Using \cref{eq:supp_posterior_mean_jacobian} with $a=1-\sigma$ and $b=\sigma$,
\begin{equation*}
  \nabla\mu_\sigma(z)=\frac{1-\sigma}{\sigma^2}\Sigma_k(z).
\end{equation*}
Differentiation of \cref{eq:supp_flow_posterior_update} gives
\begin{equation*}
  \nabla F_k(z)=\frac{\tau}{\sigma}\mathbf I+
    \frac{(\sigma-\tau)(1-\sigma)}{\sigma^3}\Sigma_k(z),
\end{equation*}
which proves \cref{eq:supp_psd_flow}. Both coefficients are nonnegative. The covariance coefficient vanishes at $\sigma=1$ and is positive for $0<\sigma<1$.

\textbf{Symmetry and eigenvalue bound.} In both cases $\nabla F_k=s_k\mathbf I+c_k\Sigma_k$ with $s_k,c_k\ge0$. The covariance matrix is symmetric, and for any $v\in\mathbb R^d$,
\begin{equation}
  \begin{aligned}
    v^\top\nabla F_k(z)v
      &=s_k\|v\|^2+c_k\operatorname{Var}(v^\top x_0\mid z_k=z)\\
      &\ge s_k\|v\|^2.
  \end{aligned}
  \label{eq:supp_posterior_psd_bound}
\end{equation}
Thus $\nabla F_k(z)\succeq s_k\mathbf I\succeq0$, and minimizing over unit vectors proves the claimed eigenvalue bound.
\end{proof}

\subsection{Proof of \cref{thm:surrogate}}
\label{sec:supp_proof_thm2}

\begin{proof}
\textbf{Bounding the difference between the gradients.}
Fix $\kappa>0$ and write $p=\kappa^{-1}\nabla_\delta\ell_{\rm sh},$ $q=\nabla_\delta\ell_{\rm full}.$
The chain rule gives $\nabla_\delta\ell_P=A_P^\top g_P$. Therefore, adding and subtracting $A_{\rm full}^\top g_{\rm sh}$ yields
\begin{equation}
    p-q = (\kappa^{-1}A_{\rm sh}-A_{\rm full})^\top g_{\rm sh} \; + \; A_{\rm full}^\top(g_{\rm sh}-g_{\rm full}).
\end{equation}
By the triangle inequality and the definition of $b_\delta$,
\begin{equation}
  \begin{aligned}
    \|p-q\|
      &\le\|(\kappa^{-1}A_{\rm sh}-A_{\rm full})^\top g_{\rm sh}\| + \|A_{\rm full}^\top(g_{\rm sh}-g_{\rm full})\|\\
      &=b_\delta.
  \end{aligned}
\end{equation}
Using the induced matrix norm also gives
\begin{equation}
    b_\delta \le \|\kappa^{-1}A_{\rm sh}-A_{\rm full}\|\,\|g_{\rm sh}\| + \|A_{\rm full}\|\,\|g_{\rm sh}-g_{\rm full}\|.
  \label{eq:supp_opnorm_bias}
\end{equation}

\textbf{Positive alignment and the cosine bound.}
Suppose $b_\delta<\|q\|$. Then $q\ne0$, and the reverse triangle inequality gives $\|p\|\ge\|q\|-b_\delta>0$. By Cauchy--Schwarz, we have:
\begin{equation}
  \begin{aligned}
    \langle p,q\rangle
      &=\|q\|^2+\langle p-q,q\rangle\\
      &\ge\|q\|^2-\|p-q\|\,\|q\|\\
      &\ge\|q\|\underbrace{(\|q\|-b_\delta)}_{>0}>0.
  \end{aligned}
\end{equation}
Let $\theta=\angle(p,q)$. The positive inner product gives $\cos\theta>0$. Expanding the squared distance and completing the square,
\begin{equation}
  \begin{aligned}
    b_\delta^2
      &\ge\|p-q\|^2\\
      &=\|p\|^2+\|q\|^2-2\|p\|\,\|q\|\cos\theta\\
      &=\underbrace{(\|p\|-\|q\|\cos\theta)^2}_{\ge0}
        +\|q\|^2\sin^2\theta\\
      &\ge\|q\|^2(1-\cos^2\theta).
  \end{aligned}
\end{equation}
Rearranging and using $\cos\theta>0$ gives $\cos\theta\ge\sqrt{1-b_\delta^2/\|q\|^2}$. Since $\kappa>0$, rescaling the short-path gradient does not change its angle with $q$. Hence
\begin{equation*}
  \begin{aligned}
    \cos\angle(\nabla_\delta\ell_{\rm sh},
                \nabla_\delta\ell_{\rm full})
      &=\cos\theta\\
      &\ge\sqrt{1-\frac{b_\delta^2}{\|q\|^2}}>0.
  \end{aligned}
\end{equation*}

\textbf{Descent with respect to the perturbation.}
The directional derivative of the full-sampler loss along the negative short-path gradient is
\begin{equation}
    \left.\frac{d}{dt}\ell_{\rm full}
      (\delta-t\nabla_\delta\ell_{\rm sh})\right|_{t=0} = -\langle q,\kappa p\rangle = -\kappa\underbrace{\langle p,q\rangle}_{>0}<0.
\end{equation}
Thus $-\nabla_\delta\ell_{\rm sh}$ is a descent direction for $\ell_{\rm full}$.

\textbf{Descent with respect to the encoder parameters.}
With the decoder and sampling noise fixed, both losses depend on $\phi$ only through $\delta=E_\phi(s)$. Let $B=\partial\delta/\partial\phi$, $p_\phi=\kappa^{-1}\nabla_\phi\ell_{\rm sh}$, and $q_\phi=\nabla_\phi\ell_{\rm full}$. The chain rule gives
\begin{equation}
  p_\phi-q_\phi=B^\top(p-q),\qquad
  \|p_\phi-q_\phi\|\le\|B\|b_\delta.
\end{equation}
If $\|B\|b_\delta<\|q_\phi\|$, then
\begin{equation}
  \begin{aligned}
    \langle p_\phi,q_\phi\rangle
      &=\|q_\phi\|^2+\langle p_\phi-q_\phi,q_\phi\rangle\\
      &\ge\|q_\phi\|^2-\|B\|b_\delta\|q_\phi\|\\
      &=\|q_\phi\|\underbrace{(\|q_\phi\|-\|B\|b_\delta)}_{>0}>0.
  \end{aligned}
\end{equation}
Consequently,
\begin{equation}
    \left.\frac{d}{dt}\ell_{\rm full}
      (\phi-t\nabla_\phi\ell_{\rm sh})\right|_{t=0} = -\kappa\langle p_\phi,q_\phi\rangle<0.
\end{equation}
Therefore $-\nabla_\phi\ell_{\rm sh}$ is also a descent direction for the full-sampler loss.
\end{proof}

\subsubsection{Further mathematical consequences}
\label{sec:supp_surrogate_consequences}
We use the same notation $p=\kappa^{-1}\nabla_\delta\ell_{\rm sh}$ and $q=\nabla_\delta\ell_{\rm full}$ below.

\textbf{A sufficient condition based on singular values.}
Since $q=A_{\rm full}^\top g_{\rm full}$, the smallest singular value gives $\|q\|\ge\sigma_{\min}(A_{\rm full})\|g_{\rm full}\|.$ Thus, $ b_\delta<\sigma_{\min}(A_{\rm full})\|g_{\rm full}\|$ is sufficient for $b_\delta<\|q\|$ and the perturbation-space descent conclusion of \cref{thm:surrogate}. This implication uses differentiability at the point under consideration and does not require the radius condition from \cref{thm:channel}(i).

\textbf{Equality in the cosine bound.}
In dimension at least two, the vector inequality used in the proof can become an equality. For $0\le b_\delta<\|q\|$, choose a unit vector $u_\perp$ orthogonal to $q$ and define
\begin{equation}
    p_*=\left(1-\frac{b_\delta^2}{\|q\|^2}\right)q + b_\delta\sqrt{1-\frac{b_\delta^2}{\|q\|^2}}\,u_\perp.
\end{equation}
Orthogonality gives
\begin{equation*}
  \begin{aligned}
    \|p_*-q\|^2&=b_\delta^2,\\
    \|p_*\|^2&=\|q\|^2-b_\delta^2,\\
    \langle p_*,q\rangle&=\|q\|^2-b_\delta^2.
  \end{aligned}
\end{equation*}
Consequently,
\begin{equation}
  \cos\angle(p_*,q)
    =\frac{\langle p_*,q\rangle}{\|p_*\|\,\|q\|}
    =\sqrt{1-\frac{b_\delta^2}{\|q\|^2}}.
\end{equation}

\textbf{Equal decoder gradients.}
When $g_{\rm sh}=g_{\rm full}$, let $p_0=\nabla_\delta\ell_{\rm sh}$ be the unscaled short-path gradient. The second term in the definition of $b_\delta$ vanishes, so $b_\delta=\|\kappa^{-1}p_0-q\|$. If $\langle p_0,q\rangle>0$, minimizing over $t>0$ gives $t_*=\underset{t>0}{\arg\min}\,\|tp_0-q\|^2 = \frac{\langle p_0,q\rangle}{\|p_0\|^2}>0.$ Hence,
\begin{equation}
  \begin{aligned}
    \min_{t>0}\|tp_0-q\|^2
      &=\|q\|^2-\frac{\langle p_0,q\rangle^2}{\|p_0\|^2}\\
      &=\|q\|^2\sin^2\angle(p_0,q)<\|q\|^2.
  \end{aligned}
\end{equation}
Choosing $\kappa^{-1}=t_*$ therefore gives $b_\delta<\|q\|$.
Conversely, if $\|tp_0-q\|<\|q\|$ for some $t>0$, expanding the squared norms yields $t^2\|p_0\|^2-2t\langle p_0,q\rangle<0,$ and division by $t>0$ gives $2\langle p_0,q\rangle>t\|p_0\|^2\ge0$. Thus, when the decoder gradients are equal, a positive scale satisfying $b_\delta<\|q\|$ exists if and only if the two unscaled gradients have a positive inner product. Using the matrix-norm upper bound in \cref{eq:supp_opnorm_bias} can impose a stronger sufficient condition.

\subsubsection{Empirical evaluation}
\label{sec:supp_surrogate_empirical}
The empirical evaluation of short- and full-path gradient alignment with respect to $\delta$ and $\phi$ is presented in \cref{sec:gradient-alignment}.

\subsection{An illustrative Gaussian model}
\label{sec:supp_proof_prop1}
The following Gaussian model has constant Jacobians that share an eigenbasis. It illustrates how the timing of imprinting changes the watermark strength relative to natural latent variation. Part~(b) also illustrates the gradient alignment in \cref{thm:surrogate} when the decoder gradients are equal. The model is an idealized example: the tested samplers can amplify and rotate watermark perturbations at early steps, and their Jacobians over the late window need not be symmetric (\cref{sec:supp_measured}). The general bounds in the main paper do not rely on this Gaussian assumption.

\begin{proposition}[Gaussian latents]
\label{prop:gauss}
Let the data prior be $\mathcal N(\mu_{\rm p},\Sigma_{\rm p})$, where
$\Sigma_{\rm p}=\sum_i\sigma_i^2e_ie_i^\top\succ0$ and $\{e_i\}$ is
orthonormal. Both paths use deterministic DDIM with the exact unconditional minimum mean-square error (MMSE) noise predictor for this prior on strictly increasing grids $\alpha_k = \bar\alpha_{t_k}\in(0,1]$ ($\alpha_1<\dots<\alpha_{N+1}$ for the full path and $\alpha^{\rm sh}_1<\dots<\alpha^{\rm sh}_{K+1}$ for the short path). Let the full path use weights $\eta_k\ge0$ with $\bar\eta=\sum_k\eta_k>0$, and the short path weights $\eta^{\rm sh}_k\ge0$ with $\sum_{k=1}^K\eta^{\rm sh}_k>0$. Then both response Jacobians $A_{\rm full}$ and $A_{\rm sh}$ are constant and diagonal in
$\{e_i\}$, with positive eigenvalues $a_i^{\rm full}$ and $a_i^{\rm sh}$ given in \cref{eq:supp_gaussian_eigen}, and the following hold.
\begin{enumerate}[label=(\alph*),leftmargin=*]
      \item \emph{Detectability.} For $z_T\sim\mathcal N(0,\mathbf I)$ and fixed $\delta$, the likelihood-ratio test between clean and watermarked $z_0$ satisfies $\mathrm{TPR}=Q\big(Q^{-1}(\mathrm{FPR})-d\big).$ Here, $Q$ is the standard normal tail and $d^2=\bar\eta^2\sum_i\gamma_i^2\delta_i^2$, with $\delta_i=\langle e_i,\delta\rangle$ and $\gamma_i=\bar\eta^{-1}\sum_{k=1}^N\eta_k/\pi_{k,i}$. The quantity $\pi_{k,i}$ is the Jacobian eigenvalue of $z_T\mapsto z_k^\circ$ along $e_i$. Initial-noise imprinting gives $\gamma_i=1$. If $\sigma_i<1$, moving weight later strictly increases $\gamma_i$, which exceeds one whenever $\sum_{k>1}\eta_k>0$.

      \item \emph{Alignment.} If $g_{\rm sh}=g_{\rm full}\neq0$, let $\kappa_{\min}$ and $\kappa_{\max}$ be the minimum and maximum of $\kappa_i=a_i^{\rm sh}/a_i^{\rm full}$. Then
      \begin{equation}
          \cos\angle\big(
            \nabla_\delta\ell_{\rm sh},
            \nabla_\delta\ell_{\rm full}\big) \ge \frac{2\sqrt{\kappa_{\min}\kappa_{\max}}}
               {\kappa_{\min}+\kappa_{\max}}>0.
        \label{eq:gaussian_alignment}
      \end{equation}
    \end{enumerate}
\end{proposition}

\begin{proof}
Write $X_i=\langle e_i,X\rangle$ and $\mu_{{\rm p},i}=\langle e_i,\mu_{\rm p}\rangle$. For $X\sim\mathcal N(\mu_{\rm p},\Sigma_{\rm p})$ and independent $\epsilon\sim\mathcal N(0,\mathbf I)$, the noisy latent coordinate
$Y_{k,i}=\sqrt{\alpha_k}X_i+\sqrt{1-\alpha_k}\epsilon_i$ has mean $\sqrt{\alpha_k}\mu_{{\rm p},i}$ and variance $v_{k,i}=\alpha_k\sigma_i^2+1-\alpha_k>0$. Gaussian conditioning gives the exact predictors $\widehat X_{k,i}(y)=\mu_{{\rm p},i} + \frac{\sqrt{\alpha_k}\sigma_i^2}{v_{k,i}}
(y-\sqrt{\alpha_k}\mu_{{\rm p},i})$ and $\widehat\epsilon_{k,i}(y) = \frac{\sqrt{1-\alpha_k}}{v_{k,i}}(y - \sqrt{\alpha_k}\mu_{{\rm p},i})$, which satisfy $\sqrt{\alpha_k}\widehat X_{k,i}(y)+\sqrt{1-\alpha_k}\widehat\epsilon_{k,i}(y)=y$. Substitution into the deterministic DDIM update yields
\begin{equation}
  \begin{aligned}
    F_{k,i}(y)
      &=\sqrt{\alpha_{k+1}}\widehat X_{k,i}(y)
        +\sqrt{1-\alpha_{k+1}}\widehat\epsilon_{k,i}(y)\\
      &=\underbrace{
        \frac{\sqrt{\alpha_k\alpha_{k+1}}\sigma_i^2
          +\sqrt{(1-\alpha_k)(1-\alpha_{k+1})}}{v_{k,i}}
        }_{\varphi_{k,i}}\,y\\
      &\quad+\underbrace{
        (\sqrt{\alpha_{k+1}}-\varphi_{k,i}\sqrt{\alpha_k})\mu_{{\rm p},i}
        }_{\beta_{k,i}}.
  \end{aligned}
  \label{eq:supp_gaussian_ddim}
\end{equation}
For $u_{k,i}=(\sqrt{\alpha_k}\sigma_i, \sqrt{1-\alpha_k})^\top$, we have $\|u_{k,i}\|^2=v_{k,i}$, and Cauchy--Schwarz gives
$0<\varphi_{k,i}=u_{k,i}^\top u_{k+1,i}/\|u_{k,i}\|^2\le\sqrt{v_{k+1,i}/v_{k,i}}$; positivity uses $\alpha_k>0$ and $\sigma_i>0$.

Set $\pi_{k,i}=\prod_{j=1}^{k-1}\varphi_{j,i}$, with empty products equal to one, and
$\mu_{0,i}=\sum_{k=1}^N\beta_{k,i} \prod_{j=k+1}^N\varphi_{j,i}$.
Unrolling $z_{k+1,i}=F_{k,i}(z_{k,i}+\eta_k\delta_i)$ from $z_{1,i}=z_{T,i}$ gives
\begin{equation}
    z_{0,i}
      = \underbrace{\pi_{N+1,i}z_{T,i}+\mu_{0,i}}_{\text{clean latent}}+a_i^{\rm full}\delta_i,
\end{equation}
with
\begin{equation}
  \begin{aligned}
  a_i^{\rm full} &= \sum_{k=1}^N\eta_k\prod_{j=k}^N\varphi_{j,i} = \pi_{N+1,i}\sum_{k=1}^N\frac{\eta_k}{\pi_{k,i}}>0,\\
  a_i^{\rm sh}&=\sum_{k=1}^K\eta^{\rm sh}_k\prod_{j=k}^K\varphi^{\rm sh}_{j,i}>0,
  \end{aligned}
  \label{eq:supp_gaussian_eigen}
\end{equation}
where $\varphi^{\rm sh}_{j,i}$ is defined by \cref{eq:supp_gaussian_ddim} on the short-path grid, and the second expression follows by repeating the derivation on that grid. Positivity holds because every $\varphi>0$, all weights are nonnegative, and at least one weight is positive. Thus $A_{\rm full}=\sum_i a_i^{\rm full}e_ie_i^\top$ and $A_{\rm sh}=\sum_i a_i^{\rm sh}e_ie_i^\top$ are constant and diagonal.

\emph{(a) Detectability.} Let $\Delta_z=A_{\rm full}\delta$, $\mu_0=\sum_i \mu_{0,i}e_i$, and
$\Sigma_0=\sum_i\pi_{N+1,i}^2e_ie_i^\top\succ0$. Since $z_T\sim\mathcal N(0,\mathbf I)$, the clean and watermarked hypotheses are $H_0:z_0\sim\mathcal N(\mu_0,\Sigma_0)$ and $H_1:z_0\sim\mathcal N(\mu_0+\Delta_z,\Sigma_0)$.
Their log-likelihood ratio satisfies
\begin{equation}
    \log\frac{p_{H_1}(z)}{p_{H_0}(z)}
      =\Delta_z^\top \Sigma_0^{-1}(z-\mu_0)
        -\tfrac12\underbrace{\Delta_z^\top \Sigma_0^{-1}\Delta_z}_{d^2},
\end{equation}
with $d^2 = \sum_i\left(\frac{a_i^{\rm full}\delta_i}{\pi_{N+1,i}}\right)^2 = \bar\eta^2\sum_i\gamma_i^2\delta_i^2.$
For $d>0$, the statistic $\mathrm{LR}(z)=\Delta_z^\top\Sigma_0^{-1}(z-\mu_0)/d$ has distributions $\mathrm{LR}\mid H_0\sim\mathcal N(0,1)$ and $\mathrm{LR}\mid H_1\sim\mathcal N(d,1)$. The Neyman--Pearson test~\cite{dmNeyman1933} rejects when $\mathrm{LR}(z)>Q^{-1}(f)$, giving $\mathrm{FPR}=f$ and
$\mathrm{TPR}=Q(Q^{-1}(f)-d)$ for $0<f<1$. When $d=0$, the hypotheses coincide and TPR equals FPR.

Initial-noise imprinting has $\eta_1=\bar\eta$ and $\eta_k=0$ for $k>1$, so $\pi_{1,i}=1$ gives $\gamma_i=1$. If $\sigma_i<1$, then $v_{k+1,i}-v_{k,i}=(\alpha_{k+1}-\alpha_k)(\sigma_i^2-1)<0$. Hence $0<\varphi_{k,i}<1$ and $\pi_{k+1,i}^{-1}>\pi_{k,i}^{-1}$.
Moving weight $\Delta\eta>0$ from step $k$ to step $l>k$ increases $\gamma_i$ by
$\frac{\Delta\eta}{\bar\eta}\underbrace{(\pi_{l,i}^{-1}-\pi_{k,i}^{-1})}_{>0}$. Consequently, $\gamma_i>1$ whenever any weight is placed after the first step, proving (a).

\emph{(b) Alignment.} Let $g=g_{\rm sh}=g_{\rm full}\neq0$ and $g_i=\langle e_i,g\rangle$. Since both response matrices are symmetric, $\nabla_\delta\ell_P=A_P^\top g=\sum_i a_i^P g_i e_i$ for $P\in\{\mathrm{sh},\mathrm{full}\}$. Define $\varpi_i=(a_i^{\rm full})^2g_i^2/\sum_j(a_j^{\rm full})^2g_j^2$, so $\varpi_i\ge0$ and
$\sum_i\varpi_i=1$. Set $\bar\kappa=\sum_i\varpi_i\kappa_i$ and $\overline{\kappa^2}=\sum_i\varpi_i\kappa_i^2$, and note $\kappa_{\min}\le\bar\kappa\le\kappa_{\max}$. The inequality $(\kappa_i-\kappa_{\min})(\kappa_{\max}-\kappa_i)\ge0$, averaged with the weights $\varpi_i$, gives
$\overline{\kappa^2}\le U_\kappa$, where $U_\kappa=(\kappa_{\min}+\kappa_{\max})\bar\kappa-\kappa_{\min}\kappa_{\max}\ge \kappa_{\min}^2>0$. Abbreviate $\kappa_{+}=\kappa_{\min}+\kappa_{\max}$ and $\kappa_{\times}=\kappa_{\min}\kappa_{\max}$. Therefore,
\begin{equation}
  \begin{aligned}
    \cos^2\angle\big(
      \nabla_\delta\ell_{\rm sh},
      \nabla_\delta\ell_{\rm full}\big)
      &=\frac{(\sum_i\varpi_i\kappa_i)^2}
        {\sum_i\varpi_i\kappa_i^2}=\frac{\bar\kappa^2}{\overline{\kappa^2}}
      \ge\frac{\bar\kappa^2}{U_\kappa}\\
      &=\frac{4\kappa_{\times}}{\kappa_{+}^2}
        +\underbrace{
          \frac{\big(\kappa_{+}\bar\kappa-2\kappa_{\times}\big)^2}{\kappa_{+}^2U_\kappa}
          }_{\ge0}\\
      &\ge\frac{4\kappa_{\times}}{\kappa_{+}^2}.
  \end{aligned}
\end{equation}
Here the first equality uses $\langle\nabla_\delta\ell_{\rm sh},\nabla_\delta\ell_{\rm full}\rangle=\sum_i\kappa_i(a_i^{\rm full})^2g_i^2$ and $\|\nabla_\delta\ell_{\rm sh}\|^2=\sum_i\kappa_i^2(a_i^{\rm full})^2g_i^2$. Since all $\kappa_i$ are positive, the cosine is positive. Taking square roots proves the Kantorovich-type bound~\cite{dmGreub1959} in \cref{eq:gaussian_alignment}. If $\kappa_i$ is constant, the gradients are proportional and their cosine equals one.
\end{proof}

\begin{table}[ht]
  \centering
  \footnotesize
  \caption{\textbf{Key-pair separation.} Signed gains for 100 independent-key and 100 one-bit-flip pairs per generator. Minimum gains and nonpositive counts include both pair types, while median normalized gains use independent-key pairs only.}
  \label{tab:supp_measured}
  \setlength{\tabcolsep}{4pt}
  \renewcommand{\arraystretch}{1.08}
  \begin{tabular}{@{}p{0.53\columnwidth}rrr@{}}
    \toprule
    Measurement & SD~2.1 & SD3 & FLUX \\
    \midrule
    Minimum signed gain $\bar m_{\delta,\delta'}$ & 0.281 & 0.901 & 0.455 \\
    Nonpositive gains & 0/200 & 0/200 & 0/200 \\
    Median $\bar m_{\delta,\delta'}/\bar\eta$ (independent keys) & 0.70 & 0.68 & 0.31 \\
    \bottomrule
  \end{tabular}
\end{table}

\begin{figure}[ht]
  \centering
  \includegraphics[width=\columnwidth]{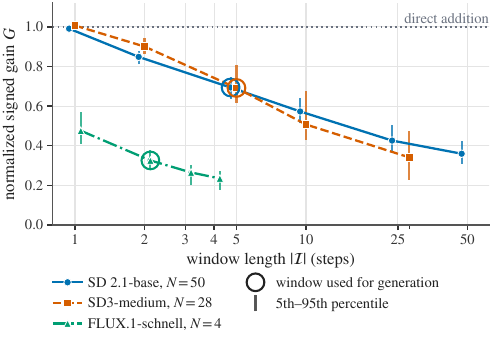}
  \caption{\textbf{Effect of imprinting-window length at fixed backbone-specific $\bar\eta$.} Median normalized signed gain $G=\bar m_{\delta,\delta'}/\bar\eta$ (5th--95th percentiles) over 20 prompt--seed settings with independent key pairs. Imprinting uses the last $|\mathcal I|$ steps with $\eta_k=\bar\eta/|\mathcal I|$, where $\bar\eta=0.5$, $2.0$, and $3.0$ for SD~2.1, SD3, and FLUX, respectively. Rings mark the generation windows, $G=1$ indicates direct addition, and small horizontal offsets aid visibility.}
  \label{fig:supp_theory_window}
\end{figure}

\subsection{Experimental analysis of watermark propagation}
\label{sec:supp_measured}

\subsubsection{Purpose and notation}
\label{sec:supp_measured_purpose}
We evaluate the two conclusions of \cref{thm:channel} on the tested samplers: the dependence of watermark propagation on the imprinting window and the separation of distinct encoded keys. Let $\Phi(\delta)=z_0$ denote the final latent before frequency-based embedding, margin correction, VAE decoding, or attacks. For two distinct perturbations, define $u=(\delta-\delta')/\|\delta-\delta'\|$ and the signed gain
\begin{equation}
  \bar m_{\delta,\delta'}
  =\frac{u^\top[\Phi(\delta)-\Phi(\delta')]}{\|\delta-\delta'\|}.
  \label{eq:supp_measured_separation}
\end{equation}
A positive signed gain guarantees separation of the corresponding pair by \cref{thm:channel}(ii). To compare the response with direct addition, we use the normalized gain $G=\bar m_{\delta,\delta'}/\bar\eta$, where $\bar\eta=\sum_k\eta_k$ is the total imprinting weight.

\begin{figure*}[ht]
  \centering\color{darkred}
  \includegraphics[width=0.99\textwidth]{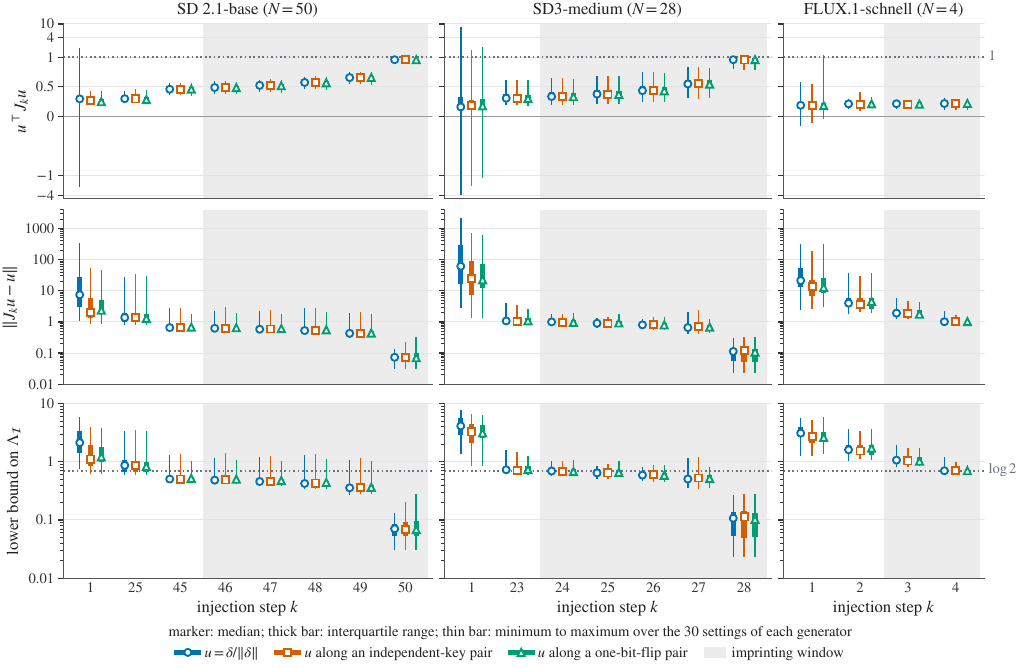}
  \caption{\textbf{Propagation from different injection steps.} For sampled key-difference directions $u$, the projected response $u^\top J_ku$ and deviation $\|J_ku-u\|$ show how perturbations inserted before step $k$ reach the final latent. The third row tests the necessary lower bound $\Lambda_{\mathcal I}\ge\log(1+\|J_ku-u\|)$ against $\log 2$. The shaded region marks the imprinting window used for generation. Steps run to $N=50$, $28$, and $4$ for SD~2.1, SD3, and FLUX, respectively.}
  \label{fig:supp_theory_jacobians}
\end{figure*}

\subsubsection{Protocol}
\label{sec:supp_measured_protocol}
We use the trained encoders and generation settings in \cref{tab:supp_hparams}. SD~2.1 uses 50 DDIM steps, SD3 uses 28 Euler steps, and FLUX uses four distilled steps. Their default imprinting windows cover the final 5, 5, and 2 steps, respectively. For each generator, 100 prompt--seed settings yield 100 independent-key pairs and 100 one-bit-flip pairs. We measure their signed gains at the perturbation amplitude and imprinting weights used for generation (\cref{tab:supp_measured}).

To study window length, we vary $|\mathcal I|$ while holding the total weight $\bar\eta$ fixed for each backbone, using $\eta_k=\bar\eta/|\mathcal I|$ on the last $|\mathcal I|$ steps. We report normalized signed gains for independent-key pairs over 20 prompt--seed settings (\cref{fig:supp_theory_window}). A separate set of 30 settings per generator measures stepwise Jacobian-vector responses along the clean trajectories (\cref{fig:supp_theory_jacobians}).

% Each result uses one of two arithmetic precisions. Signed gains are computed from final latents sampled at generation precision, with the denoiser in fp16 (SD~2.1, SD3) or bf16 (FLUX), so they describe the latents from which images are decoded. The Jacobian responses in \cref{fig:supp_theory_jacobians} are computed by forward-mode differentiation along clean trajectories in fp64 (SD~2.1, SD3) or fp32 (FLUX). \hl{The two precisions produce different trajectories: for the window used for generation, the response $\Phi(\delta)-\Phi(0)$ at generation precision differs from the higher-precision response by a median of 3\%, 39\%, and 77\% of its norm on SD~2.1, SD3, and FLUX.} At that window, the median normalized signed gain of the independent-key pairs differs by at most 0.002 between the two precisions. \hl{On SD3, the two runs also used different GPU types (H100 and A100), whose sampled SD3 trajectories differ.}

\subsubsection{Results}
\label{sec:supp_measured_results}
\textbf{Distinct keys remain separated.}
We evaluate 100 independent-key pairs and 100 one-bit-flip pairs per generator. All 200 pairs have positive signed gains (\cref{tab:supp_measured}), with minima of 0.281, 0.901, and 0.455 for SD~2.1, SD3, and FLUX, respectively. These results confirm the separation condition in \cref{thm:channel}(ii) for all tested pairs.

\textbf{Effect of imprinting-window length.}
Late imprinting better preserves key directions (\cref{fig:supp_theory_window}). At fixed $\bar\eta$, imprinting only at the final step achieves median normalized signed gains $G$ of 0.99, 1.01, and 0.48 on SD~2.1, SD3, and FLUX, respectively. The default five-step windows also maintain positive gains of approximately 0.69 on SD~2.1 and SD3. Importantly, all sampled key pairs retain positive signed gains across all evaluated window lengths (minimum $G=0.17$), confirming their separation at the final latent. These findings are consistent with the window-dependent bound in \cref{thm:channel}(i). The image-quality effects of imprinting placement are examined separately in \cref{sec:injection-window}.

\textbf{Late imprinting better preserves key directions.}
Early-step Jacobians can amplify and rotate perturbations. On SD3, the median magnitude $\|J_1u\|$ is approximately 30, while the projections $u^\top J_ku$ can be positive or negative at early steps on SD~2.1 and SD3. In contrast, the final-step projection is approximately 0.96 on both generators (\cref{fig:supp_theory_jacobians}). These findings suggest that imprinting closer to the final step better preserves the original direction of the key perturbation.

% =====================================================================
\section{Implementation Details}
\label{sec:supp_impl}

\subsection{Generation and extraction settings}
\label{sec:supp_inference_settings}
\cref{tab:supp_hparams} summarizes the settings for each backbone. All images are $512\times512$, and SD3 is run without T5. The denoiser, VAE, and text encoders remain frozen throughout training. During generation, $\delta=a\mu$ is computed once per image and imprinted with $\eta_k=\bar\eta/|\mathcal I|$ for $k\in\mathcal I$ and $\eta_k=0$ otherwise; \cref{eq:subcarrier} is applied once after step $N$, and \cref{eq:supp_correction} uses $(\tau_c,\gamma_c)=(1,1.5)$.

\begin{table}[ht]
  \centering\footnotesize
  \caption{\textbf{Settings per generator.} $\mathcal I$ and $\eta_k$ apply to generation only.}
  \label{tab:supp_hparams}
  \resizebox{\linewidth}{!}{%
  \begin{tabular}{@{}lccc@{}}
    \toprule
    & SD~2.1-base & SD3-medium & FLUX.1-schnell \\
    \midrule
    $C\times h\times w$ & $4\times64\times64$ & $16\times64\times64$ & $16\times64\times64$ \\
    Sampler & DDIM & Flow Euler & Distilled \\
    $N$ / guidance $w$ & 50 / 7.5 & 28 / 7.0 & 4 / none \\
    $|\mathcal I|$ (last steps) & 5 & 5 & 2 \\
    $\bar\eta$ & 0.5 & 2.0 & 3.0 \\
    $\eta_k$, $k\in\mathcal I$ & 0.1 & 0.4 & 1.5 \\
    $\sigma^\star$ & 0.40 & 0.06 & 0.06 \\
    $[p_{\rm lo},p_{\rm hi}]$ (latent pixels) & $[6,16]$ & $[6,16]$ & $[6,16]$ \\
    $n_c$ & 1 & 2 & 1 \\
    $c_0$ & 8 & 16 & 16 \\
    $\beta_{\rm sc}$ & 0.15 & 0.1875 & 0.15 \\
    $R$ & 4 & 4 & 12 \\
    $K$ & 2 & 2 & 4 \\
    $n_{\rm img}$ (pattern statistics) & 96 & 96 & 96 \\
    \bottomrule
  \end{tabular}}
\end{table}

\textbf{Initialization and modes.} $b_{\ell,v}\sim\mathcal N(0,0.02^2)$. The $\log\sigma^2$ head has zero weights and bias $-2$, with output clamped to $[-10,2]$. $B$ is resized bilinearly when $h,w\neq64$. $\delta=a(\mu+\sigma\odot\epsilon)$ only in Stage~1 and $\delta=a\mu$ afterwards. From Stage~3, the BatchNorm layers of $E_\phi$ use running statistics.

\textbf{Pattern construction settings.} $S$, $G$, $G_{\rm clean}$ and $D_{\rm img}$ are estimated from $n_{\rm img}=96$ images per generator, generated by the Stage~4 checkpoint with the generation rule of \cref{tab:supp_hparams}. Stage~4 trains with provisional patterns, $[p_{\rm lo},p_{\rm hi}]=[8,16]$ and $n_c=C$, and learns their $H_\ell$. After checkpoint selection, both are replaced by $P_\ell$ and $H_\ell$ from \cref{eq:supp_pattern_design}, without further training.

\subsection{Alignment search settings}
\label{sec:supp_alignment_settings}
The grid $\mathcal T$ contains the identity $I$, rotations $\theta\in\{-44^\circ,-43^\circ,\dots,44^\circ\}$, and crops with side ratio $\varsigma\in\{0.98,0.96,\dots,0.72\}$ and center offsets in steps of $1/64$ (half a latent pixel). For the coarse latent search, we compute $\operatorname{Agr}_T=\operatorname{Agr}(T(z))$ using bilinear warping with zero padding. Let $\operatorname{med}_{\rm Agr}=\operatorname{median}_T\operatorname{Agr}_T$ and $\operatorname{mad}_{\rm Agr}=1.4826\operatorname{median}_T|\operatorname{Agr}_T-\operatorname{med}_{\rm Agr}|$.

\textbf{SD~2.1.} If $\arg\max_T\operatorname{Agr}_T=I$, we retain the original alignment. Otherwise, we select up to 8 distinct candidates from the top 256. Candidates must differ by $\ge2^\circ$ in $\theta$, $\ge0.03$ in $\varsigma$, or $\ge4/64$ in center position. We score each candidate using the agreement between the two outputs of $\operatorname{ImageMargins}(x,T)$. We refine the best candidate with $\theta$ offsets $\{0,\pm0.25,\pm0.5,\pm0.75\}^\circ$, or $\varsigma$ offsets $\{-0.01,0,0.01\}$ and corner offsets $\{-4,-2,0,2,4\}^2$ pixels. We accept the refined $\hat T$ if $(\operatorname{Agr}(\hat T)-\operatorname{med}_{\rm Agr})/\operatorname{mad}_{\rm Agr}\ge z_{\min}$ and $L^{-1}\sum_\ell|o^{\rm dec}_\ell|\ge c_{\min}$, with $(z_{\min},c_{\min})=(10,1.4)$. Otherwise, we retain $I$.

\textbf{SD3 and FLUX.} We score all candidates using $\operatorname{ImageMargins}$. If $\operatorname{Agr}(I)\ge0.12$, we retain the original alignment without further search. Otherwise, we accept $\hat T$ only if all four checks pass: $(\operatorname{Agr}(\hat T)-\operatorname{med}_{\rm Agr})/\operatorname{mad}_{\rm Agr}\ge z_{\min}$, $\operatorname{Agr}(\hat T)-\operatorname{Agr}(I)\ge3\operatorname{mad}_{\rm Agr}$, $\operatorname{Agr}(\hat T)\ge0.14$, and $L^{-1}\sum_\ell|o^{\rm dec}_\ell|\ge c_{\min}$. We use $(z_{\min},c_{\min})=(6,1.0)$ and retain $I$ if any check fails.

\subsection{Training schedule and optimization}
\label{sec:implement_training}

We train on 10,000 DiffusionDB prompts using the dual-path framework, with both paths sharing $(s,c,z_T)$. The short path imprints $\delta/K$ per step, while the full path imprints $\sg(\delta)/N$ without gradient tracking. \cref{tab:supp_schedule} summarizes the training schedule.

\begin{table}[t]
  \centering\footnotesize
  \caption{\textbf{Training schedule.} Learning rates are $E_\phi$ / $D_\psi$; a dash indicates a frozen network.}
  \label{tab:supp_schedule}
  \resizebox{\linewidth}{!}{%
  \begin{tabular}{@{}lccc@{}}
    \toprule
    Stage & Steps & Global batch & Learning rates \\
    \midrule
    1 & 2,000 & 64 & $3\times10^{-4}$ / -- \\
    2a & stopping rule (1,041) & 64 & $3\times10^{-4}$ / $10^{-4}$ \\
    2b & $\le$ 30,000 & 64 & -- / $10^{-4}$ \\
    3 (SD~2.1) & 7,500 & 16 & $5\times10^{-5}$ / $8.5\times10^{-4}$ \\
    4 (SD~2.1) & 1,000 & 8 & $2\times10^{-5}$ / $10^{-4}$ \\
    4 (SD3, FLUX) & 1,000 & 8 & $5\times10^{-5}$ / $2\times10^{-4}$ \\
    \bottomrule
  \end{tabular}}
\end{table}

\textbf{Training stages.}
Stage~1 initializes the watermark encoder with magnitude and KL regularization. Stage~2 pretrains the decoder on clean and noise-perturbed watermark latents, with the encoder frozen in Stage~2b. Stage~3 introduces dual-path training with image-fidelity regularization, without frequency-based patterns or attacks. Stage~4 incorporates provisional frequency-based patterns and attack augmentation to improve robustness. SD3 and FLUX are initialized from an SD~1.5 Stage~3 checkpoint, with channel adaptation and perturbation-scale adjustment.

\textbf{Optimization and losses.}
We use AdamW with cosine learning-rate decay and gradient clipping. For \cref{eq:reconstruction}, we set $\tau_{\rm m}=2$, $w_{\rm m}=0.25$, $r=0.25$, $w_{\rm tail}=0.5$, and $\lambda_L=\sqrt{2}$. The bit weights $\omega_\ell\in[1,3]$ are updated using an exponential moving average (momentum 0.98). The magnitude loss follows \cref{eq:supp_magnitude_loss} in Stage~4, and regional pixel fidelity uses $33\times33$ windows. Other loss coefficients are provided in \cref{tab:supp_loss_weights}.

\begin{table}[t]
  \centering\footnotesize
  \caption{\textbf{Loss coefficients.} $\beta_t$ rises linearly from 0.001 to 0.05 over 625 steps. $\mathcal L_{\rm sh}$, $\mathcal L_{\rm full}$, the round trip and $\mathcal L_{\rm atk}$ have weight 1.}
  \label{tab:supp_loss_weights}
  \begin{tabular}{@{}lcc@{}}
    \toprule
    Coefficient & Stage 3 & Stage 4 \\
    \midrule
    $w_{\rm mag}$ & 10 & 25 \\
    $\beta_t$ & $0.001\to0.05$ & $0.001\to0.05$ \\
    $w_{\rm div}$ & 0.1 & 0.1 \\
    $w_{\rm prvl}$ & 1.5 & 1.5 \\
    $w_{\rm ld}$ & 0.1 & 2.0 \\
    $w_{\rm neg}$ & 0.01 & 0.01 \\
    $w_{\rm lf}$ & 0.5 & 0 \\
    \bottomrule
  \end{tabular}
\end{table}

% \textbf{Training attacks.} For each sample, $\mathcal A$ is drawn as follows and applied to both paths, with strength $q_t$ rising from 0.3 to 1 over 400 steps and $w_{\rm atk}(t)$ from 0 to 1 over 200 steps.
% \begin{itemize}[leftmargin=*,nosep]
%   \item Geometric ($p=0.35$): rotation $|\theta|\le45^\circ q_t$ or a square crop keeping area in $[1-0.5q_t,1]$, then the inverse transform with residual error up to $0.5^\circ$, 1.2\% scale and $1/4$ latent pixel.
%   \item VAE embedding ($p=0.15$): 5 PGD steps, $\|\pi\|_\infty\le8/255$, maximizing $\|\mathcal E(x+\pi)-\mathcal E(x)\|_2^2$; $\pi$ is held fixed in the backward pass.
%   \item Pixel ($p=0.50$): one of JPEG (quality $90-60q_t$), Gaussian blur ($\sigma\sim\mathcal U[1,2+10q_t]$ pixels), Gaussian noise (std $0.01+0.15q_t$), color jitter (brightness, contrast, saturation $\pm(0.1+0.3q_t)$). Differentiable on the short path; on the full path, JPEG is real compression with probability 0.5.
% \end{itemize}

\subsection{Computing resources}
All training ran on a single node with 4× NVIDIA A100 (40 GB) GPUs and 2× AMD EPYC 7452 CPUs (64 cores). Training DiffMark on SD 2.1-base took about 28 node-hours ($\approx$ 112 GPU-hours), including 24.8 h for the main training stage. Adapting the SD 2.1 weights to SD3-medium and FLUX.1 took 2.5 h and 3.1 h of fine-tuning, respectively.

% =====================================================================
\section{Attack Protocol}
\label{sec:supp_attacks}
We follow WAVES~\cite{anwaves}. Distortion attacks (\cref{tab:supp_distortions}) include geometric transformations (rotation, resized crop, and random erasing), photometric changes (brightness, contrast, and Gaussian noise), and filtering or compression (Gaussian blur and JPEG). Removal attacks (\cref{tab:supp_regen}) reconstruct the image with a generative model. \texttt{Regen-VAE} uses a compression VAE, \texttt{Regen-Diff} adds noise in the SD-v1.4 latent space and then denoises, and \texttt{Rinse-2xDiff} repeats diffusion regeneration twice. Adversarial attacks use projected gradient descent (PGD) to maximize the $\ell_2$ distance between the embeddings of the attacked and original images, subject to an $\ell_\infty$ perturbation bound. The embedding model is either the target model's KL-VAE (grey-box, \texttt{Adv-KLVAE8}) or a ResNet-18 (black-box, \texttt{Adv-RN18}).

\begin{table}[t]
  \centering\footnotesize
  \setlength{\tabcolsep}{4pt}
  \caption{\textbf{Distortion attacks.} Strength specifies the tested attack severity. Rotation is measured in degrees, crop scale denotes the retained side-length ratio, and erasing specifies the removed image area. Brightness and contrast use multiplicative factors, blur uses kernel size, noise uses standard deviation $\sigma$, and JPEG uses compression quality. Arrows indicate increasing severity toward the specified value.}
  \label{tab:supp_distortions}
  \begin{tabular}{@{}llll@{}}
    \toprule
    Attack & ID & Mechanism & Strength \\
    \midrule
    Rotation & \texttt{Rotation} & rotate by angle & $45^\circ$ \\
    Resized crop & \texttt{RCrop} & crop and resize & scale $\to0.5$ \\
    Random erasing & \texttt{Erase} & zero-fill a patch & $25\%$ area \\
    Brightness & \texttt{Bright} & enhance brightness & factor $\to2.0$ \\
    Contrast & \texttt{Contrast} & enhance contrast & factor $\to2.0$ \\
    Gaussian blur & \texttt{Blur} & blur kernel & size $0\to20$ \\
    Gaussian noise & \texttt{Noise} & $\mathcal N(0,\sigma^2)$ & $\sigma$: $0\to0.1$ \\
    JPEG & \texttt{Compress} & lossy re-encoding & quality $50\to10$ \\
    \bottomrule
  \end{tabular}
\end{table}

\begin{table}[t]
  \centering\footnotesize
  \caption{\textbf{Regeneration and adversarial attacks.}}
  \label{tab:supp_regen}
  \begin{tabular}{@{}lll@{}}
    \toprule
    Attack & ID & Strength \\
    \midrule
    VAE & \texttt{Regen-VAE} & quality 7 \\
    Diffusion & \texttt{Regen-Diff} & 200 steps \\
    Rinsing ($2\times$) & \texttt{Rinse-2xDiff} & 100 steps \\
    KL-VAE (grey-box) & \texttt{Adv-KLVAE8} & $\epsilon=8/255$ \\
    ResNet-18 (black-box) & \texttt{Adv-RN18} & $\epsilon=8/255$ \\
    \bottomrule
  \end{tabular}
\end{table}

\textbf{Stronger removal attacks.} We further evaluate LFM~\cite{aistov2026exploring} and UnMarker~\cite{kassis2025unmarker} on SD~2.1 with DiffusionDB (1,000 images). \cref{tab:supp_strong} lists the settings. We recover watermarks from attacked images using the complete extraction pipeline, including alignment search, and measure image quality against the corresponding unwatermarked images.

Latent Frequency Masking (LFM)~\cite{aistov2026exploring} encodes the watermarked image with a VAE and replaces selected low-frequency Fourier coefficients in specified latent channels before decoding the modified latent. The Gaussian variant tested here uses replacement coefficients obtained from Gaussian noise, disrupting watermark information while retaining the unmasked components.
UnMarker~\cite{kassis2025unmarker} combines cropping with two optimization stages that target high- and low-frequency components of the image spectrum. These stages optimize pixel perturbations and spatial filters under perceptual similarity constraints to suppress watermark evidence without requiring detector feedback.

\begin{table}[t]
  \centering\footnotesize
  \setlength{\tabcolsep}{4pt}
  \caption{\textbf{Stronger removal attacks.}}
  \label{tab:supp_strong}
  \resizebox{\linewidth}{!}{%
  \begin{tabular}{@{}lll@{}}
    \toprule
    Attack & ID & Setting \\
    \midrule
    LFM, Gaussian & \texttt{LFM-Gauss} & $r=15$, $\mathcal C=\{1,3\}$, $\beta\in\{0,0.2,\dots,0.8\}$ \\
    UnMarker, full & \texttt{UnMarker} & crop $0.9$, Stage~1, Stage~2 \\
    UnMarker, crop + low freq. & \texttt{UnMarker-CL} & crop $0.9$, Stage~2 \\
    \bottomrule
  \end{tabular}}
\end{table}

% =====================================================================
\section{Additional Results}
\label{sec:supp_results}

% \subsection{Robustness to stronger removal attacks}
% \label{sec:supp_strong_removal}

% \begin{figure}[ht]
%     \centering
%     \includegraphics[width=\columnwidth]{figs/design_attack_cost_detection.pdf}
%     \caption{\textbf{Robustness to stronger removal attacks on SD~2.1 with DiffusionDB.} Detection TPR@1\%FPR and bit accuracy after LFM at its strongest tested setting and two UnMarker variants.}
%     \label{fig:supp_strong_removal}
% \end{figure}

% We further evaluate \ours{} against LFM and UnMarker on SD~2.1 with DiffusionDB. For LFM, we use the strongest tested setting: Gaussian, radius 15, channels 1 and 3, and $\beta=0$. For UnMarker, we evaluate both the full attack and cropping followed by its low-frequency stage.

% \cref{fig:supp_strong_removal} shows that \ours{} retains high detection rates under both attacks. TPR@1\%FPR reaches $99.0\%$ after LFM and $99.9\%$ under both UnMarker variants. Bit accuracy is $97.5\%$ after LFM, compared with $94.2\%$ and $93.9\%$ after the full and crop-plus-low-frequency UnMarker attacks, respectively. These results show that detection remains robust despite the bit errors introduced by the tested removal attacks.

\subsection{Decoder calibration and gradient alignment} 
\label{sec:gradient-alignment}

\begin{table}[t]
  \centering
  \small
  \caption{\textbf{Gradient alignment on SD~2.1.} Cosines between two-step DDIM short-path and DDIM-50 full-path loss gradients. Values are means [95\% bootstrap confidence intervals] over 500 paired samples at each checkpoint.}
  \label{tab:sd21-gradient-alignment}
  \resizebox{0.9\columnwidth}{!}{%
    \begin{tabular}{lcc}
      \toprule
      \textbf{Training step}
      & \shortstack{\textbf{Mean gradient cosine}\\\textbf{w.r.t. $\delta$}}
      & \shortstack{\textbf{Mean gradient cosine}\\\textbf{w.r.t. $\phi$}} \\
      \midrule
      3,000 & 0.706 [0.630, 0.774] & 0.946 [0.911, 0.973] \\
      4,000 & 0.657 [0.570, 0.733] & 0.931 [0.892, 0.962] \\
      5,000 & 0.721 [0.642, 0.790] & 0.943 [0.894, 0.976] \\
      \bottomrule
    \end{tabular}%
  }
\end{table}

The short path supplies gradients to the encoder, while full-path latents and their attacked versions train the decoder. We compare short- and full-path decoding-loss gradients on 500 paired SD~2.1 samples at each of three training checkpoints (\cref{tab:sd21-gradient-alignment}). The full-path gradients serve only as a reference for this diagnostic.

\cref{thm:surrogate} gives sufficient conditions for the negative short-path gradient to decrease the full-path loss locally. For an individual sample, a positive cosine between the two gradients establishes this descent direction. The measured mean cosines are positive at every checkpoint: 0.66--0.72 with respect to $\delta$ and 0.93--0.95 with respect to the encoder parameters $\phi$. All reported confidence intervals for these means exclude zero. These measurements support positive gradient alignment on average, especially for the encoder parameters. 

\begin{figure}[ht]
  \centering
  \includegraphics[width=\columnwidth]{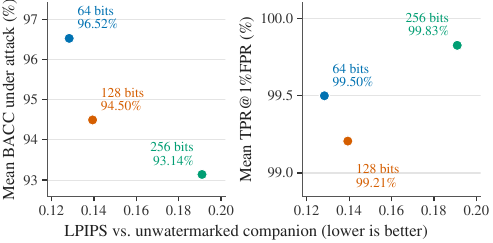}
  \caption{\textbf{Payload length, image quality, and robustness.}
  Shorter payloads have lower LPIPS to clean companions and higher mean bit accuracy after attacks.}
  \label{fig:payload-tradeoff}
\end{figure}

\subsection{Payload length trade-off}
\label{sec:supp-payload}

Shorter payloads improve image quality (\cref{fig:payload-tradeoff}). Reducing the payload from 256 to 64 bits lowers clean-companion LPIPS from approximately 0.19 to 0.13 and raises mean attacked bit accuracy from 93.14\% to 96.52\%. The 128-bit setting lies between them on both measures. Detection remains high across all lengths, with mean TPR@1\%FPR of 99.50\%, 99.21\%, and 99.83\% for 64, 128, and 256 bits, respectively. Thus, shorter keys offer better perceptual fidelity and
bit recovery, while the 256-bit setting provides greater payload capacity and the highest detection rate in this experiment.

\begin{table}[ht]
    \centering
    \caption{\textbf{Imprinting-placement ablation on DiffusionDB using the
    same SD~2.1 checkpoint}. PSNR and LPIPS are measured against
    unwatermarked images generated with the same seed. Minimum TPR is taken over
    clean decoding and all 15 attacks.}
    \label{tab:injection-window}
    \resizebox{0.8\columnwidth}{!}{%
        \begin{tabular}{@{}lrrr@{}}
            \toprule
            \textbf{Imprinting setting}
            & \textbf{PSNR}$\uparrow$
            & \textbf{LPIPS}$\downarrow$
            & \shortstack{\textbf{TPR@1}\%\textbf{FPR}} \\
            \midrule
            Initial noise only
            & 17.63 & 0.585 & 100 \\
            Every step
            & 22.40 & 0.364 & 100 \\
            Last five steps
            & 26.05 & 0.195 & 100 \\
            \bottomrule
        \end{tabular}%
    }
\end{table}

\subsection{Imprinting window} 
\label{sec:injection-window}

This separate image-level ablation compares imprinting placements using the same SD~2.1 checkpoint (\cref{tab:injection-window}). Relative to imprinting at every step, imprinting during the final five steps improves PSNR from $22.40$ to $26.05$\,dB and reduces LPIPS from $0.364$ to $0.195$, while both retain $100\%$ minimum TPR. Unlike the latent-propagation analysis in \cref{fig:supp_theory_window}, this ablation evaluates image quality and detection.

\begin{figure}[ht]
    \centering
    \includegraphics[width=\linewidth]{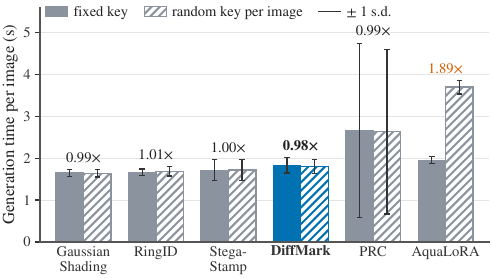}
    \caption{\textbf{Per-image key flexibility on SD~2.1.} Generation time with a fixed key or a random key per image; annotations give the random/fixed ratio.}
    \label{fig:key-flexibility}
    \vspace{-8pt}
\end{figure}

\subsection{Key flexibility}
\label{sec:key-flexibility}

Once trained for a backbone, \ours supports arbitrary per-image keys through one encoder call without retraining. In contrast, Stable Signature~\cite{fernandez2023stable} binds a fixed key to its decoder weights, so changing keys requires updating or replacing watermark-specific weights. In \cref{fig:key-flexibility}, \ours's random/fixed generation-time ratio is $0.98\times$, showing comparable runtime when keys change, whereas AquaLoRA's is $1.89\times$.

\begin{figure}[ht]
    \centering
    \includegraphics[width=1\linewidth]{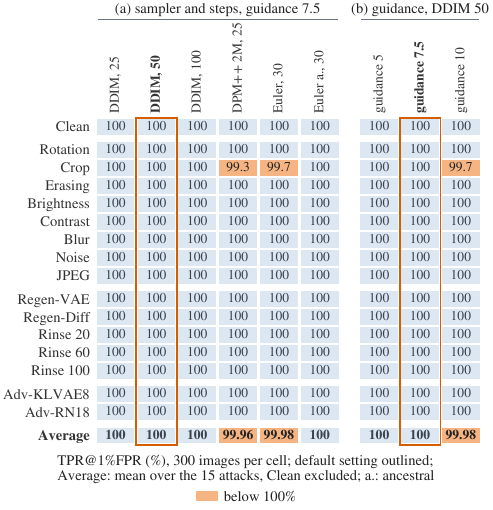}
    \caption{\textbf{Robustness across sampling settings on SD~2.1.} TPR at $1\%$ FPR by attack for (a) different samplers and step counts at guidance 7.5 and (b) different guidance scales with 50-step DDIM. Averages exclude clean images.}
    \label{fig:robust-sampler-guidance}
    \vspace{-0.5cm}
\end{figure}

\subsection{Robustness across sampling and guidance settings}
\label{sec:robust-sample-guidance}

On SD~2.1, we compare DDIM, DPM++ 2M, Euler, and Euler ancestral at guidance 7.5, vary DDIM from 25 to 100 steps, and test guidance scales 5, 7.5, and 10 with 50-step DDIM (\cref{fig:robust-sampler-guidance}). \ours maintains $99.3$--$100\%$ TPR at $1\%$ FPR across all evaluated attacks and settings, with minor drops confined to cropping. Detection is stable across these sampling choices: DDIM already achieves perfect detection with 25 steps, and additional steps provide no detection gain in this evaluation.

\begin{table*}[ht]
    \centering
    \caption{\textbf{Distortion breakdown with SD~2.1 on DiffusionDB, MS-COCO, and DALL-E3.} Values are mean TPR@1\%FPR, with standard deviations in subscript. Each cell lists DiffusionDB, MS-COCO, and DALL-E3 results from top to bottom. The best result for each attack and dataset, including ties, is in \textbf{bold}. Avg is the reported average over the eight distortions.}
    \label{tab:supp_distortion_sd21}
    \resizebox{\textwidth}{!}{%
    \footnotesize
    \setlength{\tabcolsep}{4pt}
    \renewcommand{\arraystretch}{1.05}
    \begin{tabular}{l ccccccccc}
    \toprule
    \textbf{Method} & \textbf{Rotation} & \textbf{Crop} & \textbf{Erasing} & \textbf{Brightness} & \textbf{Contrast} & \textbf{Blur} & \textbf{Noise} & \textbf{Compress} & \textbf{Avg} \\
    \cmidrule(lr){1-1} \cmidrule(lr){2-2} \cmidrule(lr){3-3} \cmidrule(lr){4-4} \cmidrule(lr){5-5} \cmidrule(lr){6-6} \cmidrule(lr){7-7} \cmidrule(lr){8-8} \cmidrule(lr){9-9} \cmidrule(lr){10-10}
    \rowcolor{rowgray}
    Tree-Ring
    & \makecell{0.127$_{\pm 0.027}$ \\ 0.119$_{\pm 0.041}$ \\ 0.109$_{\pm 0.045}$}
    & \makecell{0.100$_{\pm 0.021}$ \\ 0.077$_{\pm 0.028}$ \\ 0.083$_{\pm 0.026}$}
    & \makecell{\textbf{1.000}$_{\pm 0.000}$ \\ \textbf{1.000}$_{\pm 0.000}$ \\ \textbf{1.000}$_{\pm 0.000}$}
    & \makecell{\textbf{1.000}$_{\pm 0.000}$ \\ \textbf{1.000}$_{\pm 0.001}$ \\ 0.999$_{\pm 0.001}$}
    & \makecell{\textbf{1.000}$_{\pm 0.000}$ \\ \textbf{1.000}$_{\pm 0.000}$ \\ \textbf{1.000}$_{\pm 0.000}$}
    & \makecell{0.999$_{\pm 0.001}$ \\ 0.999$_{\pm 0.001}$ \\ 0.997$_{\pm 0.003}$}
    & \makecell{\textbf{1.000}$_{\pm 0.000}$ \\ \textbf{1.000}$_{\pm 0.000}$ \\ 0.999$_{\pm 0.001}$}
    & \makecell{\textbf{1.000}$_{\pm 0.000}$ \\ \textbf{1.000}$_{\pm 0.000}$ \\ 0.998$_{\pm 0.001}$}
    & \makecell{0.778$_{\pm 0.006}$ \\ 0.774$_{\pm 0.009}$ \\ 0.773$_{\pm 0.009}$} \\
    Shallow-Diffuse
    & \makecell{0.002$_{\pm 0.003}$ \\ 0.005$_{\pm 0.004}$ \\ 0.004$_{\pm 0.004}$}
    & \makecell{0.009$_{\pm 0.004}$ \\ 0.011$_{\pm 0.008}$ \\ 0.008$_{\pm 0.005}$}
    & \makecell{\textbf{1.000}$_{\pm 0.000}$ \\ \textbf{1.000}$_{\pm 0.000}$ \\ \textbf{1.000}$_{\pm 0.000}$}
    & \makecell{\textbf{1.000}$_{\pm 0.000}$ \\ \textbf{1.000}$_{\pm 0.000}$ \\ \textbf{1.000}$_{\pm 0.000}$}
    & \makecell{\textbf{1.000}$_{\pm 0.000}$ \\ \textbf{1.000}$_{\pm 0.000}$ \\ \textbf{1.000}$_{\pm 0.000}$}
    & \makecell{0.013$_{\pm 0.007}$ \\ 0.015$_{\pm 0.005}$ \\ 0.013$_{\pm 0.008}$}
    & \makecell{\textbf{1.000}$_{\pm 0.000}$ \\ \textbf{1.000}$_{\pm 0.000}$ \\ 0.999$_{\pm 0.001}$}
    & \makecell{\textbf{1.000}$_{\pm 0.000}$ \\ \textbf{1.000}$_{\pm 0.000}$ \\ \textbf{1.000}$_{\pm 0.000}$}
    & \makecell{0.628$_{\pm 0.002}$ \\ 0.629$_{\pm 0.002}$ \\ 0.628$_{\pm 0.002}$} \\
    \rowcolor{rowgray}
    FARI-T
    & \makecell{0.010$_{\pm 0.006}$ \\ 0.019$_{\pm 0.009}$ \\ 0.014$_{\pm 0.012}$}
    & \makecell{0.072$_{\pm 0.017}$ \\ 0.059$_{\pm 0.021}$ \\ 0.062$_{\pm 0.019}$}
    & \makecell{\textbf{1.000}$_{\pm 0.000}$ \\ \textbf{1.000}$_{\pm 0.000}$ \\ \textbf{1.000}$_{\pm 0.000}$}
    & \makecell{\textbf{1.000}$_{\pm 0.000}$ \\ \textbf{1.000}$_{\pm 0.000}$ \\ \textbf{1.000}$_{\pm 0.000}$}
    & \makecell{\textbf{1.000}$_{\pm 0.000}$ \\ \textbf{1.000}$_{\pm 0.000}$ \\ \textbf{1.000}$_{\pm 0.000}$}
    & \makecell{\textbf{1.000}$_{\pm 0.000}$ \\ \textbf{1.000}$_{\pm 0.000}$ \\ 0.999$_{\pm 0.001}$}
    & \makecell{\textbf{1.000}$_{\pm 0.000}$ \\ \textbf{1.000}$_{\pm 0.000}$ \\ \textbf{1.000}$_{\pm 0.000}$}
    & \makecell{\textbf{1.000}$_{\pm 0.000}$ \\ \textbf{1.000}$_{\pm 0.000}$ \\ \textbf{1.000}$_{\pm 0.000}$}
    & \makecell{0.760$_{\pm 0.003}$ \\ 0.760$_{\pm 0.004}$ \\ 0.759$_{\pm 0.004}$} \\
    \midrule
    StegaStamp
    & \makecell{0.006$_{\pm 0.009}$ \\ 0.011$_{\pm 0.004}$ \\ 0.011$_{\pm 0.005}$}
    & \makecell{0.437$_{\pm 0.028}$ \\ 0.433$_{\pm 0.018}$ \\ 0.409$_{\pm 0.024}$}
    & \makecell{\textbf{1.000}$_{\pm 0.000}$ \\ \textbf{1.000}$_{\pm 0.000}$ \\ \textbf{1.000}$_{\pm 0.000}$}
    & \makecell{\textbf{1.000}$_{\pm 0.000}$ \\ \textbf{1.000}$_{\pm 0.000}$ \\ 0.997$_{\pm 0.002}$}
    & \makecell{\textbf{1.000}$_{\pm 0.000}$ \\ \textbf{1.000}$_{\pm 0.000}$ \\ \textbf{1.000}$_{\pm 0.000}$}
    & \makecell{0.624$_{\pm 0.052}$ \\ 0.486$_{\pm 0.031}$ \\ 0.296$_{\pm 0.041}$}
    & \makecell{\textbf{1.000}$_{\pm 0.000}$ \\ \textbf{1.000}$_{\pm 0.000}$ \\ \textbf{1.000}$_{\pm 0.000}$}
    & \makecell{\textbf{1.000}$_{\pm 0.000}$ \\ \textbf{1.000}$_{\pm 0.000}$ \\ \textbf{1.000}$_{\pm 0.000}$}
    & \makecell{0.758$_{\pm 0.010}$ \\ 0.741$_{\pm 0.005}$ \\ 0.714$_{\pm 0.008}$} \\
    \rowcolor{rowgray}
    RingID
    & \makecell{\textbf{1.000}$_{\pm 0.000}$ \\ \textbf{1.000}$_{\pm 0.000}$ \\ \textbf{1.000}$_{\pm 0.000}$}
    & \makecell{0.067$_{\pm 0.024}$ \\ 0.093$_{\pm 0.024}$ \\ 0.056$_{\pm 0.021}$}
    & \makecell{\textbf{1.000}$_{\pm 0.000}$ \\ \textbf{1.000}$_{\pm 0.000}$ \\ \textbf{1.000}$_{\pm 0.000}$}
    & \makecell{\textbf{1.000}$_{\pm 0.000}$ \\ \textbf{1.000}$_{\pm 0.000}$ \\ \textbf{1.000}$_{\pm 0.000}$}
    & \makecell{\textbf{1.000}$_{\pm 0.000}$ \\ \textbf{1.000}$_{\pm 0.000}$ \\ \textbf{1.000}$_{\pm 0.000}$}
    & \makecell{\textbf{1.000}$_{\pm 0.000}$ \\ \textbf{1.000}$_{\pm 0.001}$ \\ \textbf{1.000}$_{\pm 0.000}$}
    & \makecell{\textbf{1.000}$_{\pm 0.000}$ \\ \textbf{1.000}$_{\pm 0.000}$ \\ \textbf{1.000}$_{\pm 0.000}$}
    & \makecell{\textbf{1.000}$_{\pm 0.000}$ \\ \textbf{1.000}$_{\pm 0.000}$ \\ \textbf{1.000}$_{\pm 0.000}$}
    & \makecell{0.883$_{\pm 0.003}$ \\ 0.887$_{\pm 0.003}$ \\ 0.882$_{\pm 0.003}$} \\
    StableSignature
    & \makecell{0.942$_{\pm 0.016}$ \\ 0.986$_{\pm 0.009}$ \\ 0.940$_{\pm 0.022}$}
    & \makecell{\textbf{0.999}$_{\pm 0.001}$ \\ \textbf{1.000}$_{\pm 0.000}$ \\ 0.996$_{\pm 0.002}$}
    & \makecell{\textbf{1.000}$_{\pm 0.000}$ \\ \textbf{1.000}$_{\pm 0.000}$ \\ 0.999$_{\pm 0.001}$}
    & \makecell{\textbf{1.000}$_{\pm 0.000}$ \\ \textbf{1.000}$_{\pm 0.000}$ \\ \textbf{1.000}$_{\pm 0.000}$}
    & \makecell{0.999$_{\pm 0.001}$ \\ \textbf{1.000}$_{\pm 0.000}$ \\ \textbf{1.000}$_{\pm 0.000}$}
    & \makecell{0.000$_{\pm 0.000}$ \\ 0.000$_{\pm 0.000}$ \\ 0.000$_{\pm 0.000}$}
    & \makecell{0.995$_{\pm 0.002}$ \\ 0.999$_{\pm 0.001}$ \\ 0.990$_{\pm 0.003}$}
    & \makecell{0.993$_{\pm 0.003}$ \\ 0.997$_{\pm 0.002}$ \\ 0.984$_{\pm 0.004}$}
    & \makecell{0.866$_{\pm 0.002}$ \\ 0.873$_{\pm 0.001}$ \\ 0.864$_{\pm 0.003}$} \\
    \rowcolor{rowgray}
    AquaLoRA
    & \makecell{0.000$_{\pm 0.000}$ \\ 0.000$_{\pm 0.001}$ \\ 0.000$_{\pm 0.001}$}
    & \makecell{0.945$_{\pm 0.008}$ \\ 0.939$_{\pm 0.013}$ \\ 0.922$_{\pm 0.016}$}
    & \makecell{0.979$_{\pm 0.005}$ \\ 0.972$_{\pm 0.008}$ \\ 0.945$_{\pm 0.011}$}
    & \makecell{0.966$_{\pm 0.006}$ \\ 0.941$_{\pm 0.010}$ \\ 0.932$_{\pm 0.011}$}
    & \makecell{0.975$_{\pm 0.006}$ \\ 0.932$_{\pm 0.013}$ \\ 0.925$_{\pm 0.011}$}
    & \makecell{0.968$_{\pm 0.006}$ \\ 0.940$_{\pm 0.010}$ \\ 0.910$_{\pm 0.013}$}
    & \makecell{0.989$_{\pm 0.003}$ \\ 0.988$_{\pm 0.004}$ \\ 0.964$_{\pm 0.006}$}
    & \makecell{0.992$_{\pm 0.003}$ \\ 0.985$_{\pm 0.005}$ \\ 0.962$_{\pm 0.008}$}
    & \makecell{0.852$_{\pm 0.003}$ \\ 0.837$_{\pm 0.007}$ \\ 0.820$_{\pm 0.009}$} \\
    GaussianShading
    & \makecell{0.007$_{\pm 0.006}$ \\ 0.009$_{\pm 0.004}$ \\ 0.003$_{\pm 0.003}$}
    & \makecell{0.049$_{\pm 0.019}$ \\ 0.051$_{\pm 0.010}$ \\ 0.029$_{\pm 0.013}$}
    & \makecell{\textbf{1.000}$_{\pm 0.000}$ \\ \textbf{1.000}$_{\pm 0.000}$ \\ \textbf{1.000}$_{\pm 0.000}$}
    & \makecell{\textbf{1.000}$_{\pm 0.000}$ \\ \textbf{1.000}$_{\pm 0.000}$ \\ \textbf{1.000}$_{\pm 0.000}$}
    & \makecell{\textbf{1.000}$_{\pm 0.000}$ \\ \textbf{1.000}$_{\pm 0.000}$ \\ \textbf{1.000}$_{\pm 0.000}$}
    & \makecell{\textbf{1.000}$_{\pm 0.000}$ \\ 0.996$_{\pm 0.002}$ \\ 0.981$_{\pm 0.006}$}
    & \makecell{\textbf{1.000}$_{\pm 0.000}$ \\ \textbf{1.000}$_{\pm 0.000}$ \\ \textbf{1.000}$_{\pm 0.000}$}
    & \makecell{\textbf{1.000}$_{\pm 0.000}$ \\ \textbf{1.000}$_{\pm 0.000}$ \\ \textbf{1.000}$_{\pm 0.000}$}
    & \makecell{0.757$_{\pm 0.003}$ \\ 0.757$_{\pm 0.002}$ \\ 0.752$_{\pm 0.002}$} \\
    \rowcolor{rowgray}
    FARI-GS
    & \makecell{0.007$_{\pm 0.004}$ \\ 0.013$_{\pm 0.004}$ \\ 0.008$_{\pm 0.004}$}
    & \makecell{0.043$_{\pm 0.013}$ \\ 0.038$_{\pm 0.007}$ \\ 0.024$_{\pm 0.008}$}
    & \makecell{\textbf{1.000}$_{\pm 0.000}$ \\ \textbf{1.000}$_{\pm 0.000}$ \\ \textbf{1.000}$_{\pm 0.000}$}
    & \makecell{\textbf{1.000}$_{\pm 0.000}$ \\ \textbf{1.000}$_{\pm 0.000}$ \\ \textbf{1.000}$_{\pm 0.000}$}
    & \makecell{\textbf{1.000}$_{\pm 0.000}$ \\ \textbf{1.000}$_{\pm 0.000}$ \\ \textbf{1.000}$_{\pm 0.000}$}
    & \makecell{\textbf{1.000}$_{\pm 0.000}$ \\ \textbf{1.000}$_{\pm 0.000}$ \\ 0.999$_{\pm 0.001}$}
    & \makecell{\textbf{1.000}$_{\pm 0.000}$ \\ \textbf{1.000}$_{\pm 0.000}$ \\ \textbf{1.000}$_{\pm 0.000}$}
    & \makecell{\textbf{1.000}$_{\pm 0.000}$ \\ \textbf{1.000}$_{\pm 0.000}$ \\ \textbf{1.000}$_{\pm 0.000}$}
    & \makecell{0.756$_{\pm 0.002}$ \\ 0.756$_{\pm 0.001}$ \\ 0.754$_{\pm 0.001}$} \\
    PRC
    & \makecell{0.007$_{\pm 0.003}$ \\ 0.002$_{\pm 0.002}$ \\ 0.003$_{\pm 0.002}$}
    & \makecell{0.007$_{\pm 0.004}$ \\ 0.007$_{\pm 0.004}$ \\ 0.003$_{\pm 0.002}$}
    & \makecell{\textbf{1.000}$_{\pm 0.000}$ \\ \textbf{1.000}$_{\pm 0.000}$ \\ 0.999$_{\pm 0.001}$}
    & \makecell{\textbf{1.000}$_{\pm 0.000}$ \\ \textbf{1.000}$_{\pm 0.000}$ \\ 0.998$_{\pm 0.001}$}
    & \makecell{\textbf{1.000}$_{\pm 0.000}$ \\ 0.999$_{\pm 0.001}$ \\ 0.999$_{\pm 0.001}$}
    & \makecell{0.169$_{\pm 0.046}$ \\ 0.148$_{\pm 0.028}$ \\ 0.162$_{\pm 0.054}$}
    & \makecell{0.960$_{\pm 0.007}$ \\ 0.966$_{\pm 0.006}$ \\ 0.961$_{\pm 0.007}$}
    & \makecell{0.986$_{\pm 0.004}$ \\ 0.984$_{\pm 0.004}$ \\ 0.973$_{\pm 0.005}$}
    & \makecell{0.641$_{\pm 0.007}$ \\ 0.638$_{\pm 0.004}$ \\ 0.637$_{\pm 0.008}$} \\
    \midrule
    \rowcolor{rowgray}
    \textbf{DiffMark}
    & \makecell{\textbf{1.000}$_{\pm 0.000}$ \\ \textbf{1.000}$_{\pm 0.000}$ \\ 0.999$_{\pm 0.001}$}
    & \makecell{\textbf{0.999}$_{\pm 0.001}$ \\ 0.999$_{\pm 0.001}$ \\ \textbf{0.999}$_{\pm 0.001}$}
    & \makecell{\textbf{1.000}$_{\pm 0.000}$ \\ \textbf{1.000}$_{\pm 0.000}$ \\ \textbf{1.000}$_{\pm 0.000}$}
    & \makecell{\textbf{1.000}$_{\pm 0.000}$ \\ \textbf{1.000}$_{\pm 0.000}$ \\ \textbf{1.000}$_{\pm 0.000}$}
    & \makecell{\textbf{1.000}$_{\pm 0.000}$ \\ \textbf{1.000}$_{\pm 0.000}$ \\ \textbf{1.000}$_{\pm 0.000}$}
    & \makecell{\textbf{1.000}$_{\pm 0.000}$ \\ \textbf{1.000}$_{\pm 0.000}$ \\ 0.997$_{\pm 0.002}$}
    & \makecell{\textbf{1.000}$_{\pm 0.000}$ \\ \textbf{1.000}$_{\pm 0.000}$ \\ \textbf{1.000}$_{\pm 0.000}$}
    & \makecell{\textbf{1.000}$_{\pm 0.000}$ \\ \textbf{1.000}$_{\pm 0.000}$ \\ \textbf{1.000}$_{\pm 0.000}$}
    & \makecell{\textbf{1.000}$_{\pm 0.000}$ \\ \textbf{1.000}$_{\pm 0.000}$ \\ \textbf{0.999}$_{\pm 0.000}$} \\
    \bottomrule
    \end{tabular}%
    }
\end{table*}

\begin{table*}[t]
    \centering
    \caption{\textbf{Distortion breakdown with SD3 on DiffusionDB, MS-COCO, and DALL-E3.} Values are mean TPR@1\%FPR, with standard deviations in subscript. Each cell lists DiffusionDB, MS-COCO, and DALL-E3 results from top to bottom. The best result for each attack and dataset, including ties, is in \textbf{bold}. Avg is the reported average over the eight distortions.}
    \label{tab:supp_distortion_sd3}
    \resizebox{\textwidth}{!}{%
    \footnotesize
    \setlength{\tabcolsep}{4pt}
    \renewcommand{\arraystretch}{1.05}
    \begin{tabular}{l ccccccccc}
    \toprule
    \textbf{Method} & \textbf{Rotation} & \textbf{Crop} & \textbf{Erasing} & \textbf{Brightness} & \textbf{Contrast} & \textbf{Blur} & \textbf{Noise} & \textbf{Compress} & \textbf{Avg} \\
    \cmidrule(lr){1-1} \cmidrule(lr){2-2} \cmidrule(lr){3-3} \cmidrule(lr){4-4} \cmidrule(lr){5-5} \cmidrule(lr){6-6} \cmidrule(lr){7-7} \cmidrule(lr){8-8} \cmidrule(lr){9-9} \cmidrule(lr){10-10}
    \rowcolor{rowgray}
    Tree-Ring
    & \makecell{0.036$_{\pm 0.012}$ \\ 0.041$_{\pm 0.012}$ \\ 0.050$_{\pm 0.017}$}
    & \makecell{0.018$_{\pm 0.006}$ \\ 0.009$_{\pm 0.004}$ \\ 0.013$_{\pm 0.004}$}
    & \makecell{\textbf{1.000}$_{\pm 0.001}$ \\ 0.989$_{\pm 0.003}$ \\ 0.976$_{\pm 0.005}$}
    & \makecell{\textbf{1.000}$_{\pm 0.000}$ \\ 0.980$_{\pm 0.005}$ \\ 0.973$_{\pm 0.006}$}
    & \makecell{0.999$_{\pm 0.001}$ \\ 0.991$_{\pm 0.003}$ \\ 0.973$_{\pm 0.006}$}
    & \makecell{0.528$_{\pm 0.041}$ \\ 0.450$_{\pm 0.030}$ \\ 0.332$_{\pm 0.037}$}
    & \makecell{0.993$_{\pm 0.003}$ \\ 0.953$_{\pm 0.008}$ \\ 0.928$_{\pm 0.010}$}
    & \makecell{0.986$_{\pm 0.004}$ \\ 0.954$_{\pm 0.008}$ \\ 0.929$_{\pm 0.012}$}
    & \makecell{0.695$_{\pm 0.007}$ \\ 0.671$_{\pm 0.007}$ \\ 0.647$_{\pm 0.010}$} \\
    ShallowDiffuse
    & \makecell{0.000$_{\pm 0.000}$ \\ 0.000$_{\pm 0.000}$ \\ 0.000$_{\pm 0.000}$}
    & \makecell{0.012$_{\pm 0.004}$ \\ 0.012$_{\pm 0.007}$ \\ 0.011$_{\pm 0.006}$}
    & \makecell{\textbf{1.000}$_{\pm 0.000}$ \\ \textbf{1.000}$_{\pm 0.000}$ \\ \textbf{1.000}$_{\pm 0.000}$}
    & \makecell{\textbf{1.000}$_{\pm 0.000}$ \\ \textbf{1.000}$_{\pm 0.000}$ \\ \textbf{1.000}$_{\pm 0.000}$}
    & \makecell{\textbf{1.000}$_{\pm 0.000}$ \\ 0.998$_{\pm 0.001}$ \\ 0.999$_{\pm 0.001}$}
    & \makecell{0.005$_{\pm 0.003}$ \\ 0.002$_{\pm 0.002}$ \\ 0.004$_{\pm 0.003}$}
    & \makecell{\textbf{1.000}$_{\pm 0.000}$ \\ 0.997$_{\pm 0.002}$ \\ 0.999$_{\pm 0.001}$}
    & \makecell{\textbf{1.000}$_{\pm 0.000}$ \\ 0.999$_{\pm 0.001}$ \\ 0.999$_{\pm 0.001}$}
    & \makecell{0.627$_{\pm 0.001}$ \\ 0.626$_{\pm 0.001}$ \\ 0.627$_{\pm 0.001}$} \\
    \midrule
    \rowcolor{rowgray}
    StegaStamp
    & \makecell{0.010$_{\pm 0.005}$ \\ 0.022$_{\pm 0.011}$ \\ 0.007$_{\pm 0.005}$}
    & \makecell{0.411$_{\pm 0.024}$ \\ 0.373$_{\pm 0.021}$ \\ 0.379$_{\pm 0.026}$}
    & \makecell{\textbf{1.000}$_{\pm 0.000}$ \\ \textbf{1.000}$_{\pm 0.000}$ \\ \textbf{1.000}$_{\pm 0.000}$}
    & \makecell{\textbf{1.000}$_{\pm 0.000}$ \\ 0.996$_{\pm 0.002}$ \\ 0.992$_{\pm 0.003}$}
    & \makecell{0.998$_{\pm 0.001}$ \\ 0.996$_{\pm 0.002}$ \\ 0.981$_{\pm 0.005}$}
    & \makecell{0.420$_{\pm 0.047}$ \\ 0.381$_{\pm 0.039}$ \\ 0.355$_{\pm 0.056}$}
    & \makecell{\textbf{1.000}$_{\pm 0.000}$ \\ \textbf{1.000}$_{\pm 0.000}$ \\ \textbf{1.000}$_{\pm 0.000}$}
    & \makecell{\textbf{1.000}$_{\pm 0.000}$ \\ \textbf{1.000}$_{\pm 0.000}$ \\ \textbf{1.000}$_{\pm 0.000}$}
    & \makecell{0.730$_{\pm 0.009}$ \\ 0.721$_{\pm 0.008}$ \\ 0.714$_{\pm 0.011}$} \\
    RingID
    & \makecell{0.997$_{\pm 0.002}$ \\ 0.987$_{\pm 0.006}$ \\ 0.951$_{\pm 0.008}$}
    & \makecell{0.031$_{\pm 0.009}$ \\ 0.024$_{\pm 0.008}$ \\ 0.021$_{\pm 0.006}$}
    & \makecell{\textbf{1.000}$_{\pm 0.000}$ \\ \textbf{1.000}$_{\pm 0.000}$ \\ \textbf{1.000}$_{\pm 0.000}$}
    & \makecell{\textbf{1.000}$_{\pm 0.000}$ \\ 0.998$_{\pm 0.001}$ \\ 0.999$_{\pm 0.001}$}
    & \makecell{\textbf{1.000}$_{\pm 0.000}$ \\ \textbf{1.000}$_{\pm 0.000}$ \\ 0.997$_{\pm 0.002}$}
    & \makecell{0.970$_{\pm 0.006}$ \\ 0.912$_{\pm 0.015}$ \\ 0.829$_{\pm 0.017}$}
    & \makecell{0.999$_{\pm 0.001}$ \\ 0.995$_{\pm 0.002}$ \\ 0.991$_{\pm 0.004}$}
    & \makecell{\textbf{1.000}$_{\pm 0.000}$ \\ 0.994$_{\pm 0.003}$ \\ 0.992$_{\pm 0.003}$}
    & \makecell{0.875$_{\pm 0.002}$ \\ 0.864$_{\pm 0.004}$ \\ 0.847$_{\pm 0.004}$} \\
    \rowcolor{rowgray}
    GaussianShading
    & \makecell{0.008$_{\pm 0.004}$ \\ 0.020$_{\pm 0.008}$ \\ 0.012$_{\pm 0.005}$}
    & \makecell{0.037$_{\pm 0.014}$ \\ 0.027$_{\pm 0.009}$ \\ 0.022$_{\pm 0.008}$}
    & \makecell{\textbf{1.000}$_{\pm 0.000}$ \\ \textbf{1.000}$_{\pm 0.000}$ \\ \textbf{1.000}$_{\pm 0.000}$}
    & \makecell{\textbf{1.000}$_{\pm 0.000}$ \\ \textbf{1.000}$_{\pm 0.000}$ \\ \textbf{1.000}$_{\pm 0.000}$}
    & \makecell{\textbf{1.000}$_{\pm 0.000}$ \\ \textbf{1.000}$_{\pm 0.000}$ \\ \textbf{1.000}$_{\pm 0.000}$}
    & \makecell{0.986$_{\pm 0.005}$ \\ 0.954$_{\pm 0.010}$ \\ 0.903$_{\pm 0.014}$}
    & \makecell{\textbf{1.000}$_{\pm 0.000}$ \\ \textbf{1.000}$_{\pm 0.000}$ \\ 0.999$_{\pm 0.001}$}
    & \makecell{\textbf{1.000}$_{\pm 0.000}$ \\ \textbf{1.000}$_{\pm 0.000}$ \\ \textbf{1.000}$_{\pm 0.000}$}
    & \makecell{0.754$_{\pm 0.003}$ \\ 0.750$_{\pm 0.003}$ \\ 0.742$_{\pm 0.003}$} \\
    PRC
    & \makecell{0.000$_{\pm 0.000}$ \\ 0.000$_{\pm 0.000}$ \\ 0.000$_{\pm 0.000}$}
    & \makecell{0.001$_{\pm 0.001}$ \\ 0.000$_{\pm 0.001}$ \\ 0.000$_{\pm 0.000}$}
    & \makecell{\textbf{1.000}$_{\pm 0.000}$ \\ 0.996$_{\pm 0.003}$ \\ 0.980$_{\pm 0.005}$}
    & \makecell{0.998$_{\pm 0.002}$ \\ 0.911$_{\pm 0.010}$ \\ 0.898$_{\pm 0.010}$}
    & \makecell{0.994$_{\pm 0.002}$ \\ 0.940$_{\pm 0.008}$ \\ 0.862$_{\pm 0.011}$}
    & \makecell{0.985$_{\pm 0.012}$ \\ 0.828$_{\pm 0.068}$ \\ 0.807$_{\pm 0.045}$}
    & \makecell{0.881$_{\pm 0.011}$ \\ 0.696$_{\pm 0.015}$ \\ 0.675$_{\pm 0.015}$}
    & \makecell{0.802$_{\pm 0.013}$ \\ 0.624$_{\pm 0.016}$ \\ 0.556$_{\pm 0.016}$}
    & \makecell{0.708$_{\pm 0.004}$ \\ 0.624$_{\pm 0.011}$ \\ 0.597$_{\pm 0.009}$} \\
    \midrule
    \rowcolor{rowgray}
    \textbf{DiffMark}
    & \makecell{\textbf{1.000}$_{\pm 0.000}$ \\ \textbf{1.000}$_{\pm 0.000}$ \\ \textbf{0.999}$_{\pm 0.001}$}
    & \makecell{\textbf{1.000}$_{\pm 0.000}$ \\ \textbf{1.000}$_{\pm 0.000}$ \\ \textbf{0.992}$_{\pm 0.003}$}
    & \makecell{\textbf{1.000}$_{\pm 0.000}$ \\ \textbf{1.000}$_{\pm 0.000}$ \\ \textbf{1.000}$_{\pm 0.000}$}
    & \makecell{\textbf{1.000}$_{\pm 0.000}$ \\ \textbf{1.000}$_{\pm 0.000}$ \\ \textbf{1.000}$_{\pm 0.000}$}
    & \makecell{\textbf{1.000}$_{\pm 0.000}$ \\ 0.999$_{\pm 0.001}$ \\ \textbf{1.000}$_{\pm 0.000}$}
    & \makecell{\textbf{1.000}$_{\pm 0.000}$ \\ \textbf{1.000}$_{\pm 0.000}$ \\ \textbf{0.999}$_{\pm 0.001}$}
    & \makecell{\textbf{1.000}$_{\pm 0.000}$ \\ \textbf{1.000}$_{\pm 0.000}$ \\ \textbf{1.000}$_{\pm 0.000}$}
    & \makecell{\textbf{1.000}$_{\pm 0.000}$ \\ \textbf{1.000}$_{\pm 0.000}$ \\ \textbf{1.000}$_{\pm 0.000}$}
    & \makecell{\textbf{1.000}$_{\pm 0.000}$ \\ \textbf{1.000}$_{\pm 0.000}$ \\ \textbf{0.999}$_{\pm 0.000}$} \\
    \bottomrule
    \end{tabular}%
    }
    \vspace{-0.3cm}
\end{table*}

\begin{table*}[ht]
    \centering
    \caption{\textbf{Distortion breakdown with FLUX on DiffusionDB, MS-COCO, and DALL-E3.} Values are mean TPR@1\%FPR, with standard deviations in subscript. Each cell lists DiffusionDB, MS-COCO, and DALL-E3 results from top to bottom. The best result for each attack and dataset, including ties, is in \textbf{bold}. Avg is the reported average over the eight distortions.}
    \label{tab:supp_distortion_flux}
    \resizebox{\textwidth}{!}{%
    \footnotesize
    \setlength{\tabcolsep}{4pt}
    \renewcommand{\arraystretch}{1.05}
    \begin{tabular}{l ccccccccc}
    \toprule
    \textbf{Method} & \textbf{Rotation} & \textbf{Crop} & \textbf{Erasing} & \textbf{Brightness} & \textbf{Contrast} & \textbf{Blur} & \textbf{Noise} & \textbf{Compress} & \textbf{Avg} \\
    \cmidrule(lr){1-1} \cmidrule(lr){2-2} \cmidrule(lr){3-3} \cmidrule(lr){4-4} \cmidrule(lr){5-5} \cmidrule(lr){6-6} \cmidrule(lr){7-7} \cmidrule(lr){8-8} \cmidrule(lr){9-9} \cmidrule(lr){10-10}
    \rowcolor{rowgray}
    Tree-Ring
    & \makecell{0.004$_{\pm 0.003}$ \\ 0.009$_{\pm 0.005}$ \\ 0.014$_{\pm 0.006}$}
    & \makecell{0.001$_{\pm 0.002}$ \\ 0.000$_{\pm 0.001}$ \\ 0.002$_{\pm 0.002}$}
    & \makecell{\textbf{1.000}$_{\pm 0.000}$ \\ \textbf{1.000}$_{\pm 0.000}$ \\ 0.999$_{\pm 0.001}$}
    & \makecell{\textbf{1.000}$_{\pm 0.000}$ \\ 0.996$_{\pm 0.002}$ \\ 0.991$_{\pm 0.003}$}
    & \makecell{\textbf{1.000}$_{\pm 0.000}$ \\ 0.999$_{\pm 0.001}$ \\ 0.996$_{\pm 0.002}$}
    & \makecell{0.023$_{\pm 0.010}$ \\ 0.036$_{\pm 0.010}$ \\ 0.040$_{\pm 0.014}$}
    & \makecell{0.810$_{\pm 0.031}$ \\ 0.558$_{\pm 0.041}$ \\ 0.744$_{\pm 0.032}$}
    & \makecell{0.750$_{\pm 0.039}$ \\ 0.538$_{\pm 0.040}$ \\ 0.675$_{\pm 0.034}$}
    & \makecell{0.574$_{\pm 0.010}$ \\ 0.517$_{\pm 0.011}$ \\ 0.558$_{\pm 0.010}$} \\
    ShallowDiffuse
    & \makecell{0.002$_{\pm 0.002}$ \\ 0.007$_{\pm 0.003}$ \\ 0.002$_{\pm 0.003}$}
    & \makecell{0.033$_{\pm 0.008}$ \\ 0.022$_{\pm 0.007}$ \\ 0.029$_{\pm 0.015}$}
    & \makecell{\textbf{1.000}$_{\pm 0.000}$ \\ \textbf{1.000}$_{\pm 0.000}$ \\ \textbf{1.000}$_{\pm 0.000}$}
    & \makecell{\textbf{1.000}$_{\pm 0.000}$ \\ \textbf{1.000}$_{\pm 0.000}$ \\ \textbf{1.000}$_{\pm 0.000}$}
    & \makecell{\textbf{1.000}$_{\pm 0.000}$ \\ \textbf{1.000}$_{\pm 0.000}$ \\ \textbf{1.000}$_{\pm 0.000}$}
    & \makecell{0.007$_{\pm 0.004}$ \\ 0.006$_{\pm 0.003}$ \\ 0.005$_{\pm 0.004}$}
    & \makecell{\textbf{1.000}$_{\pm 0.000}$ \\ \textbf{1.000}$_{\pm 0.000}$ \\ \textbf{1.000}$_{\pm 0.000}$}
    & \makecell{\textbf{1.000}$_{\pm 0.000}$ \\ \textbf{1.000}$_{\pm 0.000}$ \\ \textbf{1.000}$_{\pm 0.000}$}
    & \makecell{0.630$_{\pm 0.001}$ \\ 0.629$_{\pm 0.001}$ \\ 0.630$_{\pm 0.003}$} \\
    \midrule
    \rowcolor{rowgray}
    StegaStamp
    & \makecell{0.006$_{\pm 0.004}$ \\ 0.031$_{\pm 0.015}$ \\ 0.007$_{\pm 0.008}$}
    & \makecell{0.439$_{\pm 0.018}$ \\ 0.460$_{\pm 0.025}$ \\ 0.423$_{\pm 0.027}$}
    & \makecell{\textbf{1.000}$_{\pm 0.000}$ \\ \textbf{1.000}$_{\pm 0.000}$ \\ \textbf{1.000}$_{\pm 0.000}$}
    & \makecell{\textbf{1.000}$_{\pm 0.000}$ \\ \textbf{1.000}$_{\pm 0.000}$ \\ 0.999$_{\pm 0.001}$}
    & \makecell{\textbf{1.000}$_{\pm 0.000}$ \\ \textbf{1.000}$_{\pm 0.000}$ \\ \textbf{1.000}$_{\pm 0.000}$}
    & \makecell{0.671$_{\pm 0.032}$ \\ 0.447$_{\pm 0.038}$ \\ 0.535$_{\pm 0.046}$}
    & \makecell{\textbf{1.000}$_{\pm 0.000}$ \\ \textbf{1.000}$_{\pm 0.000}$ \\ \textbf{1.000}$_{\pm 0.000}$}
    & \makecell{\textbf{1.000}$_{\pm 0.000}$ \\ \textbf{1.000}$_{\pm 0.000}$ \\ \textbf{1.000}$_{\pm 0.000}$}
    & \makecell{0.764$_{\pm 0.006}$ \\ 0.742$_{\pm 0.009}$ \\ 0.745$_{\pm 0.009}$} \\
    RingID
    & \makecell{0.101$_{\pm 0.022}$ \\ 0.069$_{\pm 0.023}$ \\ 0.097$_{\pm 0.025}$}
    & \makecell{0.002$_{\pm 0.002}$ \\ 0.003$_{\pm 0.002}$ \\ 0.005$_{\pm 0.003}$}
    & \makecell{\textbf{1.000}$_{\pm 0.000}$ \\ \textbf{1.000}$_{\pm 0.000}$ \\ \textbf{1.000}$_{\pm 0.000}$}
    & \makecell{\textbf{1.000}$_{\pm 0.000}$ \\ \textbf{1.000}$_{\pm 0.000}$ \\ 0.998$_{\pm 0.001}$}
    & \makecell{\textbf{1.000}$_{\pm 0.000}$ \\ \textbf{1.000}$_{\pm 0.000}$ \\ \textbf{1.000}$_{\pm 0.000}$}
    & \makecell{0.068$_{\pm 0.023}$ \\ 0.081$_{\pm 0.027}$ \\ 0.075$_{\pm 0.019}$}
    & \makecell{0.995$_{\pm 0.002}$ \\ 0.960$_{\pm 0.010}$ \\ 0.981$_{\pm 0.006}$}
    & \makecell{0.980$_{\pm 0.005}$ \\ 0.954$_{\pm 0.011}$ \\ 0.979$_{\pm 0.005}$}
    & \makecell{0.643$_{\pm 0.006}$ \\ 0.633$_{\pm 0.008}$ \\ 0.642$_{\pm 0.006}$} \\
    \rowcolor{rowgray}
    GaussianShading
    & \makecell{0.010$_{\pm 0.004}$ \\ 0.015$_{\pm 0.007}$ \\ 0.009$_{\pm 0.005}$}
    & \makecell{0.023$_{\pm 0.007}$ \\ 0.029$_{\pm 0.011}$ \\ 0.019$_{\pm 0.006}$}
    & \makecell{\textbf{1.000}$_{\pm 0.000}$ \\ \textbf{1.000}$_{\pm 0.000}$ \\ \textbf{1.000}$_{\pm 0.000}$}
    & \makecell{\textbf{1.000}$_{\pm 0.000}$ \\ \textbf{1.000}$_{\pm 0.000}$ \\ 0.999$_{\pm 0.001}$}
    & \makecell{\textbf{1.000}$_{\pm 0.000}$ \\ \textbf{1.000}$_{\pm 0.000}$ \\ \textbf{1.000}$_{\pm 0.000}$}
    & \makecell{0.948$_{\pm 0.010}$ \\ 0.907$_{\pm 0.021}$ \\ 0.867$_{\pm 0.026}$}
    & \makecell{\textbf{1.000}$_{\pm 0.000}$ \\ \textbf{1.000}$_{\pm 0.000}$ \\ \textbf{1.000}$_{\pm 0.000}$}
    & \makecell{\textbf{1.000}$_{\pm 0.000}$ \\ \textbf{1.000}$_{\pm 0.000}$ \\ \textbf{1.000}$_{\pm 0.000}$}
    & \makecell{0.748$_{\pm 0.002}$ \\ 0.744$_{\pm 0.004}$ \\ 0.737$_{\pm 0.004}$} \\
    PRC
    & \makecell{0.000$_{\pm 0.000}$ \\ 0.000$_{\pm 0.000}$ \\ 0.000$_{\pm 0.000}$}
    & \makecell{0.000$_{\pm 0.001}$ \\ 0.003$_{\pm 0.002}$ \\ 0.000$_{\pm 0.000}$}
    & \makecell{\textbf{1.000}$_{\pm 0.000}$ \\ 0.999$_{\pm 0.001}$ \\ 0.996$_{\pm 0.002}$}
    & \makecell{\textbf{1.000}$_{\pm 0.000}$ \\ 0.944$_{\pm 0.008}$ \\ 0.961$_{\pm 0.006}$}
    & \makecell{\textbf{1.000}$_{\pm 0.000}$ \\ 0.994$_{\pm 0.003}$ \\ 0.986$_{\pm 0.004}$}
    & \makecell{0.006$_{\pm 0.003}$ \\ 0.005$_{\pm 0.002}$ \\ 0.005$_{\pm 0.003}$}
    & \makecell{0.735$_{\pm 0.015}$ \\ 0.430$_{\pm 0.016}$ \\ 0.630$_{\pm 0.017}$}
    & \makecell{0.763$_{\pm 0.014}$ \\ 0.461$_{\pm 0.018}$ \\ 0.603$_{\pm 0.018}$}
    & \makecell{0.563$_{\pm 0.004}$ \\ 0.479$_{\pm 0.004}$ \\ 0.523$_{\pm 0.005}$} \\
    \midrule
    \rowcolor{rowgray}
    \textbf{DiffMark}
    & \makecell{\textbf{1.000}$_{\pm 0.000}$ \\ \textbf{0.997}$_{\pm 0.002}$ \\ \textbf{0.984}$_{\pm 0.004}$}
    & \makecell{\textbf{1.000}$_{\pm 0.000}$ \\ \textbf{0.993}$_{\pm 0.003}$ \\ \textbf{0.971}$_{\pm 0.005}$}
    & \makecell{\textbf{1.000}$_{\pm 0.000}$ \\ \textbf{1.000}$_{\pm 0.000}$ \\ \textbf{1.000}$_{\pm 0.000}$}
    & \makecell{\textbf{1.000}$_{\pm 0.000}$ \\ 0.999$_{\pm 0.001}$ \\ 0.999$_{\pm 0.001}$}
    & \makecell{\textbf{1.000}$_{\pm 0.000}$ \\ 0.999$_{\pm 0.001}$ \\ \textbf{1.000}$_{\pm 0.000}$}
    & \makecell{\textbf{1.000}$_{\pm 0.000}$ \\ \textbf{1.000}$_{\pm 0.000}$ \\ \textbf{1.000}$_{\pm 0.000}$}
    & \makecell{\textbf{1.000}$_{\pm 0.000}$ \\ \textbf{1.000}$_{\pm 0.000}$ \\ \textbf{1.000}$_{\pm 0.000}$}
    & \makecell{\textbf{1.000}$_{\pm 0.000}$ \\ \textbf{1.000}$_{\pm 0.000}$ \\ \textbf{1.000}$_{\pm 0.000}$}
    & \makecell{\textbf{1.000}$_{\pm 0.000}$ \\ \textbf{0.998}$_{\pm 0.001}$ \\ \textbf{0.994}$_{\pm 0.001}$} \\
    \bottomrule
    \end{tabular}%
    }
    \vspace{-0.5cm}
\end{table*}

\subsection{Distortion attack breakdown}
\label{sec:supp_distortion_results}
\cref{tab:supp_distortion_sd21,tab:supp_distortion_sd3,tab:supp_distortion_flux} report TPR@$1\%$FPR for SD~2.1, SD3, and FLUX on DiffusionDB, MS-COCO, and DALL-E3. Each generator has one table, with DiffusionDB, MS-COCO, and DALL-E3 results listed from top to bottom in every cell. The eight distortions and their strengths are specified in \cref{tab:supp_distortions}. Standard deviations are shown in subscript. Avg. is the reported average over the eight distortions, excluding clean images. Only methods with results for the corresponding generator are included.

\textbf{Overall robustness.} \ours{} maintains an average TPR of at least $0.994$ across all nine generator--dataset settings. Several baselines retain high TPR under brightness and contrast changes, but their aggregate results are limited by rotation, cropping, or blur. The following breakdown identifies these weaknesses.

\textbf{Geometric distortions.} Rotation and resized cropping are the clearest sources of failure for many baselines. RingID preserves rotation detection on SD~2.1 but loses it on FLUX, and its crop TPR remains low across all three generators. Stable Signature and AquaLoRA-robust tolerate cropping on SD~2.1 but fail under blur and rotation, respectively. \ours{} handles both transformations across generators. Its lowest geometric TPR occurs on FLUX with DALL-E3, at $0.984$ for rotation and $0.971$ for cropping. Random erasing is less disruptive for most methods, and \ours{} reaches $1.000$ in every setting.

\textbf{Brightness, contrast, and noise.} Most methods tolerate brightness and contrast changes, so these attacks show smaller performance differences than rotation and cropping. Gaussian noise reveals larger differences on SD3 and FLUX, particularly for PRC and Tree-Ring. \ours{} stays at or above $0.999$ under brightness and contrast changes and reaches $1.000$ under noise in all nine settings.

\textbf{Blur and compression.} Blur distinguishes methods even when their brightness, contrast, and noise results are strong. Shallow Diffuse has blur TPR at or below $0.015$ across all settings, and Stable Signature falls to $0.000$ on SD~2.1. Tree-Ring and RingID also lose substantial blur robustness on FLUX. \ours{} maintains blur TPR of at least $0.997$ and compression TPR of $1.000$ throughout. These results support robustness to the tested distortions across generators and datasets.

\subsection{Removal and adversarial attack breakdown}
\label{sec:supp_removal_adv_results}
\cref{tab:supp_removal_adv_sd21,tab:supp_removal_adv_sd3,tab:supp_removal_adv_flux} give the removal and adversarial results for each generator. The removal group contains Regen-VAE, Regen-Diff, and the three Rinse-2x settings. The adversarial group contains Adv-KLVAE8 and Adv-RN18. Avg gives the reported mean for each attack group, and each cell follows the same dataset order as the distortion tables. Values use TPR@1\%FPR with standard deviations in subscript.

\begin{table*}[ht]
    \centering
    \caption{\textbf{Removal and adversarial attack breakdown with SD~2.1 on DiffusionDB, MS-COCO, and DALL-E3.} Values are mean TPR@1\%FPR, with standard deviations in subscript. Each cell lists DiffusionDB, MS-COCO, and DALL-E3 results from top to bottom. The best result for each attack and dataset, including ties, is in \textbf{bold}. Avg is the reported average within each attack group.}
    \label{tab:supp_removal_adv_sd21}
    \resizebox{\textwidth}{!}{%
    \footnotesize
    \setlength{\tabcolsep}{4pt}
    \renewcommand{\arraystretch}{1.05}
    \begin{tabular}{l cccccc @{\hspace{10pt}} ccc}
    \toprule
    \multirow{2}{*}{\textbf{Method}}
    & \multicolumn{6}{c}{\textbf{Removal Attacks}}
    & \multicolumn{3}{c}{\textbf{Adversarial Attacks}} \\
    \cmidrule(lr){2-7} \cmidrule(lr){8-10}
    & \textbf{Regen-VAE} & \textbf{Regen-Diff} & \textbf{Rinse-2x-20} & \textbf{Rinse-2x-60} & \textbf{Rinse-2x-100} & \textbf{Avg} & \textbf{Adv-KLVAE8} & \textbf{Adv-RN18} & \textbf{Avg} \\
    \cmidrule(lr){1-1} \cmidrule(lr){2-2} \cmidrule(lr){3-3} \cmidrule(lr){4-4} \cmidrule(lr){5-5} \cmidrule(lr){6-6} \cmidrule(lr){7-7} \cmidrule(lr){8-8} \cmidrule(lr){9-9} \cmidrule(lr){10-10}
    \rowcolor{rowgray}
    Tree-Ring
    & \makecell{0.997$_{\pm 0.001}$ \\ 0.981$_{\pm 0.002}$ \\ 0.964$_{\pm 0.004}$}
    & \makecell{0.999$_{\pm 0.001}$ \\ 0.998$_{\pm 0.002}$ \\ 0.994$_{\pm 0.002}$}
    & \makecell{0.998$_{\pm 0.002}$ \\ 0.998$_{\pm 0.000}$ \\ 0.985$_{\pm 0.001}$}
    & \makecell{0.983$_{\pm 0.007}$ \\ 0.974$_{\pm 0.004}$ \\ 0.953$_{\pm 0.008}$}
    & \makecell{0.945$_{\pm 0.006}$ \\ 0.907$_{\pm 0.010}$ \\ 0.899$_{\pm 0.006}$}
    & \makecell{0.985$_{\pm 0.003}$ \\ 0.971$_{\pm 0.003}$ \\ 0.959$_{\pm 0.003}$}
    & \makecell{0.684$_{\pm 0.036}$ \\ 0.719$_{\pm 0.010}$ \\ 0.689$_{\pm 0.006}$}
    & \makecell{0.998$_{\pm 0.000}$ \\ 0.999$_{\pm 0.001}$ \\ 0.990$_{\pm 0.000}$}
    & \makecell{0.841$_{\pm 0.018}$ \\ 0.859$_{\pm 0.005}$ \\ 0.840$_{\pm 0.003}$} \\
    ShallowDiffuse
    & \makecell{0.899$_{\pm 0.022}$ \\ 0.977$_{\pm 0.007}$ \\ 0.933$_{\pm 0.013}$}
    & \makecell{\textbf{1.000}$_{\pm 0.000}$ \\ \textbf{1.000}$_{\pm 0.000}$ \\ \textbf{1.000}$_{\pm 0.000}$}
    & \makecell{\textbf{1.000}$_{\pm 0.000}$ \\ \textbf{1.000}$_{\pm 0.000}$ \\ 0.996$_{\pm 0.002}$}
    & \makecell{0.999$_{\pm 0.001}$ \\ 0.999$_{\pm 0.002}$ \\ 0.993$_{\pm 0.001}$}
    & \makecell{0.994$_{\pm 0.000}$ \\ 0.981$_{\pm 0.001}$ \\ 0.981$_{\pm 0.006}$}
    & \makecell{0.978$_{\pm 0.005}$ \\ 0.991$_{\pm 0.002}$ \\ 0.981$_{\pm 0.002}$}
    & \makecell{0.091$_{\pm 0.022}$ \\ 0.407$_{\pm 0.016}$ \\ 0.417$_{\pm 0.035}$}
    & \makecell{\textbf{1.000}$_{\pm 0.000}$ \\ \textbf{1.000}$_{\pm 0.000}$ \\ 0.999$_{\pm 0.001}$}
    & \makecell{0.545$_{\pm 0.011}$ \\ 0.703$_{\pm 0.008}$ \\ 0.708$_{\pm 0.017}$} \\
    \rowcolor{rowgray}
    FARI-T
    & \makecell{0.999$_{\pm 0.001}$ \\ 0.989$_{\pm 0.001}$ \\ 0.973$_{\pm 0.005}$}
    & \makecell{0.999$_{\pm 0.001}$ \\ 0.997$_{\pm 0.001}$ \\ 0.987$_{\pm 0.003}$}
    & \makecell{0.990$_{\pm 0.005}$ \\ 0.991$_{\pm 0.004}$ \\ 0.973$_{\pm 0.006}$}
    & \makecell{0.969$_{\pm 0.008}$ \\ 0.954$_{\pm 0.012}$ \\ 0.933$_{\pm 0.004}$}
    & \makecell{0.912$_{\pm 0.003}$ \\ 0.867$_{\pm 0.020}$ \\ 0.831$_{\pm 0.006}$}
    & \makecell{0.974$_{\pm 0.003}$ \\ 0.959$_{\pm 0.008}$ \\ 0.939$_{\pm 0.003}$}
    & \makecell{0.817$_{\pm 0.025}$ \\ 0.878$_{\pm 0.017}$ \\ 0.811$_{\pm 0.008}$}
    & \makecell{0.997$_{\pm 0.001}$ \\ 0.999$_{\pm 0.001}$ \\ 0.991$_{\pm 0.001}$}
    & \makecell{0.907$_{\pm 0.013}$ \\ 0.939$_{\pm 0.009}$ \\ 0.901$_{\pm 0.005}$} \\
    \midrule
    StegaStamp
    & \makecell{\textbf{1.000}$_{\pm 0.000}$ \\ \textbf{1.000}$_{\pm 0.000}$ \\ \textbf{1.000}$_{\pm 0.000}$}
    & \makecell{\textbf{1.000}$_{\pm 0.000}$ \\ \textbf{1.000}$_{\pm 0.000}$ \\ 0.999$_{\pm 0.001}$}
    & \makecell{0.999$_{\pm 0.001}$ \\ 0.997$_{\pm 0.001}$ \\ 0.994$_{\pm 0.002}$}
    & \makecell{0.931$_{\pm 0.019}$ \\ 0.930$_{\pm 0.007}$ \\ 0.883$_{\pm 0.008}$}
    & \makecell{0.631$_{\pm 0.072}$ \\ 0.681$_{\pm 0.012}$ \\ 0.651$_{\pm 0.004}$}
    & \makecell{0.912$_{\pm 0.018}$ \\ 0.922$_{\pm 0.004}$ \\ 0.905$_{\pm 0.002}$}
    & \makecell{\textbf{1.000}$_{\pm 0.000}$ \\ \textbf{1.000}$_{\pm 0.000}$ \\ \textbf{1.000}$_{\pm 0.000}$}
    & \makecell{\textbf{1.000}$_{\pm 0.000}$ \\ \textbf{1.000}$_{\pm 0.000}$ \\ \textbf{1.000}$_{\pm 0.000}$}
    & \makecell{\textbf{1.000}$_{\pm 0.000}$ \\ \textbf{1.000}$_{\pm 0.000}$ \\ \textbf{1.000}$_{\pm 0.000}$} \\
    \rowcolor{rowgray}
    RingID
    & \makecell{\textbf{1.000}$_{\pm 0.000}$ \\ \textbf{1.000}$_{\pm 0.000}$ \\ \textbf{1.000}$_{\pm 0.000}$}
    & \makecell{\textbf{1.000}$_{\pm 0.000}$ \\ \textbf{1.000}$_{\pm 0.000}$ \\ \textbf{1.000}$_{\pm 0.000}$}
    & \makecell{\textbf{1.000}$_{\pm 0.000}$ \\ \textbf{1.000}$_{\pm 0.000}$ \\ \textbf{1.000}$_{\pm 0.000}$}
    & \makecell{\textbf{1.000}$_{\pm 0.000}$ \\ \textbf{1.000}$_{\pm 0.000}$ \\ 0.997$_{\pm 0.002}$}
    & \makecell{0.999$_{\pm 0.001}$ \\ \textbf{1.000}$_{\pm 0.000}$ \\ 0.993$_{\pm 0.001}$}
    & \makecell{\textbf{1.000}$_{\pm 0.000}$ \\ \textbf{1.000}$_{\pm 0.000}$ \\ 0.998$_{\pm 0.001}$}
    & \makecell{0.745$_{\pm 0.006}$ \\ 0.857$_{\pm 0.011}$ \\ 0.821$_{\pm 0.008}$}
    & \makecell{\textbf{1.000}$_{\pm 0.000}$ \\ \textbf{1.000}$_{\pm 0.000}$ \\ \textbf{1.000}$_{\pm 0.000}$}
    & \makecell{0.873$_{\pm 0.003}$ \\ 0.929$_{\pm 0.006}$ \\ 0.911$_{\pm 0.004}$} \\
    StableSignature
    & \makecell{0.003$_{\pm 0.001}$ \\ 0.015$_{\pm 0.008}$ \\ 0.009$_{\pm 0.004}$}
    & \makecell{0.009$_{\pm 0.004}$ \\ 0.026$_{\pm 0.011}$ \\ 0.011$_{\pm 0.005}$}
    & \makecell{0.004$_{\pm 0.003}$ \\ 0.016$_{\pm 0.009}$ \\ 0.009$_{\pm 0.006}$}
    & \makecell{0.001$_{\pm 0.001}$ \\ 0.012$_{\pm 0.004}$ \\ 0.006$_{\pm 0.005}$}
    & \makecell{0.001$_{\pm 0.001}$ \\ 0.007$_{\pm 0.004}$ \\ 0.006$_{\pm 0.005}$}
    & \makecell{0.004$_{\pm 0.001}$ \\ 0.015$_{\pm 0.007}$ \\ 0.008$_{\pm 0.005}$}
    & \makecell{0.999$_{\pm 0.001}$ \\ \textbf{1.000}$_{\pm 0.000}$ \\ 0.997$_{\pm 0.003}$}
    & \makecell{\textbf{1.000}$_{\pm 0.000}$ \\ \textbf{1.000}$_{\pm 0.000}$ \\ 0.999$_{\pm 0.001}$}
    & \makecell{0.999$_{\pm 0.001}$ \\ \textbf{1.000}$_{\pm 0.000}$ \\ 0.998$_{\pm 0.002}$} \\
    \rowcolor{rowgray}
    AquaLoRA-robust
    & \makecell{0.961$_{\pm 0.001}$ \\ 0.910$_{\pm 0.027}$ \\ 0.880$_{\pm 0.023}$}
    & \makecell{0.977$_{\pm 0.001}$ \\ 0.953$_{\pm 0.009}$ \\ 0.943$_{\pm 0.008}$}
    & \makecell{0.967$_{\pm 0.009}$ \\ 0.917$_{\pm 0.013}$ \\ 0.890$_{\pm 0.020}$}
    & \makecell{0.891$_{\pm 0.008}$ \\ 0.854$_{\pm 0.041}$ \\ 0.804$_{\pm 0.028}$}
    & \makecell{0.731$_{\pm 0.006}$ \\ 0.658$_{\pm 0.057}$ \\ 0.607$_{\pm 0.048}$}
    & \makecell{0.906$_{\pm 0.003}$ \\ 0.859$_{\pm 0.029}$ \\ 0.825$_{\pm 0.024}$}
    & \makecell{0.975$_{\pm 0.004}$ \\ 0.946$_{\pm 0.003}$ \\ 0.888$_{\pm 0.009}$}
    & \makecell{0.981$_{\pm 0.004}$ \\ 0.947$_{\pm 0.013}$ \\ 0.929$_{\pm 0.010}$}
    & \makecell{0.978$_{\pm 0.004}$ \\ 0.947$_{\pm 0.008}$ \\ 0.908$_{\pm 0.009}$} \\
    GaussianShading
    & \makecell{\textbf{1.000}$_{\pm 0.000}$ \\ \textbf{1.000}$_{\pm 0.000}$ \\ 0.999$_{\pm 0.001}$}
    & \makecell{\textbf{1.000}$_{\pm 0.000}$ \\ \textbf{1.000}$_{\pm 0.000}$ \\ \textbf{1.000}$_{\pm 0.000}$}
    & \makecell{\textbf{1.000}$_{\pm 0.000}$ \\ \textbf{1.000}$_{\pm 0.000}$ \\ \textbf{1.000}$_{\pm 0.000}$}
    & \makecell{\textbf{1.000}$_{\pm 0.000}$ \\ \textbf{1.000}$_{\pm 0.000}$ \\ \textbf{1.000}$_{\pm 0.000}$}
    & \makecell{\textbf{1.000}$_{\pm 0.000}$ \\ \textbf{1.000}$_{\pm 0.000}$ \\ \textbf{1.000}$_{\pm 0.000}$}
    & \makecell{\textbf{1.000}$_{\pm 0.000}$ \\ \textbf{1.000}$_{\pm 0.000}$ \\ \textbf{1.000}$_{\pm 0.000}$}
    & \makecell{0.331$_{\pm 0.006}$ \\ 0.689$_{\pm 0.013}$ \\ 0.666$_{\pm 0.034}$}
    & \makecell{\textbf{1.000}$_{\pm 0.000}$ \\ \textbf{1.000}$_{\pm 0.000}$ \\ \textbf{1.000}$_{\pm 0.000}$}
    & \makecell{0.666$_{\pm 0.003}$ \\ 0.844$_{\pm 0.007}$ \\ 0.833$_{\pm 0.017}$} \\
    \rowcolor{rowgray}
    FARI-GS
    & \makecell{\textbf{1.000}$_{\pm 0.000}$ \\ \textbf{1.000}$_{\pm 0.000}$ \\ \textbf{1.000}$_{\pm 0.000}$}
    & \makecell{\textbf{1.000}$_{\pm 0.000}$ \\ \textbf{1.000}$_{\pm 0.000}$ \\ \textbf{1.000}$_{\pm 0.000}$}
    & \makecell{\textbf{1.000}$_{\pm 0.000}$ \\ \textbf{1.000}$_{\pm 0.000}$ \\ \textbf{1.000}$_{\pm 0.000}$}
    & \makecell{\textbf{1.000}$_{\pm 0.000}$ \\ \textbf{1.000}$_{\pm 0.000}$ \\ \textbf{1.000}$_{\pm 0.000}$}
    & \makecell{\textbf{1.000}$_{\pm 0.000}$ \\ \textbf{1.000}$_{\pm 0.000}$ \\ \textbf{1.000}$_{\pm 0.000}$}
    & \makecell{\textbf{1.000}$_{\pm 0.000}$ \\ \textbf{1.000}$_{\pm 0.000}$ \\ \textbf{1.000}$_{\pm 0.000}$}
    & \makecell{0.540$_{\pm 0.017}$ \\ 0.888$_{\pm 0.009}$ \\ 0.825$_{\pm 0.017}$}
    & \makecell{\textbf{1.000}$_{\pm 0.000}$ \\ \textbf{1.000}$_{\pm 0.000}$ \\ \textbf{1.000}$_{\pm 0.000}$}
    & \makecell{0.770$_{\pm 0.009}$ \\ 0.944$_{\pm 0.005}$ \\ 0.913$_{\pm 0.009}$} \\
    PRC
    & \makecell{0.472$_{\pm 0.024}$ \\ 0.590$_{\pm 0.016}$ \\ 0.657$_{\pm 0.011}$}
    & \makecell{0.999$_{\pm 0.002}$ \\ 0.983$_{\pm 0.004}$ \\ 0.973$_{\pm 0.005}$}
    & \makecell{0.995$_{\pm 0.003}$ \\ 0.954$_{\pm 0.007}$ \\ 0.949$_{\pm 0.005}$}
    & \makecell{0.960$_{\pm 0.010}$ \\ 0.839$_{\pm 0.023}$ \\ 0.883$_{\pm 0.004}$}
    & \makecell{0.791$_{\pm 0.012}$ \\ 0.655$_{\pm 0.029}$ \\ 0.722$_{\pm 0.020}$}
    & \makecell{0.843$_{\pm 0.005}$ \\ 0.804$_{\pm 0.014}$ \\ 0.837$_{\pm 0.006}$}
    & \makecell{0.099$_{\pm 0.009}$ \\ 0.301$_{\pm 0.021}$ \\ 0.374$_{\pm 0.024}$}
    & \makecell{0.967$_{\pm 0.006}$ \\ 0.962$_{\pm 0.007}$ \\ 0.950$_{\pm 0.006}$}
    & \makecell{0.533$_{\pm 0.004}$ \\ 0.631$_{\pm 0.014}$ \\ 0.662$_{\pm 0.010}$} \\
    \midrule
    \rowcolor{rowgray}
    \textbf{DiffMark}
    & \makecell{\textbf{1.000}$_{\pm 0.000}$ \\ \textbf{1.000}$_{\pm 0.000}$ \\ 0.999$_{\pm 0.001}$}
    & \makecell{\textbf{1.000}$_{\pm 0.000}$ \\ \textbf{1.000}$_{\pm 0.000}$ \\ \textbf{1.000}$_{\pm 0.000}$}
    & \makecell{\textbf{1.000}$_{\pm 0.000}$ \\ \textbf{1.000}$_{\pm 0.000}$ \\ \textbf{1.000}$_{\pm 0.000}$}
    & \makecell{\textbf{1.000}$_{\pm 0.000}$ \\ \textbf{1.000}$_{\pm 0.000}$ \\ \textbf{1.000}$_{\pm 0.000}$}
    & \makecell{\textbf{1.000}$_{\pm 0.000}$ \\ \textbf{1.000}$_{\pm 0.000}$ \\ \textbf{1.000}$_{\pm 0.000}$}
    & \makecell{\textbf{1.000}$_{\pm 0.000}$ \\ \textbf{1.000}$_{\pm 0.000}$ \\ \textbf{1.000}$_{\pm 0.000}$}
    & \makecell{\textbf{1.000}$_{\pm 0.000}$ \\ \textbf{1.000}$_{\pm 0.000}$ \\ \textbf{1.000}$_{\pm 0.000}$}
    & \makecell{\textbf{1.000}$_{\pm 0.000}$ \\ \textbf{1.000}$_{\pm 0.000}$ \\ \textbf{1.000}$_{\pm 0.000}$}
    & \makecell{\textbf{1.000}$_{\pm 0.000}$ \\ \textbf{1.000}$_{\pm 0.000}$ \\ \textbf{1.000}$_{\pm 0.000}$} \\
    \bottomrule
    \end{tabular}%
    }
\end{table*}

\begin{table*}[ht]
    \centering
    \caption{\textbf{Removal and adversarial attack breakdown with SD3 on DiffusionDB, MS-COCO, and DALL-E3.} Values are mean TPR@1\%FPR, with standard deviations in subscript. Each cell lists DiffusionDB, MS-COCO, and DALL-E3 results from top to bottom. The best result for each attack and dataset, including ties, is in \textbf{bold}. Avg is the reported average within each attack group.}
    \label{tab:supp_removal_adv_sd3}
    \resizebox{\textwidth}{!}{%
    \footnotesize
    \setlength{\tabcolsep}{4pt}
    \renewcommand{\arraystretch}{1.05}
    \begin{tabular}{l cccccc @{\hspace{10pt}} ccc}
    \toprule
    \multirow{2}{*}{\textbf{Method}}
    & \multicolumn{6}{c}{\textbf{Removal Attacks}}
    & \multicolumn{3}{c}{\textbf{Adversarial Attacks}} \\
    \cmidrule(lr){2-7} \cmidrule(lr){8-10}
    & \textbf{Regen-VAE} & \textbf{Regen-Diff} & \textbf{Rinse-2x-20} & \textbf{Rinse-2x-60} & \textbf{Rinse-2x-100} & \textbf{Avg} & \textbf{Adv-KLVAE8} & \textbf{Adv-RN18} & \textbf{Avg} \\
    \cmidrule(lr){1-1} \cmidrule(lr){2-2} \cmidrule(lr){3-3} \cmidrule(lr){4-4} \cmidrule(lr){5-5} \cmidrule(lr){6-6} \cmidrule(lr){7-7} \cmidrule(lr){8-8} \cmidrule(lr){9-9} \cmidrule(lr){10-10}
    \rowcolor{rowgray}
    Tree-Ring
    & \makecell{0.653$_{\pm 0.069}$ \\ 0.542$_{\pm 0.013}$ \\ 0.531$_{\pm 0.034}$}
    & \makecell{0.859$_{\pm 0.032}$ \\ 0.675$_{\pm 0.001}$ \\ 0.658$_{\pm 0.018}$}
    & \makecell{0.817$_{\pm 0.049}$ \\ 0.607$_{\pm 0.008}$ \\ 0.624$_{\pm 0.032}$}
    & \makecell{0.643$_{\pm 0.043}$ \\ 0.441$_{\pm 0.020}$ \\ 0.449$_{\pm 0.008}$}
    & \makecell{0.522$_{\pm 0.054}$ \\ 0.331$_{\pm 0.012}$ \\ 0.353$_{\pm 0.032}$}
    & \makecell{0.699$_{\pm 0.049}$ \\ 0.519$_{\pm 0.008}$ \\ 0.523$_{\pm 0.023}$}
    & \makecell{0.917$_{\pm 0.016}$ \\ 0.883$_{\pm 0.008}$ \\ 0.853$_{\pm 0.014}$}
    & \makecell{0.979$_{\pm 0.012}$ \\ 0.911$_{\pm 0.011}$ \\ 0.889$_{\pm 0.012}$}
    & \makecell{0.948$_{\pm 0.014}$ \\ 0.897$_{\pm 0.009}$ \\ 0.871$_{\pm 0.012}$} \\
    ShallowDiffuse
    & \makecell{0.981$_{\pm 0.006}$ \\ 0.937$_{\pm 0.012}$ \\ 0.936$_{\pm 0.006}$}
    & \makecell{\textbf{1.000}$_{\pm 0.000}$ \\ 0.979$_{\pm 0.008}$ \\ 0.969$_{\pm 0.004}$}
    & \makecell{0.999$_{\pm 0.001}$ \\ 0.969$_{\pm 0.015}$ \\ 0.949$_{\pm 0.003}$}
    & \makecell{0.996$_{\pm 0.002}$ \\ 0.939$_{\pm 0.021}$ \\ 0.908$_{\pm 0.005}$}
    & \makecell{0.983$_{\pm 0.005}$ \\ 0.865$_{\pm 0.024}$ \\ 0.853$_{\pm 0.012}$}
    & \makecell{0.992$_{\pm 0.002}$ \\ 0.938$_{\pm 0.016}$ \\ 0.923$_{\pm 0.005}$}
    & \makecell{\textbf{1.000}$_{\pm 0.000}$ \\ 0.997$_{\pm 0.002}$ \\ 0.997$_{\pm 0.003}$}
    & \makecell{\textbf{1.000}$_{\pm 0.000}$ \\ 0.999$_{\pm 0.001}$ \\ 0.996$_{\pm 0.003}$}
    & \makecell{\textbf{1.000}$_{\pm 0.000}$ \\ 0.998$_{\pm 0.002}$ \\ 0.997$_{\pm 0.003}$} \\
    \midrule
    \rowcolor{rowgray}
    StegaStamp
    & \makecell{\textbf{1.000}$_{\pm 0.000}$ \\ \textbf{1.000}$_{\pm 0.000}$ \\ \textbf{1.000}$_{\pm 0.000}$}
    & \makecell{\textbf{1.000}$_{\pm 0.000}$ \\ 0.999$_{\pm 0.001}$ \\ 0.998$_{\pm 0.002}$}
    & \makecell{0.999$_{\pm 0.001}$ \\ 0.995$_{\pm 0.001}$ \\ 0.989$_{\pm 0.004}$}
    & \makecell{0.979$_{\pm 0.009}$ \\ 0.957$_{\pm 0.011}$ \\ 0.935$_{\pm 0.010}$}
    & \makecell{0.810$_{\pm 0.007}$ \\ 0.783$_{\pm 0.045}$ \\ 0.752$_{\pm 0.026}$}
    & \makecell{0.957$_{\pm 0.003}$ \\ 0.947$_{\pm 0.011}$ \\ 0.935$_{\pm 0.008}$}
    & \makecell{\textbf{1.000}$_{\pm 0.000}$ \\ \textbf{1.000}$_{\pm 0.000}$ \\ \textbf{1.000}$_{\pm 0.000}$}
    & \makecell{\textbf{1.000}$_{\pm 0.000}$ \\ \textbf{1.000}$_{\pm 0.000}$ \\ \textbf{1.000}$_{\pm 0.000}$}
    & \makecell{\textbf{1.000}$_{\pm 0.000}$ \\ \textbf{1.000}$_{\pm 0.000}$ \\ \textbf{1.000}$_{\pm 0.000}$} \\
    RingID
    & \makecell{0.978$_{\pm 0.014}$ \\ 0.933$_{\pm 0.008}$ \\ 0.909$_{\pm 0.015}$}
    & \makecell{0.997$_{\pm 0.001}$ \\ 0.971$_{\pm 0.004}$ \\ 0.939$_{\pm 0.010}$}
    & \makecell{0.999$_{\pm 0.001}$ \\ 0.963$_{\pm 0.006}$ \\ 0.931$_{\pm 0.005}$}
    & \makecell{0.992$_{\pm 0.004}$ \\ 0.890$_{\pm 0.013}$ \\ 0.874$_{\pm 0.011}$}
    & \makecell{0.980$_{\pm 0.008}$ \\ 0.813$_{\pm 0.025}$ \\ 0.822$_{\pm 0.021}$}
    & \makecell{0.989$_{\pm 0.005}$ \\ 0.914$_{\pm 0.008}$ \\ 0.895$_{\pm 0.011}$}
    & \makecell{\textbf{1.000}$_{\pm 0.000}$ \\ 0.989$_{\pm 0.003}$ \\ 0.983$_{\pm 0.001}$}
    & \makecell{\textbf{1.000}$_{\pm 0.000}$ \\ 0.991$_{\pm 0.002}$ \\ 0.988$_{\pm 0.003}$}
    & \makecell{\textbf{1.000}$_{\pm 0.000}$ \\ 0.990$_{\pm 0.003}$ \\ 0.985$_{\pm 0.002}$} \\
    \rowcolor{rowgray}
    GaussianShading
    & \makecell{\textbf{1.000}$_{\pm 0.000}$ \\ 0.991$_{\pm 0.001}$ \\ 0.993$_{\pm 0.002}$}
    & \makecell{\textbf{1.000}$_{\pm 0.000}$ \\ \textbf{1.000}$_{\pm 0.000}$ \\ 0.995$_{\pm 0.001}$}
    & \makecell{\textbf{1.000}$_{\pm 0.000}$ \\ 0.997$_{\pm 0.003}$ \\ 0.995$_{\pm 0.002}$}
    & \makecell{\textbf{1.000}$_{\pm 0.000}$ \\ 0.998$_{\pm 0.002}$ \\ 0.992$_{\pm 0.003}$}
    & \makecell{\textbf{1.000}$_{\pm 0.000}$ \\ 0.997$_{\pm 0.001}$ \\ 0.993$_{\pm 0.001}$}
    & \makecell{\textbf{1.000}$_{\pm 0.000}$ \\ 0.997$_{\pm 0.001}$ \\ 0.994$_{\pm 0.001}$}
    & \makecell{\textbf{1.000}$_{\pm 0.000}$ \\ \textbf{1.000}$_{\pm 0.000}$ \\ \textbf{1.000}$_{\pm 0.000}$}
    & \makecell{\textbf{1.000}$_{\pm 0.000}$ \\ \textbf{1.000}$_{\pm 0.000}$ \\ \textbf{1.000}$_{\pm 0.000}$}
    & \makecell{\textbf{1.000}$_{\pm 0.000}$ \\ \textbf{1.000}$_{\pm 0.000}$ \\ \textbf{1.000}$_{\pm 0.000}$} \\
    PRC
    & \makecell{0.001$_{\pm 0.001}$ \\ 0.000$_{\pm 0.000}$ \\ 0.000$_{\pm 0.000}$}
    & \makecell{0.000$_{\pm 0.000}$ \\ 0.000$_{\pm 0.000}$ \\ 0.002$_{\pm 0.000}$}
    & \makecell{0.002$_{\pm 0.000}$ \\ 0.000$_{\pm 0.000}$ \\ 0.000$_{\pm 0.000}$}
    & \makecell{0.000$_{\pm 0.000}$ \\ 0.000$_{\pm 0.000}$ \\ 0.000$_{\pm 0.000}$}
    & \makecell{0.000$_{\pm 0.000}$ \\ 0.000$_{\pm 0.000}$ \\ 0.000$_{\pm 0.000}$}
    & \makecell{0.001$_{\pm 0.000}$ \\ 0.000$_{\pm 0.000}$ \\ 0.000$_{\pm 0.000}$}
    & \makecell{0.930$_{\pm 0.017}$ \\ 0.751$_{\pm 0.015}$ \\ 0.746$_{\pm 0.011}$}
    & \makecell{0.933$_{\pm 0.018}$ \\ 0.747$_{\pm 0.014}$ \\ 0.719$_{\pm 0.014}$}
    & \makecell{0.932$_{\pm 0.018}$ \\ 0.749$_{\pm 0.014}$ \\ 0.732$_{\pm 0.012}$} \\
    \midrule
    \rowcolor{rowgray}
    \textbf{DiffMark}
    & \makecell{\textbf{1.000}$_{\pm 0.000}$ \\ 0.999$_{\pm 0.001}$ \\ \textbf{1.000}$_{\pm 0.000}$}
    & \makecell{\textbf{1.000}$_{\pm 0.000}$ \\ \textbf{1.000}$_{\pm 0.000}$ \\ \textbf{1.000}$_{\pm 0.000}$}
    & \makecell{\textbf{1.000}$_{\pm 0.000}$ \\ \textbf{1.000}$_{\pm 0.000}$ \\ \textbf{1.000}$_{\pm 0.000}$}
    & \makecell{\textbf{1.000}$_{\pm 0.000}$ \\ \textbf{1.000}$_{\pm 0.000}$ \\ \textbf{1.000}$_{\pm 0.000}$}
    & \makecell{\textbf{1.000}$_{\pm 0.000}$ \\ \textbf{1.000}$_{\pm 0.000}$ \\ \textbf{1.000}$_{\pm 0.000}$}
    & \makecell{\textbf{1.000}$_{\pm 0.000}$ \\ \textbf{1.000}$_{\pm 0.000}$ \\ \textbf{1.000}$_{\pm 0.000}$}
    & \makecell{\textbf{1.000}$_{\pm 0.000}$ \\ \textbf{1.000}$_{\pm 0.000}$ \\ \textbf{1.000}$_{\pm 0.000}$}
    & \makecell{\textbf{1.000}$_{\pm 0.000}$ \\ \textbf{1.000}$_{\pm 0.000}$ \\ \textbf{1.000}$_{\pm 0.000}$}
    & \makecell{\textbf{1.000}$_{\pm 0.000}$ \\ \textbf{1.000}$_{\pm 0.000}$ \\ \textbf{1.000}$_{\pm 0.000}$} \\
    \bottomrule
    \end{tabular}%
    }
\end{table*}

\begin{table*}[ht]
    \centering
    \caption{\textbf{Removal and adversarial attack breakdown with FLUX on DiffusionDB, MS-COCO, and DALL-E3.} Values are mean TPR@1\%FPR, with standard deviations in subscript. Each cell lists DiffusionDB, MS-COCO, and DALL-E3 results from top to bottom. The best result for each attack and dataset, including ties, is in \textbf{bold}. Avg is the reported average within each attack group.}
    \label{tab:supp_removal_adv_flux}
    \resizebox{\textwidth}{!}{%
    \footnotesize
    \setlength{\tabcolsep}{4pt}
    \renewcommand{\arraystretch}{1.05}
    \begin{tabular}{l cccccc @{\hspace{10pt}} ccc}
    \toprule
    \multirow{2}{*}{\textbf{Method}}
    & \multicolumn{6}{c}{\textbf{Removal Attacks}}
    & \multicolumn{3}{c}{\textbf{Adversarial Attacks}} \\
    \cmidrule(lr){2-7} \cmidrule(lr){8-10}
    & \textbf{Regen-VAE} & \textbf{Regen-Diff} & \textbf{Rinse-2x-20} & \textbf{Rinse-2x-60} & \textbf{Rinse-2x-100} & \textbf{Avg} & \textbf{Adv-KLVAE8} & \textbf{Adv-RN18} & \textbf{Avg} \\
    \cmidrule(lr){1-1} \cmidrule(lr){2-2} \cmidrule(lr){3-3} \cmidrule(lr){4-4} \cmidrule(lr){5-5} \cmidrule(lr){6-6} \cmidrule(lr){7-7} \cmidrule(lr){8-8} \cmidrule(lr){9-9} \cmidrule(lr){10-10}
    \rowcolor{rowgray}
    Tree-Ring
    & \makecell{0.014$_{\pm 0.000}$ \\ 0.009$_{\pm 0.003}$ \\ 0.019$_{\pm 0.007}$}
    & \makecell{0.025$_{\pm 0.005}$ \\ 0.017$_{\pm 0.004}$ \\ 0.046$_{\pm 0.019}$}
    & \makecell{0.025$_{\pm 0.008}$ \\ 0.016$_{\pm 0.005}$ \\ 0.062$_{\pm 0.029}$}
    & \makecell{0.011$_{\pm 0.004}$ \\ 0.009$_{\pm 0.005}$ \\ 0.024$_{\pm 0.003}$}
    & \makecell{0.022$_{\pm 0.003}$ \\ 0.009$_{\pm 0.005}$ \\ 0.021$_{\pm 0.006}$}
    & \makecell{0.019$_{\pm 0.004}$ \\ 0.012$_{\pm 0.004}$ \\ 0.034$_{\pm 0.013}$}
    & \makecell{0.835$_{\pm 0.021}$ \\ 0.557$_{\pm 0.062}$ \\ 0.777$_{\pm 0.027}$}
    & \makecell{0.965$_{\pm 0.010}$ \\ 0.733$_{\pm 0.048}$ \\ 0.883$_{\pm 0.025}$}
    & \makecell{0.900$_{\pm 0.015}$ \\ 0.645$_{\pm 0.055}$ \\ 0.830$_{\pm 0.026}$} \\
    ShallowDiffuse
    & \makecell{0.997$_{\pm 0.001}$ \\ 0.966$_{\pm 0.007}$ \\ 0.982$_{\pm 0.005}$}
    & \makecell{\textbf{1.000}$_{\pm 0.000}$ \\ 0.998$_{\pm 0.000}$ \\ 0.999$_{\pm 0.001}$}
    & \makecell{\textbf{1.000}$_{\pm 0.000}$ \\ 0.998$_{\pm 0.002}$ \\ 0.999$_{\pm 0.001}$}
    & \makecell{0.999$_{\pm 0.001}$ \\ 0.989$_{\pm 0.002}$ \\ 0.997$_{\pm 0.001}$}
    & \makecell{0.995$_{\pm 0.002}$ \\ 0.935$_{\pm 0.003}$ \\ 0.966$_{\pm 0.021}$}
    & \makecell{0.998$_{\pm 0.001}$ \\ 0.977$_{\pm 0.002}$ \\ 0.989$_{\pm 0.005}$}
    & \makecell{\textbf{1.000}$_{\pm 0.000}$ \\ \textbf{1.000}$_{\pm 0.000}$ \\ \textbf{1.000}$_{\pm 0.000}$}
    & \makecell{\textbf{1.000}$_{\pm 0.000}$ \\ \textbf{1.000}$_{\pm 0.000}$ \\ \textbf{1.000}$_{\pm 0.000}$}
    & \makecell{\textbf{1.000}$_{\pm 0.000}$ \\ \textbf{1.000}$_{\pm 0.000}$ \\ \textbf{1.000}$_{\pm 0.000}$} \\
    \rowcolor{rowgray}
    StegaStamp
    & \makecell{\textbf{1.000}$_{\pm 0.000}$ \\ \textbf{1.000}$_{\pm 0.000}$ \\ \textbf{1.000}$_{\pm 0.000}$}
    & \makecell{\textbf{1.000}$_{\pm 0.000}$ \\ \textbf{1.000}$_{\pm 0.000}$ \\ \textbf{1.000}$_{\pm 0.000}$}
    & \makecell{\textbf{1.000}$_{\pm 0.000}$ \\ \textbf{1.000}$_{\pm 0.000}$ \\ \textbf{1.000}$_{\pm 0.000}$}
    & \makecell{0.979$_{\pm 0.001}$ \\ 0.993$_{\pm 0.005}$ \\ 0.959$_{\pm 0.018}$}
    & \makecell{0.747$_{\pm 0.004}$ \\ 0.920$_{\pm 0.005}$ \\ 0.756$_{\pm 0.056}$}
    & \makecell{0.945$_{\pm 0.001}$ \\ 0.983$_{\pm 0.002}$ \\ 0.943$_{\pm 0.015}$}
    & \makecell{\textbf{1.000}$_{\pm 0.000}$ \\ \textbf{1.000}$_{\pm 0.000}$ \\ \textbf{1.000}$_{\pm 0.000}$}
    & \makecell{\textbf{1.000}$_{\pm 0.000}$ \\ \textbf{1.000}$_{\pm 0.000}$ \\ \textbf{1.000}$_{\pm 0.000}$}
    & \makecell{\textbf{1.000}$_{\pm 0.000}$ \\ \textbf{1.000}$_{\pm 0.000}$ \\ \textbf{1.000}$_{\pm 0.000}$} \\
    RingID
    & \makecell{0.059$_{\pm 0.018}$ \\ 0.087$_{\pm 0.015}$ \\ 0.080$_{\pm 0.024}$}
    & \makecell{0.250$_{\pm 0.047}$ \\ 0.172$_{\pm 0.016}$ \\ 0.235$_{\pm 0.045}$}
    & \makecell{0.224$_{\pm 0.056}$ \\ 0.184$_{\pm 0.005}$ \\ 0.199$_{\pm 0.047}$}
    & \makecell{0.102$_{\pm 0.038}$ \\ 0.072$_{\pm 0.006}$ \\ 0.091$_{\pm 0.023}$}
    & \makecell{0.073$_{\pm 0.021}$ \\ 0.045$_{\pm 0.006}$ \\ 0.063$_{\pm 0.024}$}
    & \makecell{0.141$_{\pm 0.036}$ \\ 0.112$_{\pm 0.008}$ \\ 0.134$_{\pm 0.032}$}
    & \makecell{0.997$_{\pm 0.002}$ \\ 0.973$_{\pm 0.003}$ \\ 0.995$_{\pm 0.003}$}
    & \makecell{\textbf{1.000}$_{\pm 0.000}$ \\ 0.991$_{\pm 0.003}$ \\ 0.999$_{\pm 0.001}$}
    & \makecell{0.999$_{\pm 0.001}$ \\ 0.982$_{\pm 0.003}$ \\ 0.997$_{\pm 0.002}$} \\
    \rowcolor{rowgray}
    GaussianShading
    & \makecell{0.994$_{\pm 0.000}$ \\ 0.997$_{\pm 0.001}$ \\ 0.996$_{\pm 0.000}$}
    & \makecell{\textbf{1.000}$_{\pm 0.000}$ \\ \textbf{1.000}$_{\pm 0.000}$ \\ \textbf{1.000}$_{\pm 0.000}$}
    & \makecell{\textbf{1.000}$_{\pm 0.000}$ \\ \textbf{1.000}$_{\pm 0.000}$ \\ \textbf{1.000}$_{\pm 0.000}$}
    & \makecell{\textbf{1.000}$_{\pm 0.000}$ \\ 0.997$_{\pm 0.001}$ \\ 0.998$_{\pm 0.002}$}
    & \makecell{\textbf{1.000}$_{\pm 0.000}$ \\ 0.997$_{\pm 0.003}$ \\ 0.995$_{\pm 0.001}$}
    & \makecell{0.999$_{\pm 0.000}$ \\ 0.998$_{\pm 0.001}$ \\ 0.998$_{\pm 0.001}$}
    & \makecell{\textbf{1.000}$_{\pm 0.000}$ \\ \textbf{1.000}$_{\pm 0.000}$ \\ \textbf{1.000}$_{\pm 0.000}$}
    & \makecell{\textbf{1.000}$_{\pm 0.000}$ \\ \textbf{1.000}$_{\pm 0.000}$ \\ \textbf{1.000}$_{\pm 0.000}$}
    & \makecell{\textbf{1.000}$_{\pm 0.000}$ \\ \textbf{1.000}$_{\pm 0.000}$ \\ \textbf{1.000}$_{\pm 0.000}$} \\
    PRC
    & \makecell{0.000$_{\pm 0.000}$ \\ 0.000$_{\pm 0.000}$ \\ 0.000$_{\pm 0.000}$}
    & \makecell{0.000$_{\pm 0.000}$ \\ 0.000$_{\pm 0.000}$ \\ 0.001$_{\pm 0.001}$}
    & \makecell{0.000$_{\pm 0.000}$ \\ 0.001$_{\pm 0.001}$ \\ 0.000$_{\pm 0.000}$}
    & \makecell{0.000$_{\pm 0.000}$ \\ 0.000$_{\pm 0.000}$ \\ 0.000$_{\pm 0.000}$}
    & \makecell{0.000$_{\pm 0.000}$ \\ 0.000$_{\pm 0.000}$ \\ 0.000$_{\pm 0.000}$}
    & \makecell{0.000$_{\pm 0.000}$ \\ 0.000$_{\pm 0.000}$ \\ 0.000$_{\pm 0.000}$}
    & \makecell{0.799$_{\pm 0.004}$ \\ 0.492$_{\pm 0.007}$ \\ 0.680$_{\pm 0.029}$}
    & \makecell{0.942$_{\pm 0.004}$ \\ 0.630$_{\pm 0.005}$ \\ 0.813$_{\pm 0.015}$}
    & \makecell{0.870$_{\pm 0.002}$ \\ 0.561$_{\pm 0.006}$ \\ 0.746$_{\pm 0.022}$} \\
    \midrule
    \rowcolor{rowgray}
    \textbf{DiffMark}
    & \makecell{\textbf{1.000}$_{\pm 0.000}$ \\ \textbf{1.000}$_{\pm 0.000}$ \\ \textbf{1.000}$_{\pm 0.000}$}
    & \makecell{\textbf{1.000}$_{\pm 0.000}$ \\ \textbf{1.000}$_{\pm 0.000}$ \\ \textbf{1.000}$_{\pm 0.000}$}
    & \makecell{\textbf{1.000}$_{\pm 0.000}$ \\ \textbf{1.000}$_{\pm 0.000}$ \\ \textbf{1.000}$_{\pm 0.000}$}
    & \makecell{\textbf{1.000}$_{\pm 0.000}$ \\ \textbf{1.000}$_{\pm 0.000}$ \\ \textbf{1.000}$_{\pm 0.000}$}
    & \makecell{\textbf{1.000}$_{\pm 0.000}$ \\ \textbf{1.000}$_{\pm 0.000}$ \\ \textbf{1.000}$_{\pm 0.000}$}
    & \makecell{\textbf{1.000}$_{\pm 0.000}$ \\ \textbf{1.000}$_{\pm 0.000}$ \\ \textbf{1.000}$_{\pm 0.000}$}
    & \makecell{\textbf{1.000}$_{\pm 0.000}$ \\ \textbf{1.000}$_{\pm 0.000}$ \\ 0.999$_{\pm 0.001}$}
    & \makecell{\textbf{1.000}$_{\pm 0.000}$ \\ \textbf{1.000}$_{\pm 0.000}$ \\ \textbf{1.000}$_{\pm 0.000}$}
    & \makecell{\textbf{1.000}$_{\pm 0.000}$ \\ \textbf{1.000}$_{\pm 0.000}$ \\ 0.999$_{\pm 0.001}$} \\
    \bottomrule
    \end{tabular}%
    }
\end{table*}

\subsection{Visualization of watermark residuals}
\label{sec:supp_residuals}

\begin{figure*}[ht]
    \centering
    \textbf{(a) DiffusionDB}\\[3pt]
    % Replace with the actual DiffusionDB figure filename.
    \includegraphics[width=\textwidth]{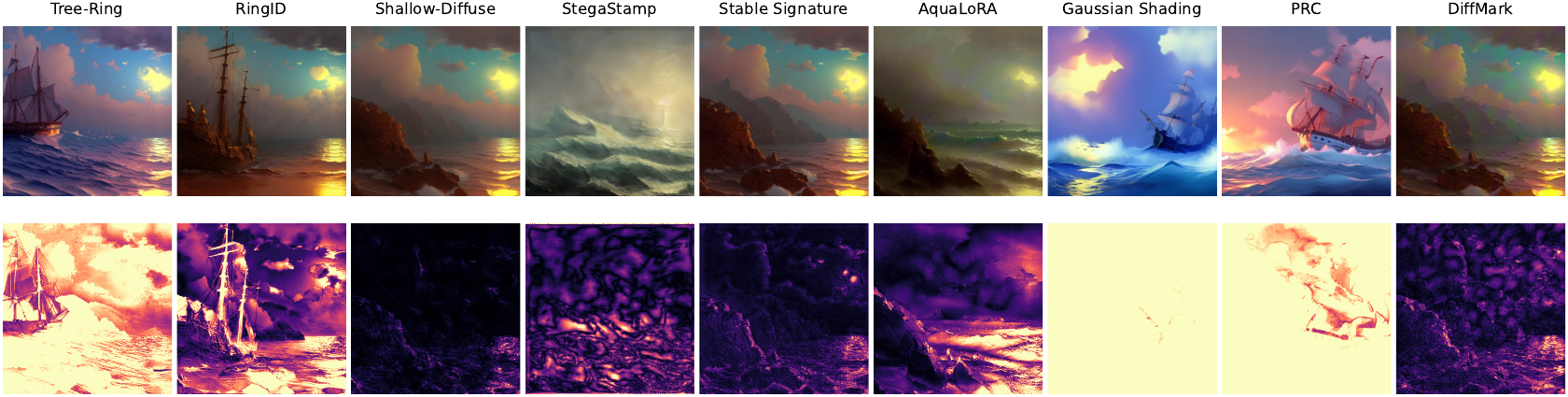}\\[6pt]
    \textbf{(b) MS-COCO}\\[3pt]
    \includegraphics[width=\textwidth]{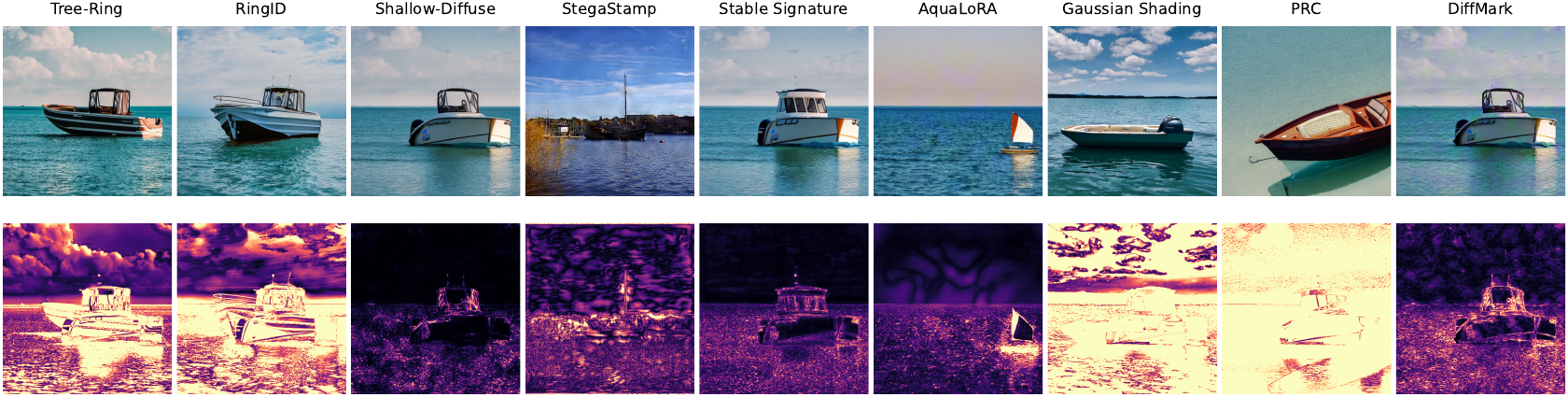}\\[6pt]
    \textbf{(c) DALL-E3}\\[3pt]
    \includegraphics[width=\textwidth]{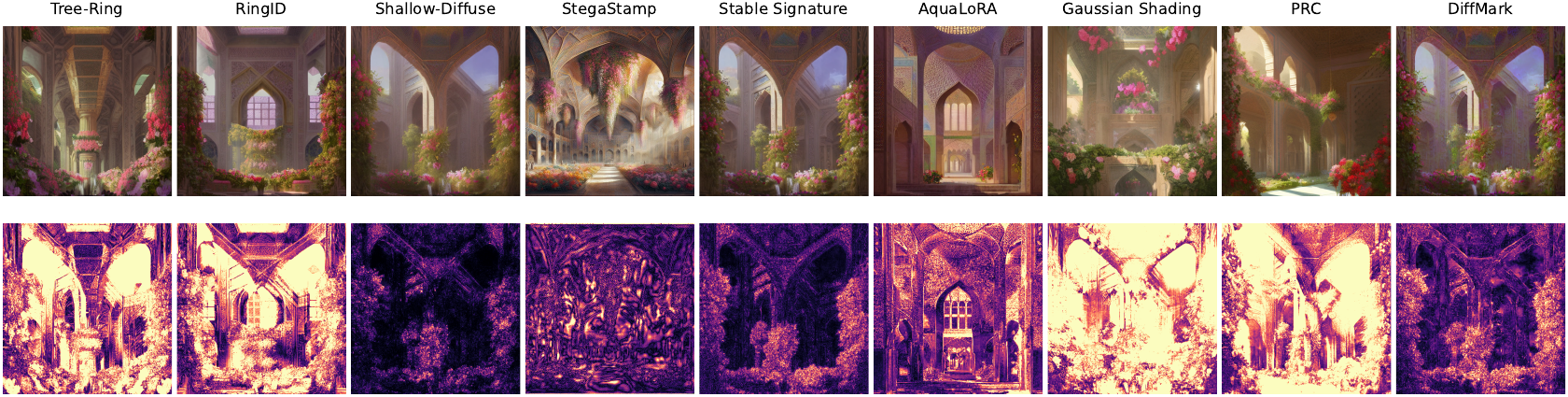}
    \caption{\textbf{Watermark residuals on three datasets.}
    For each dataset, the upper row shows watermarked images
    from \ours{} and the baselines, and the lower row visualizes
    residual magnitudes relative to their matched clean
    counterparts.}
    \label{fig:supp_residuals}
\end{figure*}

\cref{fig:supp_residuals} compares the visual changes introduced by \ours{} and the baselines on DiffusionDB, MS-COCO, and DALL-E3. For each method, we compare the watermarked image $x^{\rm wm}$
with its clean counterpart $x^{\rm clean}$, generated using the same prompt, initial noise, and sampling settings. We visualize the magnitude of the pixel-space residual
$r=x^{\rm wm}-x^{\rm clean}$. These maps reveal where watermarking changes the image and
complement the quantitative image-quality results by exposing local texture changes and broader structural differences.

\subsection{Additional qualitative visualizations}
\label{sec:supp_qualitative}

\begin{figure*}[p]
    \centering
    % Replace with the actual DiffusionDB grid filename.
    \includegraphics[width=\textwidth]{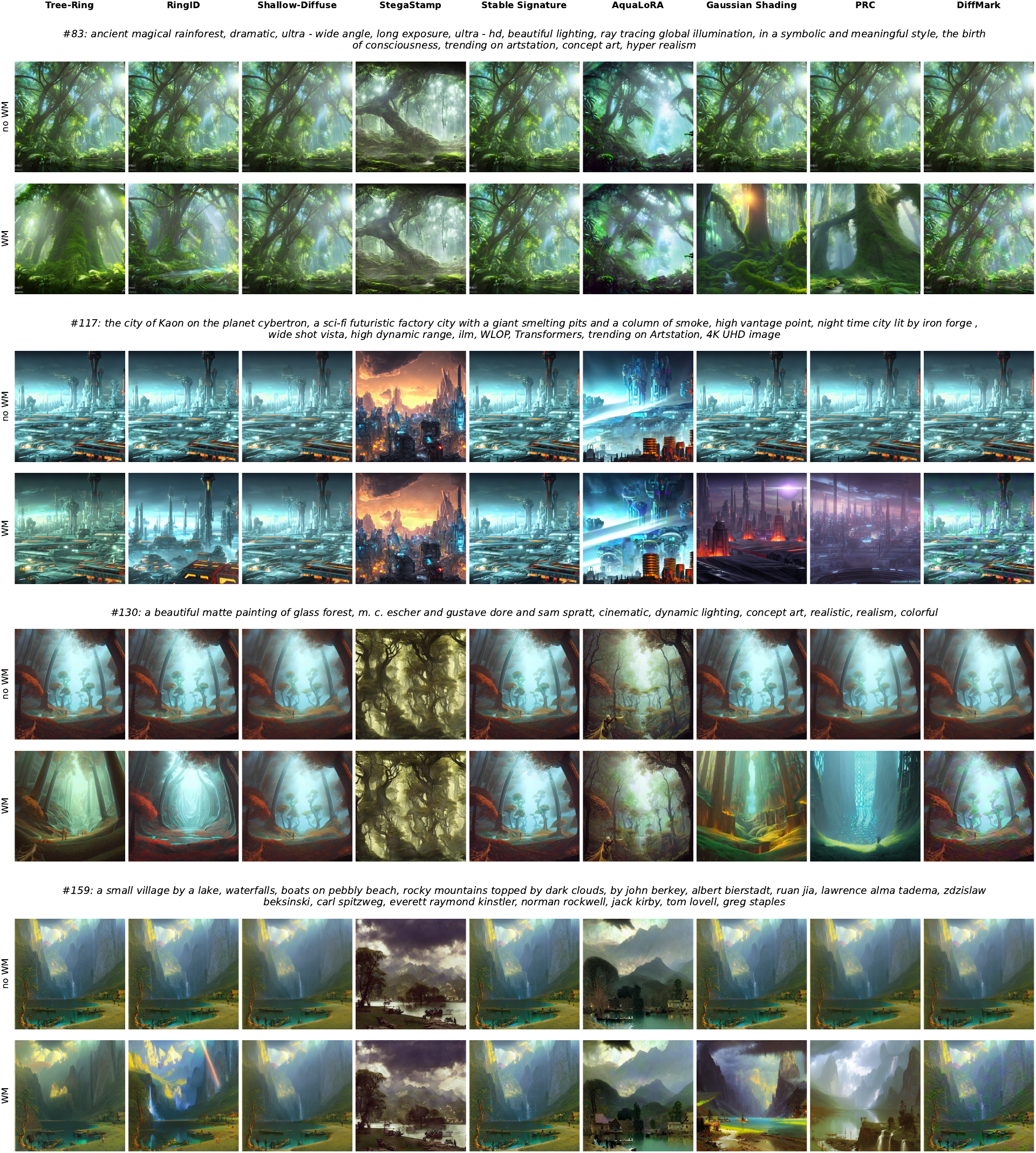}
    \caption{\textbf{Additional qualitative comparisons on
    DiffusionDB.} Each method column shows paired unwatermarked
    (no WM) and watermarked (WM) images for the same prompt
    and sampling seed. \ours{} appears in the rightmost column.}
    \label{fig:supp_qualitative_diffusiondb}
\end{figure*}

\begin{figure*}[p]
    \centering
    % Replace with the actual MS-COCO grid filename.
    \includegraphics[width=\textwidth]{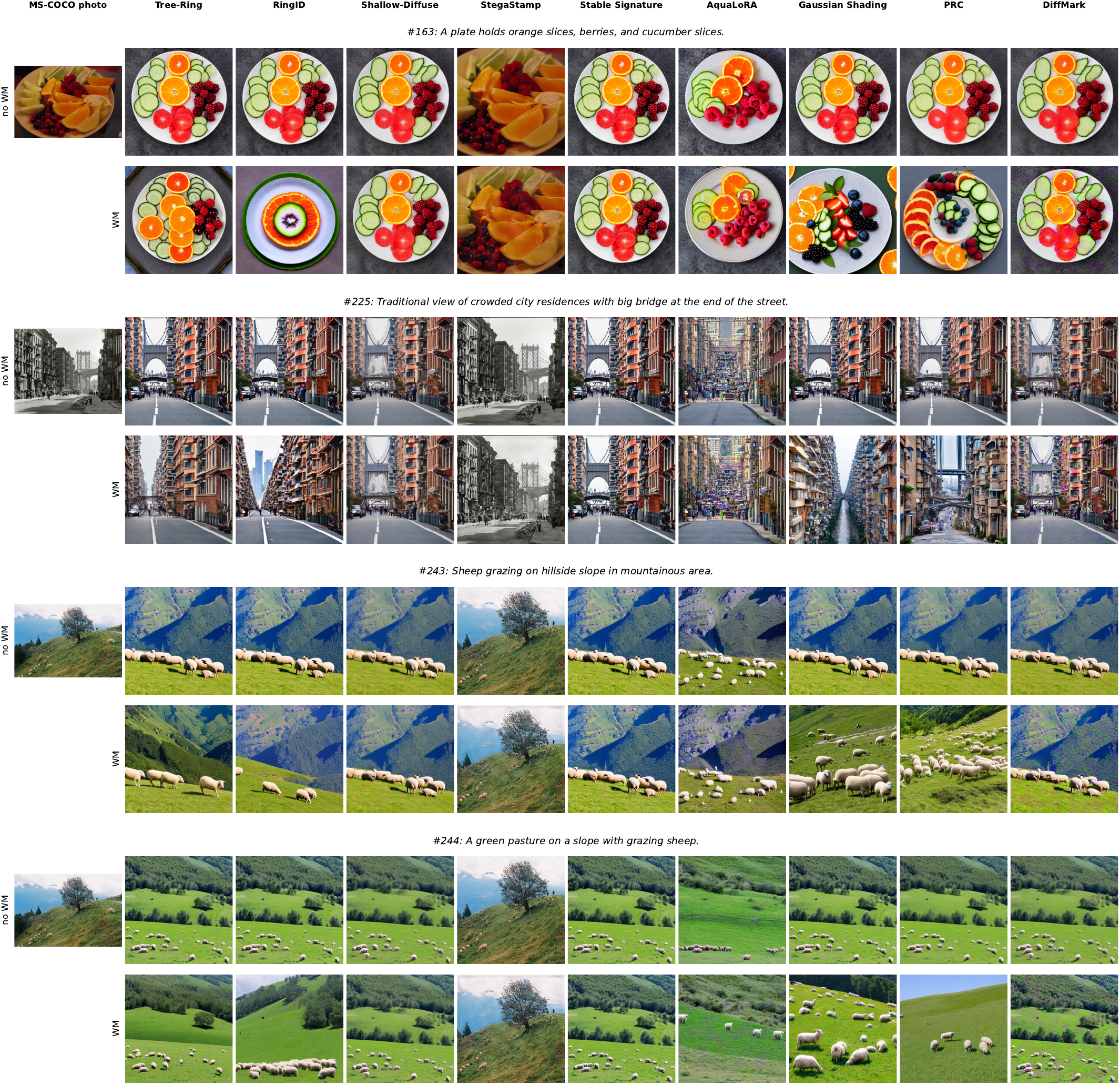}
    \caption{\textbf{Additional qualitative comparisons on
    MS-COCO.} Each method column shows paired unwatermarked
    (no WM) and watermarked (WM) images for the same prompt
    and sampling seed. \ours{} appears in the rightmost column.}
    \label{fig:supp_qualitative_mscoco}
\end{figure*}

\begin{figure*}[p]
    \centering
    \includegraphics[width=\textwidth]{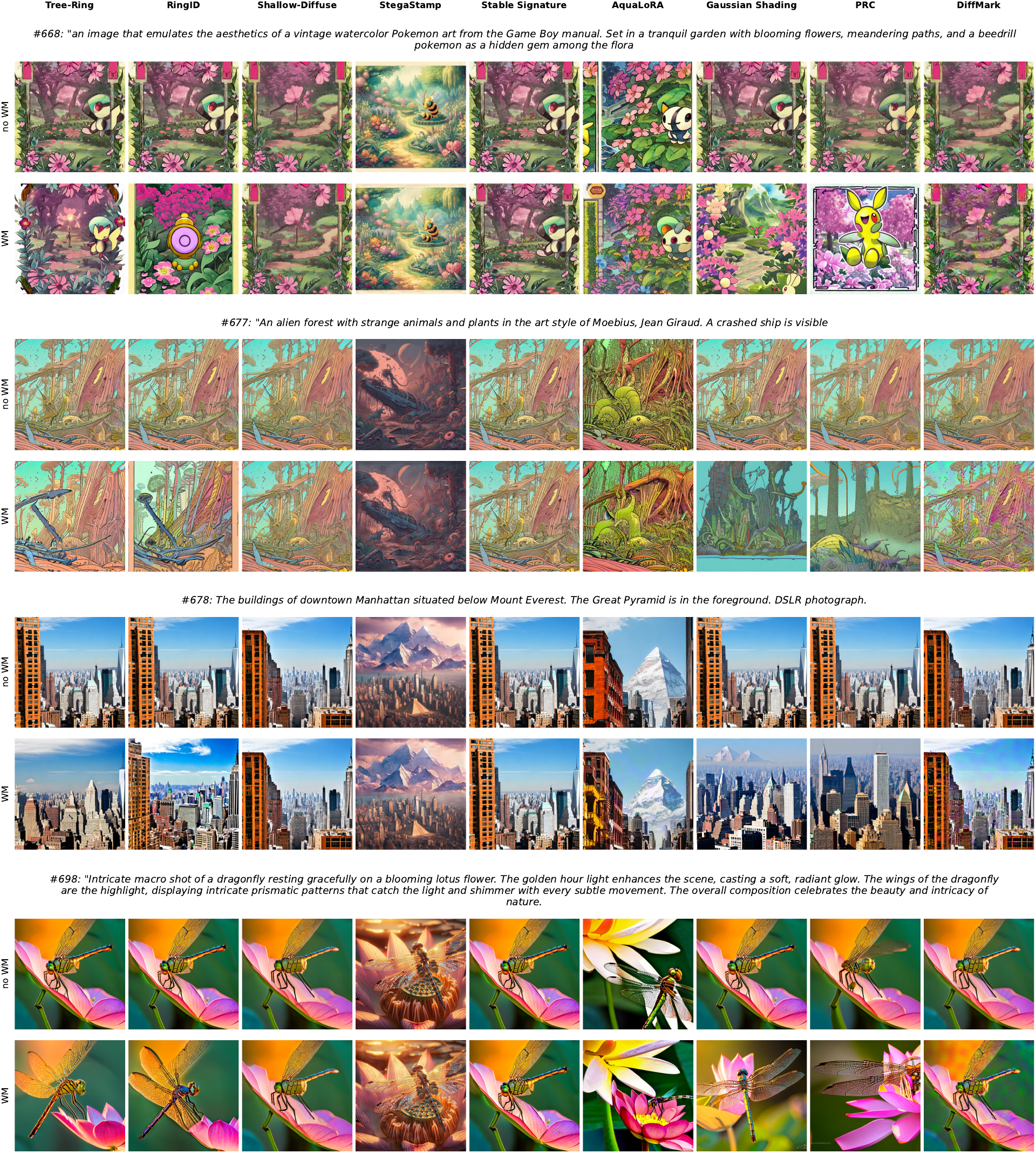}
    \caption{\textbf{Additional qualitative comparisons on
    DALL-E3.} Each method column shows paired unwatermarked
    (no WM) and watermarked (WM) images. The prompt is displayed
    above each pair of rows. \ours{} appears in the rightmost
    column.}
    \label{fig:supp_qualitative_dalle3}
\end{figure*}

\cref{fig:supp_qualitative_diffusiondb,fig:supp_qualitative_mscoco,fig:supp_qualitative_dalle3} provide additional comparisons of \ours{} and the baselines
on DiffusionDB, MS-COCO, and DALL-E3. Each example pairs an
unwatermarked image with its watermarked counterpart using
the same prompt and sampling seed. These pairs allow direct
inspection of changes in scene composition, object placement,
color, and texture.

The DALL-E3 examples cover stylized illustrations, complex
scene compositions, and macro photography. In these examples,
\ours{} largely retains the scene layout and object positions,
while introducing some local color and texture changes.
These visualizations complement the residual maps by showing
how the watermark affects the appearance of the generated image.

% =====================================================================
\section{Extended Related Work}
\label{sec:supp_related}
\textbf{Post-hoc watermarking.} Classical methods modify transform-domain coefficients, for example using DCT or SVD~\cite{al2007combined,navas2008dwt}, but are vulnerable to modern compression and editing. HiDDeN~\cite{zhu2018hidden} introduced an encoder, noise layer, and decoder architecture. MBRS~\cite{jia2021mbrs} alternates real and simulated JPEG, while StegaStamp~\cite{tancik2020stegastamp} trains against print-and-capture distortions. RivaGAN~\cite{zhang2019robust} and TrustMark~\cite{bui2023trustmark} use adversarial training, and SepMark~\cite{wu2023sepmark} separates robust and semi-robust decoders. Fernandez \etal~\cite{fernandez2022watermarking} embed watermarks in a self-supervised latent space with per-image optimization, and WAM~\cite{sander2025watermark} formulates watermarking as segmentation. These methods operate on already generated images, independently of the generation process.

Our frequency-based embedding and extraction adapt classical spread-spectrum patterns recovered through correlation~\cite{cox1997secure}, informed embedding~\cite{miller2004applying}, and search-based alignment correction~\cite{pereira2000robust} to the latent space of a frozen generator.

\textbf{One-step extraction.} FARI~\cite{yangfari} shows that inversion can use fewer steps than generation and fine-tunes LoRA adapters for one-step inversion at each checkpoint. OSI~\cite{chen2026osi} uses a one-step classifier initialized from diffusion components to recover the signs needed by Gaussian Shading. Both accelerate extraction of watermarks embedded in $z_T$ and require adaptation to each checkpoint. \ours learns to recover the key directly from VAE-encoded image latents, without reconstructing $z_T$.

\textbf{Inference-time optimization.} OptMark~\cite{xing2026optmark} optimizes a structural watermark early and a detail watermark late in the trajectory for each image. It uses regularization and adjoint gradients to control image quality and memory use. \ours trains its watermark modules before generation. Embedding then uses one encoder call, standard sampling, and up to $R$ correction rounds with VAE decoding and re-encoding.

\section{Limitations}
\label{sec:supp_limitations}

\textbf{Adaptation to new latent spaces.}
The watermark encoder and decoder operate in the generator's VAE latent space, and the frequency-based patterns are designed for that representation. Changes in latent dimensionality and sampling dynamics can therefore require module adaptation and pattern recalibration. In our experiments, the compatible checkpoints support direct reuse, while SD3 and FLUX require adaptation of the watermark modules. Generator weights remain unchanged throughout this process. Extending direct reuse to a broader range of latent representations remains a direction for future work.

The measurements show lower normalized signed gains for longer imprinting windows and positive signed gains for the tested key pairs. They characterize latent propagation, not image-level fidelity or bit recovery. The identity in \cref{thm:channel}(ii) establishes separation for these pairs, but does not provide a guarantee for all possible keys. The one-step Jacobian structure in \cref{lem:psd} assumes an exact posterior-mean denoiser without guidance; guidance and distillation can violate this structure. Similarly, the Gaussian model in \cref{prop:gauss} is illustrative and does not capture the early-step amplification observed in the tested samplers.

On SD3 and FLUX, rounding in bf16/fp16 changes the direction of the watermark response relative to higher-precision sampling, while its magnitude remains closer. We therefore evaluate key-pair separation at generation precision separately from the derivative analysis. Robustness to image manipulation is evaluated empirically on the tested generators and attacks. Tighter bounds for guided and distilled samplers could extend the theoretical guarantees.

\end{document}